%% file: main.tex
\documentclass{article}

\usepackage[preprint]{neurips_2023}

\input{math_commands.tex}

\usepackage{hyperref}
\usepackage{url}

\usepackage[utf8]{inputenc} % allow utf-8 input
\usepackage[T1]{fontenc}    % use 8-bit T1 fonts
\usepackage{booktabs}       % professional-quality tables
\usepackage{amsfonts}       % blackboard math symbols
\usepackage{nicefrac}       % compact symbols for 1/2, etc.
\usepackage{microtype}      % microtypography
\usepackage{xcolor}         % colors

\usepackage{verbatim}
\usepackage{amsmath}
\usepackage{amsthm}
\theoremstyle{definition}
\newtheorem{definition}{Definition}

\theoremstyle{assumption}
\newtheorem{assumption}{Assumption}

\newtheorem{theorem}{Theorem}[section]

\newtheorem{lemma}[theorem]{Lemma}

\usepackage{graphicx}
\usepackage{wrapfig}
\usepackage{caption}
\usepackage{subcaption}
\usepackage{pgf}
\usepackage{amssymb}
\usepackage{algorithm}              % algorithm
\usepackage[noend]{algpseudocode}   % contain algorithmc.sty

\title{Stochastic Gradient Langevin Dynamics Based on Quantization with Increasing Resolution}

\author{%
  Jinwuk Seok \\
  Future Computing Research Division, Artificial Intelligence Research Lab\\
  ETRI\\
  Daejeon, South KOREA, 34129 \\
  \texttt{jnwseok@etri.re.kr} \\
  \And
  Changsik Cho \\
  Future Computing Research Division, Artificial Intelligence Research Lab\\
  ETRI\\
  Daejeon, South KOREA, 34129 \\
  \texttt{cscho@etri.re.kr} \\
}

\begin{document}

\maketitle

%%%%%%%%%%%%%%%%%%%%%%%%%%%%%%%%%%%%%%%%%%%%%%%%%%%%%%%%%%%%
% abstract
%%%%%%%%%%%%%%%%%%%%%%%%%%%%%%%%%%%%%%%%%%%%%%%%%%%%%%%%%%%%
\begin{abstract}
Stochastic learning dynamics based on Langevin or Levy stochastic differential equations (SDEs) in deep neural networks control the variance of noise by varying the size of the mini-batch or directly those of injecting noise.
Since the noise variance affects the approximation performance, the design of the additive noise is significant in SDE-based learning and practical implementation.
In this paper, we propose an alternative stochastic descent learning equation based on quantized optimization for non-convex objective functions, adopting a stochastic analysis perspective. 
The proposed method employs a quantized optimization approach that utilizes Langevin SDE dynamics, allowing for controllable noise with an identical distribution without the need for additive noise or adjusting the mini-batch size.
Numerical experiments demonstrate the effectiveness of the proposed algorithm on vanilla convolution neural network(CNN) models and the ResNet-50 architecture across various data sets. Furthermore, we provide a simple PyTorch implementation of the proposed algorithm.
\end{abstract}

%%%%%%%%%%%%%%%%%%%%%%%%%%%%%%%%%%%%%%%%%%%%%%%%%%%%%%%%%%%%
\section{Introduction}
%%%%%%%%%%%%%%%%%%%%%%%%%%%%%%%%%%%%%%%%%%%%%%%%%%%%%%%%%%%%
Stochastic analysis for a learning equation based on stochastic gradient descent (SGD) with a finite or an infinitesimal learning rate has been an essential research topic to improve machine learning performance.
Particularly, the linear scaling rule(LSR) for SGD discovered by \citet{krizhevsky, Pratik_2018} and \citet{goyal_2018} independently provides an essential guide to select or control the learning rate corresponding to the size of the mini-batch. 
More crucially,  it gives a fundamental framework of stochastic analysis for the learning equation in current deep neural networks(DNN). 
However, the early analysis of SDE-based SGD encountered counterexamples, as demonstrated by \citet{Hoffer_2017, Shallue_2018, Zhang_NEURIPS2019}.  
Those works claim that the SGD with a momentum term or an appropriate learning rate represents superior performance to the SGD with a varying size mini-batch (\citet{Stephan_JMLR, Kidambi_2018_ITA, Liu_ICLR_2020} ), even though the noise term gives improved classification performance. 
As related research progresses, recent studies reached the following consensus: for an objective function being close to a standard convex, the SGD with mini-batch represents better performance, while for an objective function being a non-convex or curvature dominated, the SGD with momentum is better (\citet{Ma_pmlr-2018, Zhang_NEURIPS2019, Smith_ICML_2019} and \citet{Smith_ICML_2020}).  

The other research topic of the stochastic analysis for SGD is whether or not the induced SGD noise is Gaussian.
\citet{Simsekli_pmlr_2019, Nguyen_NEURIPS2019} suggested that SGD noise is a heavy-tailed distribution. 
This argument means that if SGD noise is not Gaussian, we should analyze SGD as a Levy process-based SDE instead of the standard SDE frame. 
For these claims, \citet{Wu_ICML_2020, Cheng_2020_ICML} and \citet{Li_NEURIPS2022} revealed that the third or higher order moments in SGD noise have minimal effect on accuracy performance, while the second moment has a significant impact.
In other words, the standard SDE is still valid for the stochastic analysis of SGD noise in the learning equation because the second moment of noise is the core component of the standard SDE. 
As the recent research substantiates the validation of the stochastic analysis for SGD noise based on the standard SDE,  \citet{Li-2019-JMLR, Granziol_2022_JMLR, Malladi_NEURIPS2022, Kalil_NEURIPS2022} and \citet{Li_NEURIPS2022} reinterpreted the conventional algorithm based on the standard SDE. 
Moreover, with the advent of novel algorithms and comprehensive analyses in noisy SGD research (e.g., the works of \citet{Fonseca_NEURIPS2022, Masiha_NEURIPS2022} and \citet{Altschuler_NEURIPS2022}), standard SDE-based noisy SGD algorithm is gaining widespread popularity.

Another research is stochastic gradient Langevin dynamics(SGLD), which injects an isotropic noise, such as the Wiener process, into SGD (\citet{Welling_ICML_2011, Brosse_NEURIPS_2018, Dalalyan_2019, Cheng_2020} and \citet{Zhang_ICML_2022}).
Unlike the noise derived from LSR, the noise introduced by SGLD, which originates from Markov Chain Monte Carlo, consists of independent and identically distributed (I.I.D.) components. As a result, we can readily apply this noise to a conventional learning algorithm to enhance its performance by robustness to non-convex optimization((\citet{Raginsky_2017, Xu_NEURIPS_2018, Wenlong_ICML_2018} and \citet{Wang_NEURIPS_2021}). 
However, LSR-based SGD and SGLD require additional processes such as warm-up(\citet{goyal_2018}), extra computation according to stochastic variance amplified gradient(SVAG)(\citet{Li_NEURIPS2021} and \citet{Malladi_NEURIPS2022}), or an identical random number generator for SGLD. Furthermore, with the advancement of research on distribution/federated learning, there is growing opposition to increasing the size of mini-batches due to practical considerations in optimization techniques. 
\citet{Tao_ICLR_2020} argued that in a distributed system with a heterogeneous hardware environment, including small computational devices, we could not expect a model learned using generalized large-batch SGD to be suitable. Therefore, they advocate for using small-sized batch SGD in such environments.

This paper introduces an alternative learning algorithm based on quantized optimization for SGLD, to address the practical issues related to LSR-based SGD and SGLD.
The proposed methodology makes the following contributions.

\textbf{Optimization based on Quantization}
We present the learning equation based on the quantized optimization theory, incorporating the white noise hypothesis (WNH) as suggested by \citet{Zamir_1996, Benedetto_2004, Gray:2006} and \citet{Jimnez_2007}. 
The WNH posits that the quantization error follows an I.I.D. white noise under regular conditions with sufficiently large sample data.
While the primary goal of quantization is to reduce computational burden by simplifying data processing in conventional artificial intelligence and other signal engineering(\citet{Seide_2014, Christopher_2015, Song_2015, Wen_2016, Sangil_2019} and \citet{Xiaoyun_2019}), in our work, quantization serves as the core technology for enhancing optimization performance during the learning process. 
Additionally, the quantization error effectively incorporates the various noise generated by the algorithm and establishes it as additive white noise according to the WNH, thereby facilitating the SDE analysis for optimization.

\textbf{Controllable Quantization Resolution}
By defining the quantization resolution function based on time and other parameters, we propose an algorithm that computes the quantization level.
Controlling the noise variance induced by the quantization error is essential to apply the quantization error to an optimizer effectively. While the WNH allows us to treat the quantization error as I.I.D. white noise, it alone does not guarantee optimal results if the uncontrolled variance exists.
Therefore, learning based on SGLD becomes feasible without a random number generator required by SGLD or MCMC.  
Furthermore, similar to increasing the mini-batch size in LSR-based SGD, we can develop a scheduler using controlled quantization resolution for optimization.

\textbf{Non-convex Optimization}
The proposed optimization algorithm demonstrates robust optimization characteristics for non-convex objective functions.
The quantized optimization algorithm outperforms MCMC-based optimization methods in combinatorial optimization problems, such as simulated and quantum annealing. 
Although further empirical evidence is needed, this result indicates that quantized optimization is a viable approach for non-convex optimization problems.
We analyze the proposed algorithm's weak and local convergence to substantiate this claim.

%%%%%%%%%%%%%%%%%%%%%%%%%%%%%%%%%%%%%%%%%%%%%%%%%%%%%%%%%%%%
\section{Preliminaries and Overview}
%%%%%%%%%%%%%%%%%%%%%%%%%%%%%%%%%%%%%%%%%%%%%%%%%%%%%%%%%%%%
%===========================================================
\subsection{Standard Assumptions for the Objective Function}
\label{ch02-01}
%===========================================================
We establish the objective function $f: \displaystyle \R^d \mapsto \displaystyle \R, \; f \in C^2$ for a given state parameter(e.g., weight vectors in DNN) $\boldsymbol{x}_{t_e} \in \displaystyle \R^d$ such that 
\begin{equation}
\label{ch02_eq01}
f(\boldsymbol{x}_{t_e}) 
\triangleq \frac{1}{N_T} \sum_{k=1}^{N_T} \bar{f}_k(\boldsymbol{x}_{t_e})
= \frac{1}{B \cdot n_{B}} \sum_{\tau=1}^{B} \sum_{i=1}^{n_{B}} \bar{f}_{\tau \cdot n_{B} + i} (\boldsymbol{x}_{t_e})
, \quad \bar{f}_k: \displaystyle \R^d \rightarrow \displaystyle \R
\end{equation}
, where $t_e$ denotes a discrete time index indicating the epoch, $\bar{f}_k: \displaystyle \R^d \mapsto \displaystyle \R, \;\bar{f}_k \in C^2$ denotes a loss function for the $k$-th sample data, $B \in \mathbb{Z}^+$ denotes the number of mini-batches, $n_{B}$ denotes the equivalent number of samples for each mini-batch $B_{j}$, and $N_T$ denotes the number of samples for an input data-set such that $N_T = B \cdot n_{B}$.  

In practical applications in DNN, the objective function is the summation of entropies with a distribution based on an exponential function such as the softmax or the log softmax, so the analytic assumption of the objective function (i.e. $f \in C^2$) is reasonable.
Additionally, since a practical framework for DNN updates the parameter $\boldsymbol{x}_{t_e}$ with a unit mini-batch, we can rewrite the objective function as an average of samples in a mini-batch such that 
\begin{equation}
\label{ch02_eq02}
f(\boldsymbol{x}_{t}) 
= \frac{1}{B} \sum_{\tau=0}^{B-1} \tilde{f}_{\tau}(\boldsymbol{x}_{t_e - 1 + \tau/B}),\; t \in \mathbb{R}[t_e-1, t_e] 
\quad \because \tilde{f}_{\tau}(\boldsymbol{x}_{t_e}) = \frac{1}{n_{B}} \sum_{i=1}^{n_{B}} \bar{f}_{\tau \cdot n_{B} + i} (\boldsymbol{x}_{t_e}).
\end{equation}
Under the definition of the objective function, we establish the following assumptions:
%==================================== 
\begin{assumption}  %Lipschitz continuous assumption 1
\label{assum01}
%====================================
For $\boldsymbol{x}_t \in B^o (\boldsymbol{x}^*, \rho)$, there exists a positive value $L_0$ with respect to $f$ such that
\begin{equation}
| f(\boldsymbol{x}_t) - f(\boldsymbol{x}^*) | \leq L_0 \| \boldsymbol{x}_t - \boldsymbol{x}^* \|, \quad \forall t > t_0 
\label{as01_eq01}    
\end{equation}
, where $B^o (\boldsymbol{x}^* , \rho)$ denotes an open ball $B^o (\boldsymbol{x}^* , \rho) = \{ \boldsymbol{x} | \| \boldsymbol{x} - \boldsymbol{x}^* \| < \rho \}$ for all $\rho \in \displaystyle \R^{+}$, and $\boldsymbol{x}^* \in \displaystyle \R^d$ denotes the unique globally optimal point such that $f(\boldsymbol{x}^*) < f(\boldsymbol{x}_{t})$.
Furthermore, we define the Lipschitz constants $L_1 > 0$ for the first-order derivation of $f$, such that 
\begin{equation}
\| \nabla_{\boldsymbol{x}} f(\boldsymbol{x}_t) - \nabla_{\boldsymbol{x}} f(\boldsymbol{x}^*) 
\| \leq L_1 \| \boldsymbol{x}_t - \boldsymbol{x}^* \|.
\label{as01_eq02}    
\end{equation}
\end{assumption}
In Assumption $\ref{assum01}$, we employ the time index $t \in \displaystyle \R^+$ instead of $t_e \in \mathbb{Z}^+$ to apply the assumption to an expanded continuous approximation.  
We assume that the set of the discrete epoch-unit time $\{t_e \vert t_e \in \mathbb{Z}^+ \}$ is a subset of the set of the continuous time $\{t \vert t \in \displaystyle \R^+ \}$. 
%===========================================================
\subsection{Definition and Assumptions for Quantization}
\label{ch02-02}
%===========================================================
The conventional research relevant to signal processing defines a quantization such that $x^Q \triangleq \lfloor \frac{x}{\Delta} + \frac{1}{2} \rfloor \Delta$ for $x \in \displaystyle \R$, where $\Delta \in \mathbb{Q}^+$ denotes a fixed valued quantization step. 
We provide a more detailed definition of quantization to explore the impact of the quantization error, using the quantization parameter as the reciprocal of the quantization step such that $Q_p \triangleq \Delta^{-1}$.   
%==================================== 
\begin{definition}  % definition 2
\label{def_q01}
%====================================
For $x \in \displaystyle \R$, we define the quantization of $x$ as follows:
\begin{equation}
x^Q \triangleq \frac{1}{Q_p} \lfloor Q_p \cdot  (x + 0.5 \cdot Q_p^{-1}) \rfloor
= \frac{1}{Q_p} \left( Q_p \cdot x + \varepsilon^q \right) = x + \varepsilon^q Q_p^{-1}, \quad x^Q \in \mathbb{Q}
\label{define_eq01}
\end{equation}
, where $\lfloor x \rfloor \in \mathbb{Z}$ denotes the floor function such that $\lfloor x \rfloor \leq x$ for all $x \in \displaystyle \R$, $Q_p \in \mathbb{Q}^+$ denotes the quantization parameter, and $\varepsilon^q \in \displaystyle \R$ is the factor for quantization such that $\varepsilon^q \in \displaystyle \R[-\frac{1}{2}, \frac{1}{2})$. 

Furthermore, for a given normal Euclidean basis $\{\boldsymbol{e}^{(i)}\}_{i=1}^d$, we can write a vector $\boldsymbol{x} \in \displaystyle \R^d$ such that $\boldsymbol{x} \triangleq \sum_{i=1}^d (\boldsymbol{x} \cdot \boldsymbol{e}^{(i)}) \boldsymbol{e}^{(i)}$. 
Using these notations, we can define the quantization of a vector $\boldsymbol{x}^Q \in \mathbb{Q}^d$ and the vector-valued the quantization error $\boldsymbol{\epsilon}^q \triangleq \boldsymbol{x}^Q - \boldsymbol{x} = Q_p^{-1} \boldsymbol{\varepsilon}^q \in \displaystyle \R^d$ as follows:
\begin{equation}
\boldsymbol{x}^Q \triangleq 
\sum_{i=1}^d (\boldsymbol{x} \cdot \boldsymbol{e}^{(i)})^Q \boldsymbol{e}^{(i)} 
\implies \boldsymbol{\epsilon}^q = Q_p^{-1} \boldsymbol{\varepsilon}^q = \sum_{i=1}^d \left( (\boldsymbol{x}^Q - \boldsymbol{x}) \cdot \boldsymbol{e}^{(i)} \right) \boldsymbol{e}^{(i)}.
\label{define_eq01-01}
\end{equation}
\end{definition}
We distinguish the scalar factor for quantization $\varepsilon^q \in \displaystyle \R[-1/2, 1/2)$, the vector valued factor $\boldsymbol{\varepsilon}^q \in \displaystyle \R^d[-1/2, 1/2)$, the scalar valued quantization error $\epsilon^q = Q_p^{-1}\varepsilon^q \in \displaystyle \R[-Q_p^{-1}/2, Q_p^{-1}/2)$, and the vector valued quantization error $\boldsymbol{\epsilon}^q \in \displaystyle \R^d[-\frac{Q_p^{-1}}{2}, \frac{Q_p^{-1}}{2})$ respectively.
%====================================
\begin{definition}  % definition 3 
\label{def_q02}
%====================================
We define the quantization parameter $Q_p : \displaystyle \R^d \times \displaystyle \R^{++} \mapsto \mathbb{Q}^{+}$ such that 
\begin{equation}
Q_p (\boldsymbol{\varepsilon}^q, t) = \eta(\boldsymbol{\varepsilon}^q) \cdot b^{\bar{p}(t)} 
\label{def_eq02}    
\end{equation}
, where $\eta : \displaystyle \R^d \mapsto \mathbb{Q}^{++}$ denotes the auxiliary function of the quantization parameter, $b \in \mathbb{Z}^+$ is the base, and $\bar{p} :\displaystyle \R^{++} \mapsto \mathbb{Z}^+$ denotes the power function such that $\bar{p}(t) \uparrow \infty \; \text{ as } \; t \rightarrow \infty$.
If $\eta$ is a constant value, the quantization parameter $Q_p$ is a monotone increasing function with respect to $t$. 
\end{definition}
In definition \ref{def_q02}, we can establish an intuitive stochastic approximation based on the random variable with Gaussian distribution by an appropriate transformation through the auxiliary function $\eta$, even though the probability density function of the quantization error is a uniform distribution.
We'll investigate the stochastic approximations depending on the function $\eta$ in the following chapter.

As a next step, we establish the following assumptions to define the statistical properties of the quantization error.
%==================================== 
\begin{assumption}  % assumption 2 : Statistical properties of quantization, WNH
\label{assum02}
%====================================
The probability density function of the quantization error $\epsilon^q$ is a uniform distribution $p_{\epsilon^q}$ on the quantization error's domain $\mathbb{R}[-Q_p^{-1}/2, Q_p^{-1}/2)$.  
\end{assumption}
Assumption $\ref{assum02}$ leads to the following scalar expectation and variance of the quantization error $\epsilon^q$ trivially as follows:
\begin{equation}
\forall \varepsilon^q \in \displaystyle \R, \quad
\mathbb{E}_{\varepsilon^q} Q_p^{-1} \varepsilon^q = 0, \quad
\mathbb{E}_{\varepsilon^q} Q_p^{-2}{\varepsilon^q}^2 = Q_p^{-2} \cdot \mathbb{E}_{\varepsilon^q} {\varepsilon^q}^2 = 1/12 \cdot Q_p^{-2} = c_0 Q_p^{-2}. 
\label{eq_07}
\end{equation}
%==================================== 
\begin{assumption}[\textbf{WNH from \citet{Jimnez_2007}}]   % assumption 2 : WNH
\label{assum03}
%====================================
When there is a large enough sample and the quantization parameter is sufficiently large, the quantization error is an independent and identically distributed white noise.
\end{assumption}

\textbf{Independent condition of quantization error}
Assumption $\ref{assum03}$ is reasonable when the condition is satisfied for the quantization errors as addressed in \citet{Zamir_1996, Marco_2005, Gray:2006} and \citet{Jimnez_2007}. 
However, the independence between the input signal for quantization and the quantization error is not always fulfilled, so we should check it. 
For instance, if we let a quantization value ${X}^Q = k Q_p^{-1} \in \mathbb{Q}$ for $k \in \mathbb{Z}$, we can evalutae the correlation of ${X}^Q$ and the quantization error $\epsilon^q$ such that 
\begin{equation}
\mathbb{E}_{\varepsilon^q} [X(X^Q - X)\vert X^Q=k Q_p^{-1}] = \mathbb{E}_{\varepsilon^q} [X \epsilon^q \vert X^Q=k Q_p^{-1}] = c_0 Q_p^{-2}.
\end{equation}
Accordingly, if the quantization parameter $Q_p$ is a monotone increasing function to a time index $t$ defined in Definition $\ref{def_q02}$ such that $Q_p^{-1}(t) \downarrow 0, \; t \uparrow \infty$, we can regard the quantization error as the i.i.d. white noise as described in Assumption $\ref{assum03}$, for $t \ge t_0$. 
However, even though the quantization parameter is a time-dependent monotone-increasing function, we cannot ensure the independent condition between an input signal and the quantization error at the initial stage when the time index is less than $t < t_0$.
For instance, since the quantization error is not independent of the input signal for quantization,  the quantization error represents zero when the quantized input is zero, even though the input itself is not zero. 
Such a broken independent condition can cause an early paralysis of learning since there exists nothing to update the state parameter.
On the contrary, when the quantization error is independent of an input signal, we can use the quantization error to optimize the objective function without early paralysis, despite the small valued norm of the input at the initial stage. 

We will present the compensation function to ensure the independent condition and avoid learning paralysis at the early learning stage in the other section.

\section{Learning Equation based on Quantized Optimization }
%%%%%%%%%%%%%%%%%%%%%%%%%%%%%%%%%%%%%%%%%%%%%%%%%%%%%%%%%%%%
%===========================================================
% Figures - 1 for illustration of the proposed algorithm
%===========================================================
\begin{figure}[t]
\centering
    \begin{subfigure}[b]{0.32\textwidth}
    \includegraphics[width=\textwidth]{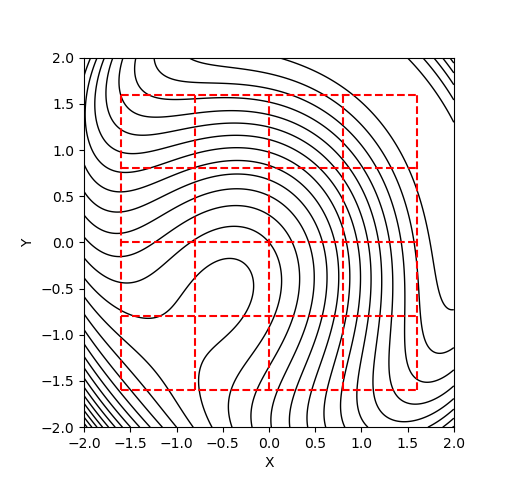}
    \caption{Small valued $Q_p(\tau)$}
    \label{fig01-000}
    \end{subfigure}
    \hfill
    \begin{subfigure}[b]{0.32\textwidth}
    \includegraphics[width=\textwidth]{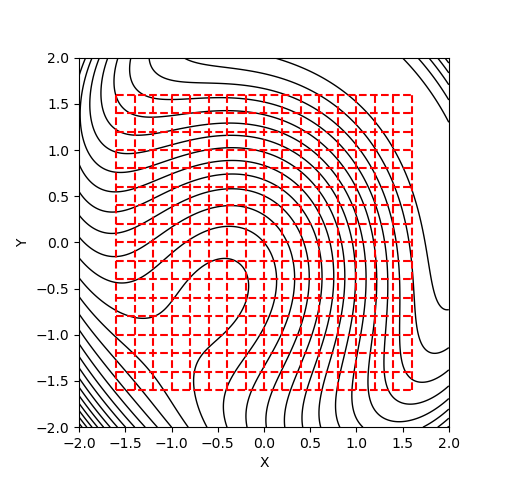}
    \caption{Large valued $Q_p(\tau)$}
    \label{fig01-001}
    \end{subfigure}
    \hfill
    \begin{subfigure}[b]{0.32\textwidth}
    \includegraphics[width=0.9\textwidth]{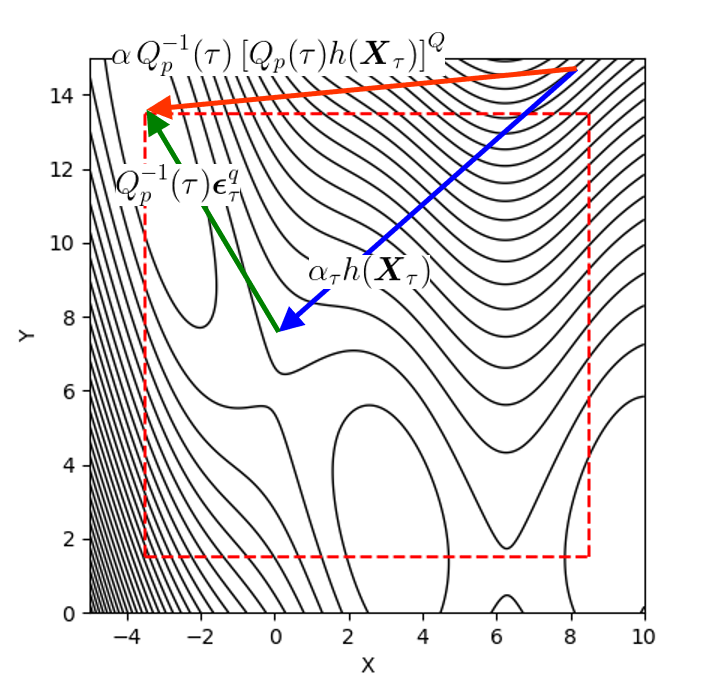}
    \caption{Search vectors}
    \label{fig01-002}
    \end{subfigure}
\caption{The concept diagram of the proposed quantization: (a) In the early stage of learning, with a small value of $Q_p(\tau)$, the search points generated by quantization are widely spaced, indicating a significant quantization error or Brownian motion process affecting the learning process. (b) As the learning process progresses, the quantization parameter $Q_p(\tau)$ increases, resulting in smaller grid sizes. This process resembles an annealing-type stochastic optimization. (c) When considering a general search vector $\alpha, h(\boldsymbol{X}_{\tau})$, quantization introduces an additional search vector, as shown.}
\end{figure}
%===========================================================
Consider that the learning equation given by:
\begin{equation}
\boldsymbol{X}_{\tau + 1} = \boldsymbol{X}_{\tau} + \lambda h(\boldsymbol{X}_{\tau}), \quad \boldsymbol{X}_{\tau} \in \mathbb{R}^d, \; \forall  \tau > 0
\label{ch04-eq01}
\end{equation}
, where $h:\displaystyle \R^d \mapsto \displaystyle \R^d$ represents the search direction, and $\tau$ denotes the time index depending on the index of a mini-batches defined in $\eqref{ch02_eq01}$ and $\eqref{ch02_eq02}$. 
Most artificial intelligence frameworks provide a learning process depending on the unit of the mini-batch size, so equation $\eqref{ch04-eq01}$ describes a real and practical learning process. 

\textbf{Main Idea of the Proposed Quantized Optimization}
The learning equation, represented as equation $\eqref{ch04-eq01}$, searches for a local minimum along a line defined by a directional vector or a conjugate direction when the equation incorporates momentum.
In the proposed quantized optimization, we create a grid on the objective function's domain and sample a point near the feasible point generated by the learning equation.
The grid size is adjustable through the quantization parameter. By considering the quantization error as a white noise following WNH, we can reduce the variance of the quantization error by increasing the quantization parameter's size. This adjustment approximates the dynamics of Langevin SDE in stochastic approximation in the sense of the central limit theorem. 
From an optimization perspective, stochastic gradient-based learning can be considered stochastic sampling.

%====================================
\subsection{Fundamental Learning Equation based on Quantization}
\label{ch04-01}
%====================================
Applying the quantization defined as $\eqref{define_eq01}$ and $\eqref{define_eq01-01}$ to the learning equation $\eqref{ch04-eq01}$, we can obtain the following fundamental quantization-based learning equation.
\begin{equation}
\boldsymbol{X}_{\tau + 1}^Q 
= \boldsymbol{X}_{\tau}^Q + [\lambda h(\boldsymbol{X}_{\tau}^Q)]^Q
= \boldsymbol{X}_{\tau}^Q + Q_p^{-1}(\tau) \left\lfloor Q_p(\tau) \cdot (\lambda h(\boldsymbol{X}_{\tau}^Q) + 0.5 Q_p^{-1})\right\rfloor, \; \boldsymbol{X}_{0}^Q \in \mathbb{Q}^d.
\label{ch04-eq02}
\end{equation}
According to the definition of quantization, we can rewrite $\eqref{ch04-eq02}$ to the following stochastic equation similar to the discrete Langevine equation : 
\begin{equation}
\boldsymbol{X}_{\tau + 1}^Q 
= \boldsymbol{X}_{\tau}^Q + \lambda \, h(\boldsymbol{X}_{\tau}^Q) + \boldsymbol{\epsilon}_{\tau}^q
= \boldsymbol{X}_{\tau}^Q + \lambda \, h(\boldsymbol{X}_{\tau}^Q) + Q_p^{-1}(\boldsymbol{\varepsilon}_{\tau}^q, \tau) \boldsymbol{\varepsilon}_{\tau}^q.
\label{ch04-eq03}
\end{equation}
, where $\boldsymbol{\epsilon}_{\tau}^q \in \mathbb{R}^d$ denotes the vector-valued quantization error, and $\boldsymbol{\varepsilon}_{\tau}^q \in \displaystyle \R^d$ denotes the vector-valued factor for the quantization, defined as $\eqref{define_eq01-01}$ respectively.

Substituting the search direction $h(\boldsymbol{X}_{\tau}^Q)$ with $-\nabla_{\boldsymbol{x}} \tilde{f}_{\tau}(\boldsymbol{X}_{\tau}^Q)$, we rewrite $\eqref{ch04-eq03}$ to the following equation:
\begin{equation}
\boldsymbol{X}_{\tau + 1}^Q = \boldsymbol{X}_{\tau}^Q - \lambda \nabla_{\boldsymbol{x}} \tilde{f}_{\tau}(\boldsymbol{X}_{\tau}^Q) +  Q_p^{-1}(\boldsymbol{\varepsilon}_{\tau}^q, \tau) \boldsymbol{\varepsilon}_{\tau}^q.
\label{ch04-eq04}
\end{equation}
While the fundamental quantized learning equations $\eqref{ch04-eq03}$ and $\eqref{ch04-eq04}$ are the formulae of a representative stochastic difference equation, we cannot analyze the dynamics of these equations as a conventional stochastic equation, owing to the quantization error as the i.i.d. white noise. 
Therefore, to analyze the dynamics of the proposed learning equation from the perspective of stochastic analysis, we suggest the following two alternative approaches:

\textbf{Transformation to Gaussian Wiener Process}
In the previous chapter, we define the fundamental form of the quantization parameter as a function of both the quantization error and the time index such as $Q_p (\boldsymbol{\varepsilon}^q, t)$, represented in $\eqref{def_eq02}$. Using such a multiple-parameterized function, we can transform a uniformly distributed random variable into a standard or approximated Gaussian-distributed random variable. Box-Muller algorithm by \citet{Box_Muller_1958}, Ziggurat algorithm by \citet{Marsaglia_1963, Marsaglia_2000}, and inverse transform sampling from \citet{Thomas_2007} are representative transforms. Since the quantization parameter including the transform can generate a Gaussian distributed independent increments $\boldsymbol{z}_{\tau} \overset{i.i.d.}{\sim} \mathcal{N}(\boldsymbol{z}; 0, \boldsymbol{I}_d)$ such that $\eta(\boldsymbol{\varepsilon}_{\tau}^q) \boldsymbol{\varepsilon}_{\tau}^q = \sqrt{\lambda} \boldsymbol{z}_{\tau}$ under the assumption of which $\lambda = 1/B$, we can rewrite $\eqref{ch04-eq04}$ as follows:
\begin{equation}
\boldsymbol{X}_{\tau + 1}^Q 
= \boldsymbol{X}_{\tau}^Q - \lambda \nabla_{\boldsymbol{x}} \tilde{f}_{\tau}(\boldsymbol{X}_{\tau}^Q) +  \sqrt{\lambda} \cdot b^{-\bar{p}(\tau)} \boldsymbol{z}_{\tau}.
\label{ch04-it-eq01}
\end{equation}
However, even though the transformation offers theoretical benefits derived from a Gaussian-distributed process, there are no advantages to implementing the learning equation, as the property of the quantization error is equivalent to that of a uniformly distributed random variable in a conventional random number generator.
Consequently, we do not treat the transformation-based algorithm in this paper due to the shortage of implementation advantage. 

\textbf{Analysis based on Central Limit Theorem}
Another approach is based on an empirical perspective under the quantization parameter depends only on the time index such that $Q_p(\boldsymbol{\varepsilon}_{\tau}^q, \tau) = Q_p(\tau)$. 
Generally, we check the performance of algorithms at the unit of epoch, not the unit of the unspecified index of a mini-batch. 
Accordingly, if there are sufficient numbers of mini-batches in an epoch and the quantization parameter is constant in a unit epoch such that $Q_p^{-1}(t_e + \tau) = Q_p^{-1}(t_e), \forall \tau \in \mathbb{Z}[0, B)$, we can analyze the summation of the learning equation to each mini-batch index as follows:
\begin{equation}
\boldsymbol{X}_{t_e + 1}^Q 
= \boldsymbol{X}_{t_e}^Q - \lambda \sum_{\tau=0}^{B-1} \nabla_{\boldsymbol{x}} \tilde{f}_{\tau}(\boldsymbol{X}_{t_e+\tau/B}^Q) + b^{-\bar{p}(t_e)} \lambda \sqrt{\frac{C_q}{c_0}}\sum_{\tau=0}^{B-1} \boldsymbol{\varepsilon}_{t_e+\tau/B}^q
\label{ch04-it-eq02}
\end{equation}
, where $Q_p^{-1}(t_e) = \lambda \sqrt{\frac{C_q}{c_0}} b^{-\bar{p}(t_e)}$. 
Herein, the summation of the factor for the quantization error converges to a Gaussian-distributed random variable such that $\sqrt{\frac{\lambda}{c_0}} \sum_{\tau=0}^{B-1} \boldsymbol{\varepsilon}_{t_e+\tau/B}^q \rightarrow \boldsymbol{z}_{t_e} \sim \mathcal{N}(\boldsymbol{z}; 0, \boldsymbol{I}_d)$ as $B \uparrow \infty$, by the central limit theorem. 
Therefore, we can regard the stochastic difference equation $\eqref{ch04-it-eq02}$ as the stochastic integrated equation with respect to $t_e$ to $t_e +1$, so we can obtain an approximated SDE to the epoch-based continuous-time index $t \in \displaystyle \R^+$.
We accept this approach as a quantization-based learning equation since (16) does not require any additional operation such as the transformation to generate a Gaussian random variable. 

\textbf{Appliance to Other Learning Algorithms}
In the quantized stochastic Langevin dynamics (\textbf{QSLD}), the search direction is not fixed as the opposite direction of the gradient vector, allowing for the application of various learning methods such as ADAM (\citet{Kingma_2015}) and alternative versions of ADAM such as ADAMW(\citet{loshchilov_ICLR_2019}), NADAM(\citet{dozat_ICLR_WH_2016}), and RADAM(\citet{Liyuan_ICLR_2020}).

\textbf{Avoid Early Paralysis of the proposed algorithm}
The initial gradient tends to vanish if the initial search point is far from optimal, especially for objective functions in deep neural networks that utilize entropy-based loss functions such as the Kullback-Leibler divergence (KL-Divergence).
The small gradient in the early stage becomes zero after the quantization process, potentially causing the deep neural network (DNN) to fall into a state of paralysis, as illustrated below: 
Assume that $\max \| \lambda h \| < 0.5\, Q_p^{-1}(\tau) - \delta$ for $\tau < \tau_0$, where $\delta$ denotes a positive value such that $\delta Q_p < 1$, and $\tau_0$ denotes a small positive integer. Then, we have 
\begin{equation}
1/Q_p \cdot \| \lfloor Q_p (\lambda h + 0.5\, Q_p^{-1} ) \rfloor \|
\leq 1/Q_p \cdot \lfloor Q_p (\max \| \lambda h \| + 0.5\, Q_p ) \rfloor
= 1/Q_p \cdot \lfloor 1 - \delta Q_p \rfloor
= 0.
\label{avoid-eq01}
\end{equation}
To prevent the paralysis depicted in $\eqref{avoid-eq01}$, a straightforward solution is to re-establish the quantized search direction by incorporating a compensation function $r(\tau) = \lfloor r(\tau) \rfloor$ into $h$, as shown: 
\begin{equation}
\begin{aligned}
&Q_p^{-1}(\tau) \cdot \| \left\lfloor Q_p (\tau) (\lambda h + r(\tau) + 0.5 Q_p^{-1}) \right\rfloor \|_{\max \| \lambda h \| < 0.5\, Q_p^{-1}(\tau) - \delta} \\
&= Q_p^{-1}(\tau) \cdot \| \lfloor Q_p \cdot (\lambda h + 0.5 \, Q_p^{-1} + r(\tau)) \rfloor \|_{\max \| \lambda h \| < 0.5\, Q_p^{-1}(\tau) - \delta} \\
&\leq Q_p^{-1}(\tau) \cdot \| \lfloor Q_p \cdot (\lambda h + 0.5 \, Q_p^{-1} \rfloor \|_{\max \| \lambda h \| < 0.5\, Q_p^{-1}(\tau) - \delta} + \| \lfloor Q_p r(\tau)) \rfloor \| \\
&\leq 0 + Q_p^{-1}(\tau) \cdot \| \lfloor Q_p r(\tau) \rfloor \|. 
\end{aligned}
\label{avoid-eq02}
\end{equation}
In equation $\eqref{avoid-eq02}$, the compensation function is responsible for increasing the magnitude of the search direction during an initial finite period.
To address this, we propose the compensation function $r(\tau)$ given by:
\begin{equation}
r(\tau, \boldsymbol{X}_{\tau}) = \lambda \cdot \left( \frac{\exp(-\varkappa(\tau - \tau_0))}{1 + \exp(-\varkappa(\tau - \tau_0))} \cdot \frac{h(\boldsymbol{X}_{\tau}^Q)}{\| h(\boldsymbol{X}_{\tau}^Q) \|} \right),\quad \tau_0 \in \mathbb{Z}^{++}
\label{avoid-eq03}
\end{equation}
, $\varkappa > 0$ is a determining parameter for the working period, and $\tau_0$ represents the half-time of the compensation. 

Moreover, the compensation function $r(\tau)$ provides the crucial property that the proposed quantization error is uncorrelated to the quantization input, such as the directional derivatives $h$ as follows: 
%====================================
\begin{theorem} % theorem for weak convergence 
\label{theorem_01-02}
%====================================
Let the quantized directional derivatives $h^Q : \mathbb{Z}^+ \times \displaystyle \R^d \mapsto \mathbb{Q}^d$ such that 
\begin{equation}
h^Q (\boldsymbol{X}_{\tau}^Q) \triangleq \frac{1}{Q_p} \left\lfloor Q_p \cdot (\lambda h(\boldsymbol{X}_{\tau}^Q) + r(\tau, \boldsymbol{X}_{\tau}^Q)) + 0.5 \right\rfloor
\end{equation}
, where $r(\tau, \boldsymbol{X}_{\tau})$ denotes a compensation function such that $r:\mathbb{Z}^+ \times \mathbb{R}^d \mapsto \displaystyle \R^d \{-1, 1\}$.
Then, the quantization input $h(\boldsymbol{X}_{\tau}^Q)$ and the quantization error $\boldsymbol{\epsilon}_{\tau}^q$ is uncorrelated such that $\mathbb{E}_{\boldsymbol{\epsilon}_{\tau}^q} [h(\boldsymbol{X}_{\tau}^Q) \boldsymbol{\epsilon}_{\tau}^q \vert h^Q(\boldsymbol{X}_{\tau}^Q) = k Q_p^{-1}] = 0$.
\end{theorem}
Theorem $\ref{theorem_01-02}$ completes the assumptions for the WNH of the quantization error referred to in Assumption $\ref{assum02}$ for stochastic analysis for the proposed learning scheme.

%====================================
\subsection{Convergence Property of QSGLD}
\label{ch04-02}
%====================================
\textbf{Weak Convergence without Convex Assumption}
Before the weak convergence analysis of QSGLD, we establish the following lemma for the approximated SDE.
%====================================
\begin{definition}[\textbf{Order-1 Weak Approximation, \citet{Li-2019-JMLR} and \citet{Malladi_NEURIPS2022}}]
\label{def_04}
%====================================
Let $\{\boldsymbol{X}_t : t\in\mathbf[0, T]\}$ and $\{\boldsymbol{X}_{\tau}^{Q} \}_{\tau=0}^{t_e \, B}$ be families of continuous and discrete stochastic processes parameterized by $\lambda$. We regard $\{\boldsymbol{X}_t\}$ and $\boldsymbol{X}_{\tau}$ are order-1 weak approximations of each other if for all test function $g$ with polynomial growth, there exists a constant $C > 0$ independent of $\lambda$ such that
\begin{equation}
\max_{ \tau \in \mathbf{Z}[0, \lfloor T/\lambda \rfloor]} 
\vert \mathbb{E} g(\boldsymbol{X}_{t}) - \mathbb{E} g(\boldsymbol{X}_{\lfloor \tau/B \rfloor}^Q) \vert \leq C_{o1} \lambda^2.
\end{equation}
\end{definition}
We provide additional definitions required in Definition $\ref{def_04}$, such as the polynomial growth of the test function in the supplementary material.
%====================================
\begin{lemma} % theorem for weak convergence 
\label{lemma_01}
%====================================
The approximated Langvin SDE for QSGLD represented in \eqref{ch04-it-eq02} is as follows: 
\begin{equation}
d\boldsymbol{X}_t = - \nabla_{\boldsymbol{x}} f(\boldsymbol{X}_t) dt + \sqrt{C_q} \cdot \sigma(t) d\boldsymbol{B}_t, \quad \forall t > t_0 \in \displaystyle \R^+, \; \because \sigma(t) \triangleq b^{-\bar{p}(t)},
\label{lemma01_eq01}
\end{equation}
The approximation $\eqref{lemma01_eq01}$ satisfies the order-1 weak approximation described in Definition $\ref{def_04}$.
\begin{comment}
If the quantization parameter is given as $Q_p(t) = \frac{1}{\lambda}\, \sqrt{\frac{c_0}{C_q}} \, b^{\bar{p}(t)}$, the QSGLD represented as \eqref{ch04-it-eq02} weakly converges to Langevin SDE with respect to $\boldsymbol{X}_t \in \mathbb{R}^d$ as follows:
\begin{equation}
d\boldsymbol{X}_t = - \nabla_{\boldsymbol{x}} f(\boldsymbol{X}_t) dt + \sqrt{C_q} \cdot \sigma(t) d\boldsymbol{B}_t, \quad \forall t > t_0 \in \displaystyle \R^+, \; \because \sigma(t) \triangleq b^{-\bar{p}(t)},
\label{lemma01_eq01}
\end{equation}
, where $\boldsymbol{B}_t$ denotes a standard vector valued Wiener process, $C_q \in \displaystyle \R^+$ denotes a constant value, and $\lambda$ denotes the reciprocal of the size of mini-batch $B$, 
\end{comment}
\end{lemma}

\textbf{Sketch of proof}
\citet{Li-2019-JMLR} introduced a rigorous analytical framework for approximating stochastic difference equations with stochastic differential equations (SDEs). Building upon this framework, \citet{Malladi_NEURIPS2022} extended the analysis for more general search directions.
We leverage the aforementioned framework to establish Lemma $\ref{lemma_01}$. We impose several bounded assumptions described in the supplementary material to derive the moment of the one-step difference in $\eqref{lemma01_eq01}$.
Subsequently, by utilizing the bound of the moment, we prove the order-1 weak approximation of QSGLD to the Langevin SDE.

\begin{comment}
We already obtained the stochastic difference equation (16), which is the inductive summation of the proposed quantization-based learning equation. 
In \eqref{lemma01_eq01}, we verify the stochastic process consisted with the summation of the quantization error $\{Q_p^{-1}(t) \sum_{\tau=0}^{B-1} \boldsymbol{\varepsilon}_{t+\tau}^q \}_{t>t_0}$ converges to a standard Wiener process, based on the central limit theorem. 
Finally, we show that \eqref{ch04-it-eq02} weakly converges to the SDE \eqref{lemma01_eq01} from the analysis framework provided by \citet{Kushner-1974}.
\end{comment}

%====================================
\begin{theorem} % theorem for weak convergence 
\label{theorem_02}
%====================================
Consider the transition probability, denoted as $p(t, \boldsymbol{X}_t, t+\bar{\tau}, \boldsymbol{x}^*)$, from an arbitrary state $\boldsymbol{X}_{t} \in \mathbb{R}^d$ to the optimal point $\boldsymbol{x}^* \in \mathbb{R}^d, \; \boldsymbol{X}_t \neq \boldsymbol{x}^*$ after a time interval $\bar{\tau} \in \displaystyle \R^+$, for all $t > t_0$. If the quantization parameter is bounded as follows:
\begin{equation}
\sup_{t \geq 0} Q_p(t) = \sqrt{\frac{1}{C} \cdot{\log (t + 2)}}, \quad C \in \displaystyle \R^{++},
\end{equation}
for $\boldsymbol{X}_t \neq \boldsymbol{\bar{X}}_t$, QSGLD represented as $\eqref{lemma01_eq01}$ converges with distribution in the sense of Cauchy convergence such that
\begin{equation}
\overline{\lim_{\bar{\tau} \rightarrow \infty}} \sup_{\boldsymbol{X}_t, \bar{\boldsymbol{X}}_{t} \in \displaystyle \R^n} \| p(t, \boldsymbol{\bar{X}}_t, t + \bar{\tau},  \boldsymbol{x}^*) - p(t, \boldsymbol{X}_t, t + \bar{\tau},  \boldsymbol{x}^*) \| \leq \tilde{C} \cdot \exp \left(-\sum_{\bar{\tau}=0}^{\infty} \delta_{t+\bar{\tau}} \right)
\end{equation}
, where $\delta_{t}$ denotes the infimum of the transition probability from time $t$ to $t+1$ given by $\delta_t = \inf_{\boldsymbol{x}, \boldsymbol{y} \in \displaystyle \R^d} p(t, \boldsymbol{x}, t+1, \boldsymbol{y})$, satisfying $\sum_{\bar{\tau}=0}^{\infty} \delta_{t+\bar{\tau}} = \infty$, and $\tilde{C}$ denotes a positive value. 
\end{theorem}
\textbf{Sketch of proof}
First, we analyze the limit supremum of the difference between the transition probabilities expressed by the infimum $\delta_t$.
Next, according to the Girsanov theorem (\citet{Bernt_2003} and \citet{Klebaner_2011}), we calculate the Radon-Nykodym derivative of the probability measure derived by the weak solution of Langevin SDE relevant to QSGLD with respect to the probability of a standard Gaussian.
Using the obtained Radon-Nykodym derivative, we calculate a lower bound for $\delta_t$ and provide proof for the theorem.

\textbf{Local Convergence under Convex Assumption}
For local convergence analysis, we suppose that the objective function around the optimal point is strictly convex. 
%====================================
\begin{assumption}  %theorem for local convergence 
\label{assum_04}
%====================================
The Hessian of the objective function $\boldsymbol{H}(f): \mathbb{R}^d \mapsto \mathbb{R}^d$ around the optimal point is non-singular and positive definite,  
\end{assumption}

%====================================
\begin{theorem} % theorem for local convergence 
\label{theorem_03}
%====================================
The expectation value of the objective function derived by the proposed QSGLD converges to a locally optimal point asymptotically under Assumption $\ref{assum_04}$. 
\end{theorem}
\textbf{Sketch of proof}
Intuitively, the proof of the theorem follows a similar structure to the conventional proof of gradient descent under the convex assumption. 
However, with the existence of the Brownian motion process, we apply the stationary probability from Theorem $\ref{theorem_02}$ into the proof. 
Accordingly, we prove the convergence not on the point-wise vector space, but on the function space induced by a stationary expectation.
%===========================================================
%\begin{comment}
\begin{algorithm}[tb]
    \caption{QSLD/QSGLD with the proposed quantization scheme}\label{alg-1}
\begin{algorithmic}[1]
\State{\bfseries Initialization} $\tau \leftarrow 0$, $\boldsymbol{X}_0 \in \mathbb{Q}^d$ \Comment {Set Initialize Discrete Time Index and state}
\Repeat
    \State Compute $h (\boldsymbol{X}_{\tau}^Q)$ at $\tau$   \Comment {Compute a Search Direction Vector}
    \State Compute $Q_p(\tau)$, and $r(\tau, h (\boldsymbol{X}_{\tau}^Q))$   \Comment {Compute Quantization Parameter and $r(\tau)$}
    %\State $\bar{h}_{\tau} \leftarrow -\alpha \nabla f(\boldsymbol{X}_{\tau}) + r(\tau)$ \Comment {Compute the Directional vector with $r(\tau)$}
    \State $h_{\tau}^Q \leftarrow \frac{1}{Q_p} \lfloor Q_p \cdot (-\lambda h(\boldsymbol{X}_{\tau}^Q) + r(\tau, h (\boldsymbol{X}_{\tau}^Q)) + 0.5 Q_p^{-1}) \rfloor$ \Comment {Quantization of Search Vector}
    \State $\boldsymbol{X}_{\tau}^Q \leftarrow \boldsymbol{X}_{\tau} + h_{\tau}^Q$ \Comment {General Updating Rule for Learning}
    \State $\tau \leftarrow \tau+1$ \Comment {General Update Discrete Time Index}
\Until {Stopping criterion is met}    
\end{algorithmic}
\end{algorithm}
%\end{comment}
%===========================================================
\begin{comment}
\begin{minipage}[t]{.45\textwidth}
\begin{algorithm}[H]
    \centering
    \caption{QSGLD with the proposed quantization scheme}\label{alg-1}
\begin{algorithmic}[1]
\State{\bfseries Initialization} $\tau \leftarrow 0$ 
\Repeat
    \State Compute $Q_p(\tau)$, and $r(\tau)$
    \State $h_{\tau} \leftarrow -\lambda \nabla f(\boldsymbol{X}_{\tau}) - r(\tau)$ 
    \State $h_{\tau}^Q \leftarrow \frac{1}{Q_p}[Q_p \cdot h_{\tau}]$ 
    \State $\boldsymbol{X}_{\tau} \leftarrow \boldsymbol{X}_{\tau} + h_{\tau}^Q$ 
    \State $\tau \leftarrow \tau+1$
\Until {Stopping criterion is met}    
\end{algorithmic}
\end{algorithm}
\end{minipage}
\begin{minipage}[t]{.45\textwidth}
\begin{algorithm}[H]
    \caption{Compute $Q_p(\tau)$ and $r(\tau)$}\label{alg-2}
\begin{algorithmic}[1]
\State{\bfseries Initialization} Set $\eta, a, b, \varkappa, C, C_l, C_r, \tau_0$ 
\State $r(\tau) \leftarrow C_rQ_p^{-1} \frac{\exp(-\varkappa(\tau - \tau_0))}{1 + \exp(-\varkappa(\tau - \tau_0))} \cdot \frac{h(\boldsymbol{X}_t)}{\| h(\boldsymbol{X}_t) \|}$
\State $h_{\tau} \leftarrow -\lambda \nabla f(\boldsymbol{X}_{\tau}) - r(\tau)$ 
\State $h_{\tau}^Q \leftarrow \frac{1}{Q_p}[Q_p \cdot h_{\tau}]$ 
\State $\boldsymbol{X}_{\tau} \leftarrow \boldsymbol{X}_{\tau} + h_{\tau}^Q$ 
\State $\tau \leftarrow \tau+1$
\end{algorithmic}
\end{algorithm}
\end{minipage}
\end{comment}
%%%%%%%%%%%%%%%%%%%%%%%%%%%%%%%%%%%%%%%%%%%%%%%%%%%%%%%%%%%%
\section{Experimental Results}
%%%%%%%%%%%%%%%%%%%%%%%%%%%%%%%%%%%%%%%%%%%%%%%%%%%%%%%%%%%%
\begin{table}[]
\caption{Comparison of test performance among optimizers with a fixed learning rate 0.01. Evaluation is based on the Top-1 accuracy of the training and testing data.}
\centering
{
\scriptsize
\begin{tabular}{l|ccc|cccccc|}
\hline
Data Set   & \multicolumn{3}{c|}{FashionMNIST}      & \multicolumn{3}{c|}{CIFAR10}                             & \multicolumn{3}{c|}{CIFAR100}       \\ \hline
Model      & \multicolumn{3}{c|}{CNN with 8-Lsyers} & \multicolumn{6}{c|}{ResNet-50}                                                                 \\ \hline
Algorithms & Training  & Testing  & Training Error  & Training & Testing & \multicolumn{1}{c|}{Training Error} & Training & Testing & Training Error \\ \hline
QSGD       & 97.10     & 91.59    & 0.085426        & 99.90    & 73.80   & \multicolumn{1}{c|}{0.009253}       & 99.04    & 37.77   & 0.030104       \\
QADAM      & 98.43     & 89.29    & 0.059952        & 99.99    & 85.09   & \multicolumn{1}{c|}{0.011456}       & 98.62    & 49.60   & 0.037855       \\
SGD        & 95.59     & 91.47    & 0.132747        & 99.99    & 63.31   & \multicolumn{1}{c|}{0.001042}       & 98.24    & 25.90   & 0.005478       \\
ASGD       & 95.60     & 91.42    & 0.130992        & 99.99    & 63.46   & \multicolumn{1}{c|}{0.001166}       & 98.36    & 26.43   & 0.004981       \\
ADAM       & 92.45     & 87.12    & 0.176379        & 99.75    & 82.08   & \multicolumn{1}{c|}{0.012421}       & 98.85    & 46.32   & 0.038741       \\
ADAMW      & 91.72     & 86.81    & 0.182867        & 99.57    & 82.20   & \multicolumn{1}{c|}{0.012551}       & 98.86    & 47.01   & 0.038002       \\
NADAM      & 96.25     & 87.55    & 0.140066        & 99.56    & 82.46   & \multicolumn{1}{c|}{0.014377}       & 98.62    & 48.56   & 0.037409       \\
RADAM      & 95.03     & 87.75    & 0.146404        & 99.65    & 82.26   & \multicolumn{1}{c|}{0.010526}       & 98.17    & 48.61   & 0.044193       \\ \hline
\end{tabular}
}
\end{table}

%===========================================================
\subsection{Configuration of Experiments}
From a practical point of view, as both the updated state $\boldsymbol{X}_{\tau+1}^Q$ and the current state $\boldsymbol{X}{\tau}^Q$ are quantized vectors, the quantization parameter should be a power of the base defined in equation $\eqref{def_eq02}$.
Although Theorem $\ref{theorem_02}$ provides an upper bound for the quantization parameter as a real value, we constrain the quantization parameter to be a rational number within a bounded range of real values. Specifically, we set the quantization parameter $Q_p$ as follows:
\begin{equation}
Q_p = \left \lfloor \sqrt{\frac{1}{C} \log(t_e + 2)} \right\rfloor
\end{equation}
, where $t_e$ represents a unit epoch defined as $t_e = \lfloor \tau/B \rfloor$, and $\tau$ denotes a unit update index such that $\tau = t_e \cdot B + k, \; k \in \mathbb{Z}^+[0, B)$. In the supplementary material, we provide detailed explanations and methods for calculating and establishing the remaining hyper-parameters.
\begin{comment}
The following theorem gives one example for the bound when we set a lower bound of the quantization parameter.
%====================================
\begin{theorem}
\label{theorem_04}
%====================================
Let $L_{Q_p}$ be the lower bound of the quantization parameter defined as $L_{Q_p} = \frac{1}{C_l} \log (\tau + 2)$, where $C_l$ is a positive value satisfying $C_l > C$. Consequently, we derive the following bound for $\bar{p}$:
\begin{equation}
\hat{g}(\tau) + \log_b\frac{C}{C_l}< \bar{p} < \hat{g}(\tau), 
\quad \because \hat{g}(\tau) \triangleq \log_b \left( \frac{1}{\eta C} \log(\tau + 2) \right)
\label{th04_eq01}
\end{equation}
\end{theorem}
\textbf{Sketch of proof}
We can derive $\eqref{th04_eq01}$ through simple calculation from the assumptions in Theorem $\ref{theorem_04}$ and equation $\eqref{def_eq02}$.
\end{comment}

\subsection{Brief Information of Experiments}
We conducted experiments to compare QSGLD with standard SGD and ASGD (\cite{Shamir_ICML_2013}).
Additionally, we compared the performance of QSLD to the proposed method with ADAM, ADAMW, NADAM, and RADAM in terms of convergence speed and generalization.
These results provide an empirical analysis and demonstrate the effectiveness of the proposed algorithms.
The network models used are ResNet-50 and a small-sized CNN network with 3-layer blocks, of which two former blocks are a bottle-neck network, and one block is a fully connected network for classification. 
Data sets used for the experiments are FashionMNIST(\citet{xiao2017fashionmnist}), CIFAR-10, and CIFAR-100. 
We give detailed information about the experiments, such as hyper-parameters of each algorithm, in the supplementary material. 

\subsection{Empirical Analysis of the Proposed Algorithm}
\textbf{Test for FashionMNIST}
For the experiment, we employed a small-sized CNN to evaluate the performance of the test algorithms. 
The FashionMNIST data-set consists of black-and-white images that are similar to MNIST data. 
Figure 2 illustrates that SGD algorithms demonstrate superior generalization performance on the test data, even though their convergence speed is slower than ADAM-based algorithms.
Table 1 shows that the proposed quantization algorithms achieve a superior final accuracy error. 
Notably, the quantization algorithm applied to ADAM algorithms surpasses the error bound and achieves even lower accuracy error.
As a result, the proposed quantization algorithm demonstrates improved classification performance for test data.

\textbf{Test for CIFAR-10 and CIFAR-100}
For the CIFAR-10 and CIFAR-100 tests, we utilized ResNet-50 as the model architecture.
As shown in Table 1, when classifying the CIFAR-10 data-set using ResNet-50, QSGLD outperformed SGD by 8\% in terms of test accuracy. As depicted in Figure 2, QSGLD exhibited significant improvements in convergence speed and error reduction compared to conventional SGD methods. On the other hand, Adam-Based QSLD showed a performance advantage of approximately 3\% for test accuracy for CIFAR-10 data-set. Similar trends were observed for the CIFAR-100 data-set. QSGLD demonstrated a performance advantage of around 11\% over conventional SGD methods for test accuracy. In contrast, Adam-Based QSLD showed an improvement of approximately 3.0\% compared to Adam optimizers and 1.0\% compared to NAdam and RAdam.

%%%%%%%%%%%%%%%%%%%%%%%%%%%%%%%%%%%%%%%%%%%%%%%%%%%%%%%%%%%%
\section{Conclusion}
%%%%%%%%%%%%%%%%%%%%%%%%%%%%%%%%%%%%%%%%%%%%%%%%%%%%%%%%%%%%
%===========================================================
% Figures - 2 for illustration of the proposed algorithm
%===========================================================
\begin{figure}[t]
\centering
    \begin{subfigure}[b]{0.32\textwidth}
    \includegraphics[width=\textwidth]{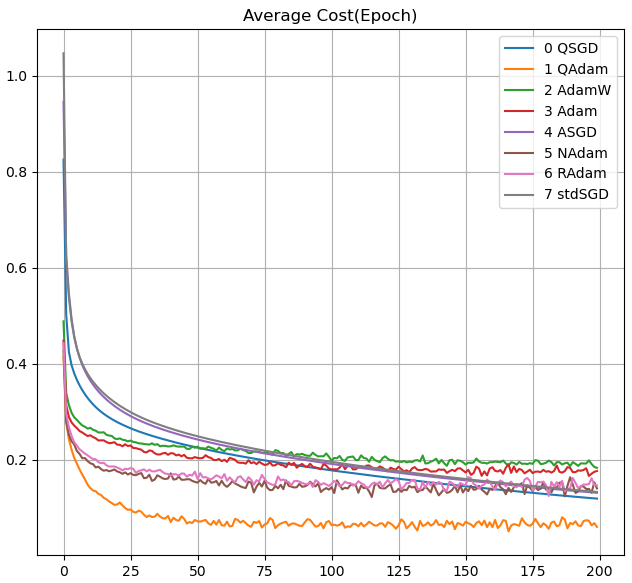}
    \caption{FashionMNIST via CNN}
    \label{fig02-000}
    \end{subfigure}
    \hfill
    \begin{subfigure}[b]{0.32\textwidth}
    \includegraphics[width=\textwidth]{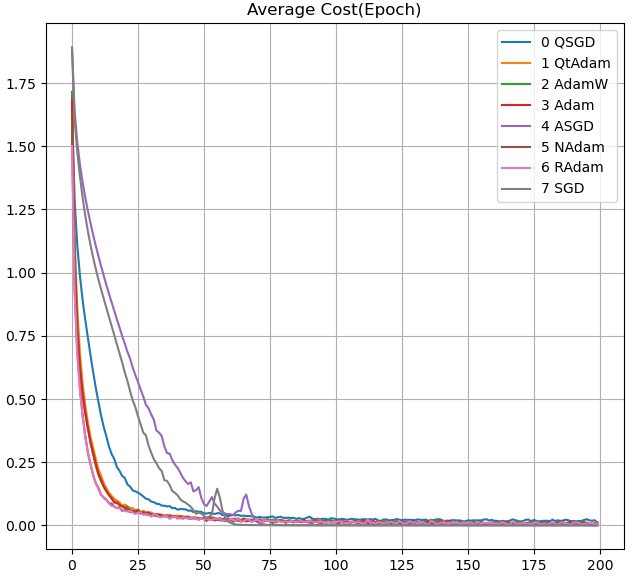}
    \caption{CIFAR10 via ResNet50}
    \label{fig02-001}
    \end{subfigure}
    \hfill
    \begin{subfigure}[b]{0.32\textwidth}
    \includegraphics[width=\textwidth]{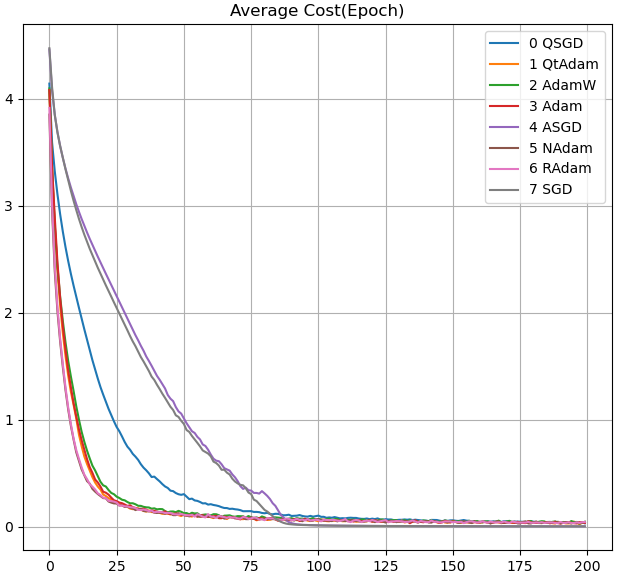}
    \caption{CIFAR100 via ResNet50}
    \label{fig02-002}
    \end{subfigure}
\caption{The error trends of test algorithms to data-set and neural models: (a) Training error trends of the CNN model on FashionMNIST data-set.(b) Training error trends of ResNet-50 on CIFAR-10 data-set. (c) Training error trends of ResNet-50 on CIFAR-100 data-set.}
\end{figure}
%===========================================================
We introduce two stochastic descent learning equations, QSLD and QSGLD, based on quantized optimization. Empirical results demonstrate that the error trends of the quantization-based stochastic learning equations exhibit stability compared to other noise-injecting algorithms. 
We provide the proposed algorithm's weak and local convergence properties with the Langevin SDE perspective.
In our view, exploring the potential of quantization techniques applied to the range of objective functions may offer limited performance gains. Hence, future research will investigate the quantization technique in the context of objective function domains.
Furthermore, we have not conducted an analysis of a more generalized version of the proposed method, QSLD, despite the empirical results suggesting its satisfactory consistency. Analyzing QSLD requires studying the properties of the general search direction, which we reserve for future investigations.

\begin{comment}
%%%%%%%%%%%%%%%%%%%%%%%%%%%%%%%%%%%%%%%%%%%%%%%%%%%%%%%%%%%%%%%%%%%%%%%
\section{Acknowledgement}
%%%%%%%%%%%%%%%%%%%%%%%%%%%%%%%%%%%%%%%%%%%%%%%%%%%%%%%%%%%%%%%%%%%%%%%
This work was supported by Institute for Information and Communications Technology Promotion(IITP) grant funded by the Korea government(MSIP) (2021-0- 00766, Development of Integrated Development Framework that supports Automatic Neural Network Generation and Deployment optimized for Runtime Environment)
\end{comment}
%%%%%%%%%%%%%%%%%%%%%%%%%%%%%%%%%%%%%%%%%%%%%%%%%%%%%%%%%%%%
\newpage
%\section*{References}
\bibliography{main}
\newpage
\appendix
\input{main_appendix}

\end{document}

%% file: math_commands.tex
\usepackage{amsmath,amsfonts,bm}

\def\eqref#1{equation~\ref{#1}}
\def\1{\bm{1}}

\DeclareMathAlphabet{\mathsfit}{\encodingdefault}{\sfdefault}{m}{sl}
\SetMathAlphabet{\mathsfit}{bold}{\encodingdefault}{\sfdefault}{bx}{n}

\newcommand{\R}{\mathbb{R}}

%% file: main_appendix.tex
\title{Supplementary Material for the  Manuscript No. 4810: "Stochastic Gradient Langevin Dynamics Based on Quantization with Increasing Resolution" }

\newtheorem{theorem-ack}{Theorem}
\newtheorem{lemma-ack}[theorem-ack]{Lemma}
\newtheorem{lemma-aux}{Lemma : Auxiliary}

\newpage
%%%%%%%%%%%%%%%%%%%%%%%%%%%%%%%%%%%%%%%%%%%%%%%%%%%%%%%%%%%%%%%%%%%%%%%
\section{Appendix : Introduction}
%%%%%%%%%%%%%%%%%%%%%%%%%%%%%%%%%%%%%%%%%%%%%%%%%%%%%%%%%%%%%%%%%%%%%%%
We provide the mathematical properties and the proofs for the theorems and lemmas featured in this manuscript. 
Firstly, we present the statistical properties of the quantization error, such as expectation, variance, and independent properties.
Moreover, we present the proof of Theorem $\ref{theorem_01-02}$ to provide the independence of the quantization error to the directional derivation $h$ for the early stage of learning.

Next, we provide the discrete variation of the proposed quantized optimization-based learning equation for stochastic analysis.
Such discrete learning formulae show how can we transform the quantization error into a standard Wiener process with the auxiliary function in the quantization parameter.

Based on the provided discrete equation, we prove that the proposed learning equation weakly converges to the stochastic differential equation known as the Langevin equation.
Using the weakly converged stochastic differential equation, we give the proof of Theorem \ref{theorem_02} about the weak convergence of the proposed algorithm.
Through the property of weak convergence, we posit that the proposed algorithm possesses a stationary distribution of the transition probability towards an optimal point, even in the presence of non-convex objective functions.

Additionally, We establish the proposed algorithm's asymptotic convergence under a convex assumption.  Despite the non-convexity assumptions, we provide proof of local convergence by considering the feasibility of assuming strict local convexity around an optimal point.

Furthermore, we provide detailed information regarding the experiments, including comprehensive results of simulations for each data set, network models utilized, and the corresponding hyperparameters for each algorithm. Finally, we offer further discussion and analysis of the specific empirical findings in these experiments.

%=====================================================
\subsection{Notations}
%=====================================================
Generally, we follow the mathematical notations shown in the manuscript provided by the ICLR formatting instructions. 
Herein, we provide some combinations of the notations for easy reading and particular expressions used in the manuscript. 

\begin{itemize}
    \item $\mathbb{R}^n$  ~ The n-dimensional space with real numbers
    \item $\mathbb{R}$    ~ $\mathbb{R}^n \vert_{n=1}$
    \item $\mathbb{R}[\alpha, \beta]$ ~ $\{ x \in \mathbb{R} | \alpha \leq  x \leq \beta, \; \alpha, \beta \in \mathbb{R} \}$
    \item $\mathbb{R}(\alpha, \beta]$ ~ $\{ x \in \mathbb{R} | \alpha <  x \leq \beta, \; \alpha, \beta \in \mathbb{R} \}$
    \item $\mathbb{R}[\alpha, \beta)$ ~ $\{ x \in \mathbb{R} | \alpha \leq  x < \beta, \; \alpha, \beta \in \mathbb{R} \}$
    \item $\mathbb{R}(\alpha, \beta)$ ~ $\{ x \in \mathbb{R} | \alpha <  x < \beta, \; \alpha, \beta \in \mathbb{R} \}$
    \item $\mathbb{Q}^n$  ~ The n-dimensional space with rational numbers
    \item $\mathbb{Q}$    ~ $\mathbb{Q}^n \vert_{n=1}$
    \item $\mathbb{Z}$    ~ The 1-dimensional space with integers. 
    \item $\mathbb{N}$    ~ The 1-dimensional space with natural numbers.  
    \item $\mathbb{R}^+$  ~ $\{ x \vert x \geq 0, \;  x \in \mathbb{R}\}$
    \item $\mathbb{R}^{++}$ ~ $\{ x \vert x > 0, \;  x \in \mathbb{R}\}$
    \item $\mathbb{Q}^+$  ~ $\{ x \vert x \geq 0, \;  x \in \mathbb{Q}\}$
    \item $\mathbb{Q}^{++}$ ~ $\{ x \vert x > 0, \;  x \in \mathbb{Q}\}$
    \item $\mathbb{Z}^+$  ~ $\{ x \vert x \geq 0, \;  x \in \mathbb{Z}\}$
    \item $\mathbb{Z}^{++}$ ~ $\{ x \vert x > 0, \;  x \in \mathbb{Z}\}$,  $\mathbb{Z}^{++}$ is equal to $\mathbb{N}$.
    \item $\lfloor x \rfloor$ ~ $\max \{ y \in \mathbb{Z}| y \leq x, \forall x \in \mathbb{R} \}$
    \item $\lceil x \rceil$ ~ $\min \{ y \in \mathbb{Z}| y \geq x, \forall x \in \mathbb{R} \}$    
    \item $n_{B_{\tau}}$ ~ The numbers of samples in the $\tau$-th mini-batch $B_{\tau}$.
    \item $B$ ~ The total number of mini-batches. If we have totally $N_T$ numbers of samples for a dataset, we have $N_T = \sum_{\tau=1}^{B} n_{B_{\tau}}$.   
    \item $t \in \mathbb{R}^+$ ~ Time index for stochastic analysis on a continuous time
    \item $\tau \in \mathbb{Z}^+$ ~ Discrete time index. Typically it denotes the index of mini-batches 
    \item $t_e \in \mathbb{Z}^+$ ~ Discrete time index. Typically it denotes the epoch, corresponding to $\tau$ such that $t_e = \tau /B$
    \item $P_{X}(x)$ ~ Probability density function of the discrete random variable $X$ with respect to a domain $\mathcal{D}(x)$ containing $x$. 
    \item $p_{X}(x)$ ~ Probability density function of the continuous random variable $X$ with respect to a domain $\mathcal{D}(x)$ containing $x$. 
    \item $\mathbb{E}_{X}(X)$ ~ Expectation value of the random variable $X$ following the distribution $P_{X}(x)$ such that $\mathbb{E}_{X}(X) = \int_{\mathcal{D}(x)} x P_{X}(x) dx$, where $\mathcal{D}(x)$ is a domain containing $x$. More detailed, $\mathbb{E}_{X}(Y(x))$ is equivalent to $\mathbb{E}(Y(x) \vert X) \triangleq \int_{-\infty}^{\infty} y(x) P_{X}(x) dx$     
    Accordingly, $\mathbb{E}_{x \sim P_{X}}$ is an equivalent expression. 
    \item $\langle \boldsymbol{a}, \boldsymbol{b} \rangle$ ~ The inner product of the vector $\boldsymbol{a} \in \mathbb{R}^d$ and $\boldsymbol{b} \in \mathbb{R}^d$. This notation is equivalent to $\boldsymbol{a} \cdot \boldsymbol{b}$.
    \item $\mathcal{U}(\boldsymbol{x}; \mu, \sigma )$ ~ Uniform distribution with the expectation $\mu$, the variance $\sigma$. The random variable $\boldsymbol{x}$ follows the uniform distribution. Additionally, we employ a notation without the random variable such as $\mathcal{U}(\mu, \sigma )$.
    \item $\mathcal{N}(\boldsymbol{x}; \mu, \sigma )$ ~ Normal distribution with the expectation $\mu$, the variance $\sigma$. The random variable $\boldsymbol{x}$ follows the normal distribution. Additionally, we employ a notation without the random variable such as $\mathcal{N}(\mu, \sigma )$.
\end{itemize}

%=====================================================
\subsection{Additional Definitions and Assumptions}
%=====================================================
We refer to the definitions and assumptions described in the main manuscript to all theorems and lemmas in the supplementary material.
However, we didn't mention the following definition for a measure of a matrix, so that we define it as follows: 
%==================================== 
\begin{definition}  % definition 1 {Matrix Norm}
\label{def_st01}
%====================================
For $\boldsymbol{A} \in \displaystyle \R^{m \times m}, m \in \mathbb{N}$, we define a metric for a matrix $A$ as follows:
\begin{equation}
\left( \boldsymbol{A} \right)_{\displaystyle \R^{m \times m}} \triangleq \langle \boldsymbol{v}, \boldsymbol{Av} \rangle , \quad \boldsymbol{v} \in \displaystyle \R^m, \; \| \boldsymbol{v} \| = 1.
\label{as01_eq03}    
\end{equation}
We define the matrix norm of $\boldsymbol{A}$, denoted as $\| \boldsymbol{A} \|_{\displaystyle \R^{m \times m}}$, using the absolute value of the metric in \eqref{as01_eq03} such that $\| \boldsymbol{A} \|_{\displaystyle \R^{m \times m}} \triangleq |(\boldsymbol{A})_{\displaystyle \R^{m \times m}}|$.
\end{definition}

%%%%%%%%%%%%%%%%%%%%%%%%%%%%%%%%%%%%%%%%%%%%%%%%%%%%%%%%%%%%%%%%%%%%%%%
\section{Statistical Properties of Quantization} % Theorem 1
%%%%%%%%%%%%%%%%%%%%%%%%%%%%%%%%%%%%%%%%%%%%%%%%%%%%%%%%%%%%%%%%%%%%%%%
%=====================================================
\subsection{Proof of Basic Statistical Properties}
%=====================================================
\textbf{Fundamental Statistical Properties of the Quantization error}

From Definition \ref{def_q01} and Assumption \ref{assum02}, we note that $\epsilon^q = Q_p^{-1} \varepsilon^q$ and $p_{{\epsilon}^q}(x) = Q_p$ for $\mathbb{R}[-\frac{1}{2 Q_p}, \frac{1}{2 Q_p})$, so that we can obtain the differential $d\epsilon^q = Q_p^{-1} d\varepsilon^q$ and the
integral domain for $\varepsilon^q$ such that $\mathcal{D}(\varepsilon^q) = [-\frac{1}{2}, \frac{1}{2})$. 
It implies that 
\begin{equation}
\mathbb{E}_{\epsilon^q} \epsilon^q 
= \mathbb{E}_{\epsilon^q} Q_p^{-1} \varepsilon^q 
= \int_{-\frac{1}{2 Q_p}}^{\frac{1}{2 Q_p}} p_{{\epsilon}^q}(x) {\epsilon^q} d{\epsilon^q} 
= \int_{-\frac{1}{2}}^{\frac{1}{2}} Q_p Q_p^{-1} \varepsilon^q Q_p^{-1} d{\varepsilon^q} 
= \frac{Q_p^{-1}}{2} {\varepsilon^q}^2 \Bigg\vert_{-\frac{1}{2}}^{\frac{1}{2}}
= 0.
\label{ack-ch02-eq01}
\end{equation}
Therefore, we simply evaluate the expectation value of the quantized input $X^Q \in \mathbb{Q}$ such that 
\begin{equation}
\begin{aligned}
\mathbb{E}_{\epsilon^q}[X^Q \vert X=x] 
&= \mathbb{E}_{\epsilon^q}[X^Q \vert X=x] \\
&= \int_{-\frac{\Delta}{2}}^{\frac{\Delta}{2}} p_{\epsilon^q}(\epsilon^q) \Delta \left\lfloor \frac{x}{\Delta} + \frac{1}{2} \right\rfloor d\epsilon^q 
= \int_{-\frac{1}{2 Q_p}}^{\frac{1}{2 Q_p}} Q_p \cdot (x + \varepsilon^q Q_p^{-1}) d\epsilon^q
= \int_{-\frac{1}{2 Q_p}}^{\frac{1}{2 Q_p}} Q_p \cdot (x + \epsilon^q) d\epsilon^q \\
&= Q_p \left( \int_{-\frac{1}{2 Q_p}}^{\frac{1}{2 Q_p}} x d\epsilon^q + \int_{-\frac{1}{2 Q_p}}^{\frac{1}{2 Q_p}} \epsilon^q d \epsilon^q \right) 
= Q_p \left( x \int_{-\frac{1}{2 Q_p}}^{\frac{1}{2 Q_p}} d\epsilon^q + \frac{{\epsilon^q}^2}{2} \Bigg\vert_{-\frac{1}{2 Q_p}}^{\frac{1}{2 Q_p}} \right) \\
&= Q_p \left( x \frac{1}{Q_p} + \frac{1}{2} \left( \frac{1}{4 Q_p^2} - \frac{1}{ 4 Q_p^2} \right) \right) \\
&= x.
\end{aligned}
\label{ack-ch02-eq01-01}
\end{equation}

We can rewrite \eqref{ack-ch02-eq01-01} to the following simplified formulae:
\begin{equation}
\mathbb{E}_{\epsilon^q}[{X}^Q \vert {X}=x] 
= \mathbb{E}_{\epsilon^q}[{X} + \epsilon^q \vert {X}=x] 
= \mathbb{E}_{\epsilon^q}[x + \epsilon^q] 
= x + \mathbb{E}_{\epsilon^q} \epsilon^q = x.
\label{ack-ch02-eq02}
\end{equation}
, where $x \in \mathbb{R}$ is a scalar value and $X \in \mathbb{R}$ is a scalar random variable as an input of quantization.
Further, we can obtain the variance of the quantization error as follows:
\begin{equation}
\begin{aligned}
Var_{\epsilon^q} \epsilon^q 
&= \mathbb{E}_{\epsilon^q} (\epsilon^q)^2 - \left( \mathbb{E}_{\epsilon^q} \epsilon^q \right)^2
= \mathbb{E}_{\epsilon^q} (\epsilon^q)^2
= \mathbb{E}_{\epsilon^q} Q_p^{-2} {\varepsilon^q}^2 \\
&= \int_{-\frac{1}{2}}^{\frac{1}{2}} Q_p Q_p^{-2} {\varepsilon^q}^2 Q_p^{-1} d{\varepsilon^q} 
= \frac{Q_p^{-2}}{3} {\varepsilon^q}^3 \Bigg\vert_{-\frac{1}{2}}^{\frac{1}{2}}
= \frac{1}{12 Q_p^2} = c_0 Q_p^{-2}.
\end{aligned}
\label{ack-ch02-eq03}
\end{equation}

\textbf{Independent Condition of the Quantization error}
The independent condition mentioned herein is relevant whether the correlation between the input of the quantization and the quantization error is zero or not.  
If we let a quantization value ${X}^Q = k Q_p^{-1} \in \mathbb{Q}$ for $k \in \mathbb{Z}$, we can evaluate the correlation of ${X}^Q$ and the quantization error $\bar{\epsilon}^q$ such that 
\begin{equation}
\begin{aligned}
&\mathbb{E}_{\bar{\varepsilon}^q} [X(X - X^Q)\vert X^Q=k Q_p^{-1}] 
= \mathbb{E}_{\bar{\varepsilon}^q} [X \bar{\epsilon}^q \vert X^Q=k Q_p^{-1}] \\
&= \int_{-\frac{1}{2}Q_p^{-1}}^{\frac{1}{2} Q_p^{-1}} Q_p x \epsilon^q d\bar{\epsilon}^q, \quad \because x = x^Q \vert_{x^Q=k Q_p^{-1}} + \bar{\epsilon}^q \vert_{\bar{\epsilon}^q = -Q_p^{-1}/2}^{\bar{\epsilon}^q = Q_p^{-1}/2}, \; dx = d\bar{\epsilon}^q \\
&= \int_{(-\frac{1}{2} + k)Q_p^{-1}}^{(\frac{1}{2} + k)Q_p^{-1}} Q_p x( x - x^Q) dx \\
&= Q_p \left[ \frac{x^3}{3} \Bigg\vert_{(-\frac{1}{2} + k)Q_p^{-1}}^{(\frac{1}{2} + k)Q_p^{-1}}
- \frac{k}{Q_p} \frac{x^2}{2} \Bigg\vert_{(-\frac{1}{2} + k)Q_p^{-1}}^{(\frac{1}{2} + k)Q_p^{-1}} \right] \\
&= Q_p \left[ \frac{1}{3 Q_p^3} \left( \left( \frac{1}{2} + k \right)^3 - \left( -\frac{1}{2} + k \right)^3 \right)
- \frac{k}{2 Q_p^3} \left( \left( \frac{1}{2} + k \right)^2 - \left( -\frac{1}{2} + k \right)^2 \right) \right] \\
&= Q_p^{-2} \left[k^2  + \frac{1}{12} - k^2 \right] 
= \frac{1}{12} Q_p^{-2} = c_0 Q_p^{-2}.
\end{aligned}
\label{supp-04}
\end{equation}
In \eqref{supp-04}, to get a positive correlation value, we establish the quantization value $X^Q$ and quantization input $X$ as the following equation:
\begin{equation}
X = X^Q + \bar{\epsilon}^q, \; \bar{\epsilon}^q = Q_P^{-1} \bar{\varepsilon}^q \in \mathbb{R}[-Q_p^{-1}/2, Q_p^{-1}/2]    
\label{supp-05}
\end{equation}
, where the scalar quantization error $\bar{\epsilon}^q$ has negative sign to originally defined quantization error $\epsilon^q$ such that $\bar{\epsilon}^q = - \epsilon^q$.

Likewise as previously mentioned, we can rewrite \eqref{supp-04} as follows
\begin{equation}
\begin{aligned}
\mathbb{E}_{\bar{\varepsilon}^q} [X \bar{\epsilon}^q \vert X^Q=k Q_p^{-1}]
&= \mathbb{E}_{\bar{\varepsilon}^q} [(X^Q + \bar{\epsilon}^q) \bar{\epsilon}^q \vert X^Q=k Q_p^{-1}]
= \mathbb{E}_{\bar{\varepsilon}^q} [(k Q_p^{-1} + \bar{\epsilon}^q)\bar{\epsilon}^q ] \\
&= k Q_p^{-1} \mathbb{E}_{\bar{\varepsilon}^q} \bar{\epsilon}^q + \mathbb{E}_{\bar{\varepsilon}^q} (\bar{\epsilon}^q)^2
= 0 + c_0 Q_p^{-2}
\end{aligned}
\end{equation}

The result of \eqref{supp-04} represents that the quantization input and quantization error are correlated with the equal variance of the quantization error, so those are not generally independent. 
This correlation enables to bring about early paralysis in the learning process under quantized directional derivation. 
Considering \eqref{supp-04}, the quantized directional derivative $h^Q \in \mathbb{R}$ can be zero, even if there exists a quantization error such that 
\begin{equation}
\mathbb{E}_{\bar{\varepsilon}^q} [h \epsilon^q \vert h^Q = kQ_p^{-1}] \bigg\vert_{k=0}
= \int_{-\frac{1}{2 Q_p}}^{\frac{1}{2 Q_p}} Q_p h( h - h^Q ) dx \bigg\vert_{h^Q = 0} = c_0 Q_p^{-2}.
\label{supp-06}
\end{equation}

If the quantization parameter is sufficiently large and leads $Q_p^{-1} \downarrow 0$, we can treat the quantization error as an independent white noise to the quantization input. 
However, the quantization parameter is not large enough in the early stage of the learning process, so such a relatively small quantization parameter raises the early paralysis of the learning algorithm.
Considering the learning equation based on the quantized directional derivative such that $X_{\tau+1}^Q = X_{\tau} + h^Q$, we can note that the dependent property between the quantization error and the quantization input leads to the diminishing of the quantized directional derivative such that
\begin{equation}
| h | \downarrow 0 \implies | \epsilon^q | \downarrow 0 \implies |h + \epsilon^q | = | h^Q | \downarrow 0.
\label{supp-07}
\end{equation}
Accordingly, the learning process does not work when we devise a learning equation based on the quantized directional derivation.
However, if there exists a compensated quantization $\breve{\epsilon}^q$ that is independent of the input $X$ such that $\mathbb{E}_{\breve{\epsilon}^q}[X \breve{\epsilon}^q | X^Q=kQ_p^{-1}] = 0$, the independent quantization error provides
\begin{equation}
| h | \downarrow 0 \text{ and } | \breve{\epsilon}^q | > 0 \implies 
|h + \breve{\epsilon}^q |_{| h | \downarrow 0} \leq |h|_{| h | \downarrow 0}  +  |\breve{\epsilon} | \triangleq |h^Q|_{| h | \downarrow 0} > 0  
\label{supp-08}
\end{equation}
This intuitive consideration states that the independent quantization error to the quantization input enables to avoidance of early learning paralysis raised by $|X| \downarrow 0$.
Consequently, we should establish a compensation function that enables it to be independent between the quantization input and the quantization error. 

%=====================================================
\subsection{Avoid Early Paralysis of the proposed algorithm}
%=====================================================
\textbf{Dithering for Independent Condition of the Quantization error}
We assume that there exists a random variable $\boldsymbol{z} \in \mathbb{R}$ defined on $\mathbb{R}[-\frac{1}{2 Q_p}, \frac{1}{2 Q_p})$ with a uniform distribution $p_{{\boldsymbol{z}}}(x) = Q_p$. 
Since the probability density function of the random variable $\boldsymbol{z}$ and $\epsilon^q$ is equal, we calculate the expectation such that 
\begin{equation}
\begin{aligned}
\mathbb{E}_{\boldsymbol{z}} [(X + \boldsymbol{z})^Q|X=x] 
&= \int_{\frac{-1}{2Q_p}}^{\frac{1}{2Q_p}} p_{{\boldsymbol{z}}}(z) \cdot \left( x + z + \epsilon^q \right) dz \\
&= \int_{x -\frac{1}{2Q_p}}^{x+\frac{1}{2Q_p}} Q_p (y + \epsilon^q) dy, \; \because y = x + z \implies dy = dz, \; y \in \mathbb{R}\left[x-\frac{1}{2Q_p}, x+\frac{1}{2Q_p}\right] \\  
&= Q_p \left( \int_{x -\frac{1}{2Q_p}}^{x+\frac{1}{2Q_p}} y dy + \int_{x -\frac{1}{2Q_p}}^{x+\frac{1}{2Q_p}} \epsilon^q dy \right)
\; \because y^Q + \epsilon = y \implies d\epsilon = dy \\
&= Q_p \left( \frac{y^2}{2} \bigg\vert_{x -\frac{1}{2Q_p}}^{x+\frac{1}{2Q_p}} y dy + \int_{-\frac{1}{Q_p}}^{\frac{1}{Q_p}} \epsilon^q d\epsilon^q \right) \; \because \epsilon^q = x \pm \frac{1}{2Q_p} - y^Q \\
&= Q_p \left(\frac{1}{2} \left[ \left(x+\frac{1}{2Q_p} \right)^2 - \left(x + \frac{1}{2Q_p}\right)^2 \right] 
+ \frac{{\epsilon^q}^2}{2} \bigg\vert_{-\frac{1}{Q_p}}^{\frac{1}{Q_p}}  \right) \\
&\because \epsilon^q \in \{\epsilon^q \in \mathbb{R} \vert x - (x \pm 1/2Q_p) \pm 1/2Q_p \} = \mathbb{R}[-1/Q_p, 1/Q_p] \\
&= Q_p \left( \frac{1}{2} \frac{2x}{Q_p}  + \frac{{\epsilon^q}^2}{2} \bigg\vert_{-\frac{1}{2Q_p}}^{\frac{1}{2Q_p}} \right) 
= Q_p \cdot \frac{x}{Q_p}  + Q_p \cdot 0 = x
\end{aligned} 
\label{supp-09}
\end{equation}

From the result of \eqref{supp-09}, we can obtain the correlation between the quantization input and the quantization error with an additional uniformly distributed noise as follows:
\begin{equation}
\begin{aligned}
&\mathbb{E}_{X, \boldsymbol{z}}[X(X - (X + \boldsymbol{z})^Q) \vert X^Q = k Q_p^{-1}]\\
&= \int_{-(\frac{1}{2}+k)Q_p^{-1}}^{(\frac{1}{2}+k)Q_p^{-1}} p_{X}(x) \int_{-\frac{Q_p^{-1}}{2}}^{\frac{Q_p^{-1}}{2}} p_{\boldsymbol{z}}(z) x (x - (x + z)^Q) dz dx \\
&= \int_{-(\frac{1}{2}+k)Q_p^{-1}}^{(\frac{1}{2}+k)Q_p^{-1}} p_{X}(x)  \left( x^2 \int_{-\frac{Q_p^{-1}}{2}}^{\frac{Q_p^{-1}}{2}} p_{\boldsymbol{z}}(z) dz - x \int_{-\frac{Q_p^{-1}}{2}}^{\frac{Q_p^{-1}}{2}} p_{\boldsymbol{z}}(z) (x + z)^Q dz \right) dx \\
&= \int_{-(\frac{1}{2}+k)Q_p^{-1}}^{(\frac{1}{2}+k)Q_p^{-1}} p_{X}(x)  (x^2 - x^2)  dx  = 0.
\end{aligned}
\label{supp-10}
\end{equation}

By the result of \eqref{supp-10}, we are aware that the additional noise $\boldsymbol{z}$ enables the quantization error and the quantization input to be independent.
The technique described in \eqref{supp-10} is known as the dithering to fulfill the independent property among the quantization error and the quantization input(provided by \citet{Marco_2005}, and \citet{Gray:2006}).

Whereas the dithering employs an additive noise with a uniform distribution herein, equation 12 presents that even an additive noise with an appropriate symmetrical distribution can satisfy the independent condition.
Based on such a conjecture, we present the compensation function to satisfy the independent condition for the quantization error in the following section. 

When we elaborate the above equations with the formulas incorporating the expectation symbol, we can rewrite \eqref{supp-09} such that
\begin{equation}
\mathbb{E}_{\boldsymbol{z}} [(X + \boldsymbol{z})^Q|X=x] 
= \mathbb{E}_{\boldsymbol{z}} [x + \boldsymbol{z} - \bar{\epsilon}^q]
= x + \mathbb{E}_{\boldsymbol{z}} \boldsymbol{z} - \mathbb{E}_{\boldsymbol{z}} \bar{\epsilon}^q
= x, \; \therefore \mathbb{E}_{\boldsymbol{z}} \boldsymbol{z} = \mathbb{E}_{\boldsymbol{z}} \bar{\epsilon}^q
\label{supp-11}
\end{equation}
The result of \eqref{supp-11} presents that we can measure the quantization error $\bar{\epsilon}^q$ or ${\epsilon}^q$ with the probability density of $\boldsymbol{z}$ due to the equal uniform distribution. 
Hence we can regard the additional noise for the dithering $\boldsymbol{z}$ as a transformation of the quantization error. 
Additionally, we note that the additional noise as the transformation for the dithering requires the changed sign of the quantization error. 

Finally, holding the dithering condition for additional noise, we rewrite \eqref{supp-10} such that  
\begin{equation}
\begin{aligned}
\mathbb{E}_{X, \boldsymbol{z}}[X(X - (X + \boldsymbol{z})^Q) \vert X^Q = k Q_p^{-1}] 
&= \mathbb{E}_{X, \boldsymbol{z}}[X(X - (X + \boldsymbol{z} + \epsilon^q )) \vert X^Q = k Q_p^{-1}] \\
&= \mathbb{E}_{X, \boldsymbol{z}}[X^2 - X^2 - X (\boldsymbol{z} + \epsilon^q) \vert X^Q = k Q_p^{-1}] \\
&= \mathbb{E}_{X, \boldsymbol{z}}[- X (\boldsymbol{z} + \epsilon^q) \vert X = k Q_p^{-1} + \bar{\epsilon}^q] \\
&= -\mathbb{E}_{\boldsymbol{z}}[(k Q_p^{-1} + \bar{\epsilon}^q) (\boldsymbol{z} + \epsilon^q) ] \\
&= - k Q_p^{-1} \left(\mathbb{E}_{\boldsymbol{z}}\boldsymbol{z} + \mathbb{E}_{\boldsymbol{z}} \epsilon^q \right) 
- \mathbb{E}_{\boldsymbol{z}} \bar{\epsilon}^q \boldsymbol{z} - \mathbb{E}_{\boldsymbol{z}} \bar{\epsilon}^q \epsilon^q \\
&= 0 - \mathbb{E}_{\boldsymbol{z}} (\boldsymbol{z})^2 + \mathbb{E}_{\epsilon^q} (\epsilon^q)^2 = 0.
\end{aligned}
\label{supp-12}
\end{equation}
In \eqref{supp-12}, we note that $\mathbb{E}_{\boldsymbol{z}} \bar{\epsilon}^q \boldsymbol{z} = \mathbb{E}_{\boldsymbol{z}} (\boldsymbol{z})^2$ and $\mathbb{E}_{\boldsymbol{z}} \bar{\epsilon}^q \epsilon^q = - \mathbb{E}_{\boldsymbol{z}} (\epsilon^q)^2 = -\mathbb{E}_{\epsilon^q} (\epsilon^q)^2$.

\textbf{The Compensation Function for Early Paralysis}
We establish the compensation function to avoid early paralysis in the quantized directional derivative-based learning process as follows:
\begin{equation}
r(\tau, \boldsymbol{X}_{\tau}) = \lambda \cdot \left( \frac{\exp(-\varkappa(\tau - \tau_0))}{1 + \exp(-\varkappa(\tau - \tau_0))} \cdot \frac{h(\boldsymbol{X}_{\tau})}{\| h(\boldsymbol{X}_{\tau}) \|} \right),\quad \tau_0 \in \mathbb{Z}^{++}
%\label{avoid-eq03}
\label{ack-ch02-eq05}
\end{equation}
We consider a directional derivative with one dimension, $h \in \mathbb{R}$, for convenience of discussing. 
Since the compensation function defined in \eqref{ack-ch02-eq05} contains a normalized directional derivative $h/\|h\|$, for $\tau \ll \tau_0$, we assume that the range of the compensation function such that 
\begin{equation}
    r(\tau, \boldsymbol{X}_{\tau}) \triangleq \lambda \cdot sgn(\boldsymbol{X}_{\tau}) , \quad  \forall \tau \ll \tau_0,\; \text{ and } \tau \gg \tau_0
\label{ack-ch02-eq06}    
\end{equation}
, where $sgn$ denotes the sign function of each element in $\boldsymbol{X}_{\tau}$ such that $sgn(\boldsymbol{X}) = \sum_{i=1}^d sgn(\boldsymbol{X} \cdot \boldsymbol{e}^{(i)}) \boldsymbol{e}^{(i)}$.

To analyze the effectiveness of the compensation function, we suppose the scalar input such that $X_{\tau} \in \mathbb{R}$. 
Intuitively, since we assume that the probability density function of the compensation function is an equal Bernoulli distribution $P_{\boldsymbol{r}}(x)$ with the value of $\{-\lambda, \lambda\}$ for the domain $\{x|x<0 \}$ and $\{x|x\ge 0 \}$, respectively, we can straightforwardly obtain the expectation and variance of the compensation function for a $\tau$ such that $| \tau - \tau_0 | \gg 0$ as follows:
\begin{equation}
\begin{aligned}
\mathbb{E}_{r} r(\tau, X) 
&= \sum_{r=-\lambda}^{r=\lambda} P_{\boldsymbol{r}}(x) r(\cdot, X) 
= \lambda \left(\int_{\{ x | x < 0 \}} P_{\boldsymbol{r}}(x) sgn(x) dx + \int_{\{ x | x \geq 0 \}} P_{\boldsymbol{r}}(x) sgn(x) dx \right) \\
&= \lambda \left(\int_{\{ x | x < 0 \} \cup \{ x | x \geq 0 \}} P_{\boldsymbol{r}}(x) sgn(x) dx \right)
= \lambda \int_{\mathbb{R}} P_{\boldsymbol{r}}(x) r(\tau, X) dx
= 0,  \\
Var_{r} r(\tau, X) &= \mathbb{E}_{r} r^2(\tau, X) = \lambda^2 \int_{\mathbb{R}} P_{\boldsymbol{r}}(x) dx = \lambda^2, \quad
\mathbb{E}_{r} r(\tau, X) r(s, X) = 0, \forall t \neq s, \; |s - \tau_0| \gg 0.
\label{ack-ch02-eq07}
\end{aligned}
\end{equation}

Moreover, the statistical quantity of the compensation function exhibits that the summation of $r(\tau, X)$, i.e., $Y_{\tau} \triangleq \sum_{k=0}^{\tau} r(\tau, X), \forall |\tau - \tau_0 | \gg 0$, yields a standard Wiener process in the sense of the following moments: 
\begin{equation}
\mathbb{E}_{r} Y_{\tau} = \sum_{k=0}^{\tau-1} \mathbb{E}_{r} r(k, X) = 0, \;\;
\mathbb{E}_{r} Y_{\tau}^2 = \sum_{k=0}^{\tau-1} \mathbb{E}_{r} r^2(k, X) + 2 \sum_{k=0}^{\tau-1}\sum_{l, l \neq k}^{\tau-1} r(k, X) r(s, X) = \lambda^2 \tau.
\label{ack-ch02-eq08}  %% eq 19
\end{equation}

Holding the definition of the quantization, we can write the quantization of the summation of the directional derivative and the compensation such that 
\begin{equation}
\begin{aligned}
(\lambda h(X) + r(\tau, X))^Q 
&= Q_p^{-1} \left[ Q_p \cdot (\lambda h(X) + r(\tau, X)) + \varepsilon^q \right] \\
&= \lambda h(X) + \varepsilon^q Q_p^{-1} + r(\tau, X) \\
&= (\lambda h(X))^Q + r(\tau, X) = (\lambda h(X))^Q + \lambda \cdot sgn(h(x))
\label{ack-ch02-eq09}  %% (20)
\end{aligned}
\end{equation}

Finally, we can expand the statistical quantities to the vector-valued compensation $r: \mathbb{R} \times \mathbb{R}^d \mapsto \mathbb{R}^d$ with the following equation:
\begin{equation}
r(\tau, \boldsymbol{X}) = \sum_{i=1}^d (r(\tau, \boldsymbol{X}) \cdot \boldsymbol{e}^{i}) \boldsymbol{e}^{i}, \quad \forall \boldsymbol{X} \in \mathbb{R}^d
\label{ack-ch02-eq10}   %% 21
\end{equation}

Thus, we can obtain the statistical moments of the vector-valued compensation for all $\boldsymbol{X} \in \mathbb{R}^d$ and $|\tau - \tau_0 | \gg 0$ as follows:
\begin{equation}
\begin{aligned}
\mathbb{E}_{r} r(\tau, \boldsymbol{X}) 
&= \sum_{i=1}^d \left( \mathbb{E}_{r} [r(\tau, \boldsymbol{X}) \cdot \boldsymbol{e}^{i}] \right) \boldsymbol{e}^{i} 
= \sum_{i=1}^d 0 \cdot \boldsymbol{e}^{i}  = 0 \\
\mathbb{E}_{r} r^2(\tau, \boldsymbol{X}) 
&= \mathbb{E}_{r} \left( \sum_{i=1}^d (r(\tau, \boldsymbol{X}) \cdot \boldsymbol{e}^{i} )\boldsymbol{e}^{i} \right) \cdot \left( \sum_{j=1}^d (r(\tau, \boldsymbol{X}) \cdot \boldsymbol{e}^{j} ) \boldsymbol{e}^{j} \right) \\
&= \mathbb{E}_{r} \left( \sum_{i=1}^d \sum_{j=1}^d (r(\tau, \boldsymbol{X}) \cdot \boldsymbol{e}^{i} )(r(\tau, \boldsymbol{X}) \cdot \boldsymbol{e}^{j} ) \boldsymbol{e}^{i} \cdot \boldsymbol{e}^{j} \right) \\
&= \mathbb{E}_{r} \sum_{i=1}^d (r(\tau, \boldsymbol{X}) \cdot \boldsymbol{e}^{i} )^2 = \lambda^2 \, d \\
Cov \, r(\tau, \boldsymbol{X}) 
&\triangleq \mathbb{E}_{r} r(\tau, \boldsymbol{X}) r(\tau, \boldsymbol{X})^T \\
&= \mathbb{E}_{r} \sum_{i=1}^d (r(\tau, \boldsymbol{X}) \cdot \boldsymbol{e}^{i} )^2 \boldsymbol{e}^{i} \otimes \boldsymbol{e}^{i} + 2 \sum_{i=1}^d \sum_{j=1, j neq i}^d (r(\tau, \boldsymbol{X}) \cdot \boldsymbol{e}^{i} )(r(\tau, \boldsymbol{X}) \cdot \boldsymbol{e}^{j} ) \boldsymbol{e}^{i} \otimes {\boldsymbol{e}^{j}} \\
&= \lambda \, \boldsymbol{I}_d
\end{aligned}
\label{ack-ch02-eq11}  %% eq22
\end{equation}
, where $\boldsymbol{I}_d$ denotes an identity matrix such that $\boldsymbol{I}_d \in \mathbb{R}^{d \times d}$.
Henceforth, we prove the following theorem using the derived statistical properties of the proposed compensation function.

%====================================
%\begin{theorem} % theorem for weak convergence  Theorem 3.1
%\label{theorem_01-02}
\textbf{Theorem 3.1}
%====================================
Let the quantized directional derivatives $h^Q : \mathbb{Z}^+ \times \displaystyle \R^d \mapsto \mathbb{Q}^d$ such that 
\begin{equation}
h^Q (\boldsymbol{X}_{\tau}^Q) \triangleq \frac{1}{Q_p} \left\lfloor Q_p \cdot (\lambda h(\boldsymbol{X}_{\tau}^Q) + r(\tau, \boldsymbol{X}_{\tau}^Q)) + 0.5\right\rfloor
\end{equation}
, where $r(\tau, \boldsymbol{X}_{\tau})$ denotes a compensation function such that $r:\mathbb{Z}^+ \times \mathbb{R}^d \mapsto \displaystyle \R^d \{-1, 1\}$.
Then, the quantization input $h(\boldsymbol{X}_{\tau}^Q)$ and the quantization error $\boldsymbol{\epsilon}_{\tau}^q$ is uncorrelated when the quantized directional derivative $h(\boldsymbol{X}_{\tau}^Q)$ is 0 such that $\mathbb{E}_{\boldsymbol{\epsilon}_{\tau}^q} [h(\boldsymbol{X}_{\tau}^Q) \boldsymbol{\epsilon}_{\tau}^q \vert h^Q(\boldsymbol{X}_{\tau}^Q) = k Q_p^{-1}] = 0$.
%\end{theorem}

\begin{proof}
From the definition of the quantization error $\boldsymbol{\bar{\epsilon}}_{\tau}^q = - \boldsymbol{\epsilon}_{\tau}^q = \boldsymbol{X}_{\tau} - \boldsymbol{X}_{\tau}^Q$ and substituting $h(\boldsymbol{X}_{\tau}^Q)$, $(h(\boldsymbol{X}_{\tau}) + r(\tau, h(\boldsymbol{X}_{\tau})))^Q$ into $X_{\tau}$, $X_{\tau}^Q$ respectively, we can rewrite the expectation $\mathbb{E}_{\boldsymbol{X}, r, \boldsymbol{\epsilon}_{\tau}^q} [h(\boldsymbol{X}_{\tau}^Q) \boldsymbol{\epsilon}_{\tau}^q \vert h^Q(\boldsymbol{X}_{\tau}^Q) = k Q_p^{-1}]$ such that 
\begin{equation}
\begin{aligned}
&\mathbb{E}_{\boldsymbol{X}, r, \boldsymbol{\epsilon}_{\tau}^q} [h(\boldsymbol{X}_{\tau}^Q) \boldsymbol{\epsilon}_{\tau}^q \vert h^Q(\boldsymbol{X}_{\tau}^Q) = k Q_p^{-1}] \\
&= \mathbb{E}_{\boldsymbol{X}, r, \boldsymbol{\epsilon}_{\tau}^q} [h(\boldsymbol{X}_{\tau}^Q) \left( (\lambda h(\boldsymbol{X}_{\tau}) + r(\tau, \boldsymbol{X}_{\tau}))^Q - \lambda h(\boldsymbol{X}_{\tau} \right) \vert h^Q(\boldsymbol{X}_{\tau}^Q) = k Q_p^{-1}] \\
&= \mathbb{E}_{\boldsymbol{X}, r, \boldsymbol{\epsilon}_{\tau}^q} [h(\boldsymbol{X}_{\tau}^Q) \left( (\lambda h(\boldsymbol{X}_{\tau}) + r(\tau, \boldsymbol{X}_{\tau}) + \boldsymbol{\epsilon}_{\tau}^q) - \lambda h(\boldsymbol{X}_{\tau} \right) \vert h^Q(\boldsymbol{X}_{\tau}^Q) = k Q_p^{-1}] \\
&= \mathbb{E}_{\boldsymbol{X}, r, \boldsymbol{\epsilon}_{\tau}^q} [\lambda (h^2(\boldsymbol{X}_{\tau}^Q) - h^2(\boldsymbol{X}_{\tau}^Q)) + r(\tau, \boldsymbol{X}_{\tau}) + \boldsymbol{\epsilon}_{\tau}^q) \vert h^Q(\boldsymbol{X}_{\tau}^Q) = k Q_p^{-1}] \\
&= \mathbb{E}_{r, \boldsymbol{\epsilon}_{\tau}^q} [r(\tau, \boldsymbol{X}_{\tau}) + \boldsymbol{\epsilon}_{\tau}^q) ] \\
&= \mathbb{E}_{r} [r(\tau, \boldsymbol{X}_{\tau})] + \mathbb{E}_{\boldsymbol{\epsilon}_{\tau}^q} [\boldsymbol{\epsilon}_{\tau}^q ] 
= 0 + 0 = 0
\end{aligned}
\label{ack-ch02-eq12}  %% eq24
\end{equation}  
\end{proof}

We can model the early paralysis of learning such that $h^Q(\boldsymbol{X}_{\tau}) = k \cdot Q_p \vert_{k=0} = 0$ for a quantization parameter $Q_p > 0$ for a $\tau > 0$. 
Theorem \ref{theorem_01-02} describes that even though the quantized directional derivative $h^Q(\boldsymbol{X}_{\tau}) = 0$, the proposed compensation function yields an independent noise to $h^Q$, and the noise can work the learning process avoiding early paralysis as follows:
\begin{equation}
\begin{aligned}
\boldsymbol{X}_{\tau + 1}^Q 
&= \boldsymbol{X}_{\tau}^Q + \frac{1}{Q_p} \left\lfloor Q_p \cdot (\lambda h(\boldsymbol{X}_{\tau}^Q) + r(\tau, \boldsymbol{X}_{\tau}^Q)) + 0.5\right\rfloor \\
&= \boldsymbol{X}_{\tau}^Q + \frac{1}{Q_p} \left\lfloor Q_p \cdot (\lambda h(\boldsymbol{X}_{\tau}^Q) + 0.5) + Q_p r(\tau, h(\boldsymbol{X}_{\tau}^Q)) \right\rfloor \\
&= \boldsymbol{X}_{\tau}^Q + \frac{1}{Q_p} \left\lfloor Q_p \cdot (\lambda h(\boldsymbol{X}_{\tau}^Q) + 0.5) + Q_p \lambda sgn(h(\boldsymbol{X}_{\tau}^Q)) \right\rfloor \bigg\vert_{\lambda = k \cdot Q_p, \, k \in \mathbb{Z}^+}\\
&= \boldsymbol{X}_{\tau}^Q + \frac{1}{Q_p} \left\lfloor Q_p \cdot (\lambda h(\boldsymbol{X}_{\tau}^Q) + 0.5) \right\rfloor + \frac{1}{Q_p} \cdot Q_p \lambda \, sgn(h(\boldsymbol{X}_{\tau}^Q)) \\
&= \boldsymbol{X}_{\tau}^Q + 0 + \lambda \, sgn(h(\boldsymbol{X}_{\tau}^Q)). 
\end{aligned}
\label{ack-ch02-eq13}  %% eq25
\end{equation}
In \eqref{ack-ch02-eq13}, we assume that $\lambda = k \cdot Q_p, \; k \in \mathbb{Z}^+$ for convenience. 

While the proposed compensation function allows for avoiding early paralysis, we can't verify the convergence of the learning equation based on the proposed method due to the variance of the process $Y_t$ represented in \eqref{ack-ch02-eq08}.   
In the proposed compensation function, we designed the critical time index $\tau_0$ where the function denotes 0.5 and drops to zero after the critical time. 
While the proposed time-dependent compensation approach facilitates the convergence of the learning algorithm to a stable point beyond the critical time, our design does not provide an absolute guarantee of complete convergence for the learning algorithm.

Consequently, we will design an improved compensation to secure convergence by adding a time-dependent variance. 
One possible intuitive design we suggest is as follows: 
\begin{equation}
r(\tau, h(\boldsymbol{X}_{\tau}^Q)) \triangleq \lambda \sqrt{\frac{1}{\log \log (\kappa \tau + c_0)}} \frac{h(\boldsymbol{X}_{\tau})}{\| h(\boldsymbol{X}_{\tau}) \|}.   
\end{equation}

%%%%%%%%%%%%%%%%%%%%%%%%%%%%%%%%%%%%%%%%%%%%%%%%%%%%%%%%%%%%%%%%%%%%%%%
\section{Fundamental Learning Equation based on Quantization}  % Theorem 3
%%%%%%%%%%%%%%%%%%%%%%%%%%%%%%%%%%%%%%%%%%%%%%%%%%%%%%%%%%%%%%%%%%%%%%%
For convenience of discussion, we analyze the Quantized Stochastic Gradient Langevin Dynamics(QSGLD) instead of the Quantized Stochastic Langevin Dynamics(QSLD). 
We provide the learning equation of QSGLD is  as follows :
\begin{equation}
\boldsymbol{X}_{\tau + 1}^Q = \boldsymbol{X}_{\tau}^Q - \lambda \nabla_{\boldsymbol{x}} \tilde{f}_{\tau}(\boldsymbol{X}_{\tau}) +  Q_p^{-1}(\tau) \boldsymbol{\varepsilon}_{\tau}^q.
%\label{ch04-eq04}
\label{ack-ch03-eq01}
\end{equation}
For the analysis of the proposed learning equation, we establish the two systems of time indexes.
One is a single-sided time system based on a mini-batch index $\tau$, the other is a double-sided time system based on $\tau$ and an epoch $t_e$ as the unit summation of mini-batches.

In the single-sided time system, we define the time index with a time difference $\delta_{\tau}$ intuitively such that
\begin{equation}
    \tau + 1 \triangleq \tau + \delta_{\tau}, \quad \delta_{\tau} \triangleq \frac{1}{B} 
\label{ack-ch03-eq02}
\end{equation}
, where $B$ is the number of mini-batches for a unit epoch.
We assume that the learning rate $\lambda$ is the reciprocal of the number of mini-batches $B$, so we can rewrite the time index as follows:
\begin{equation}
    \tau + 1 \cdot \lambda \triangleq \tau + \lambda, \quad \lambda \triangleq \frac{1}{B} = \delta_{\tau}.
\label{ack-ch03-eq03}
\end{equation}
Consequently, we treat the increased time index $\tau+1$ as equivalent to $\tau+\lambda$, so we abbreviate the learning rate behind the time index number when we use the number-based single-sided time system such that $\tau + 1 \cdot (\lambda)$.

In the double-sided time system, we define the epoch-based time index $t_e$ with the mini-batch-based time index $\tau$ such that 
\begin{equation}
    t_e + 1 = t_e + \sum_{\tau=0}^{B-1} \delta_{\tau} = t_e + \lambda B, \quad \delta_{\tau} = \frac{1}{B}, \; \forall \tau \in \mathbb{Z}[0, B).
\label{ack-ch03-eq04}    
\end{equation}

Therefore, we establish a connection with $\tau$ and $t_e$ such that
\begin{equation}
    \tau = t_e \cdot B + k, \; k \in \mathbb{Z}[0, B), \quad t_e = \left\lfloor \frac{\tau}{B} \right\rfloor.
\label{ack-ch03-eq05}        
\end{equation}

Using these time indexing systems, we derive a stochastic difference equation based on the uniform distributed quantization error according to WNH in the following section.

Before wrapping up the section, we discuss the sample-based time system.
Each mini-batch contains $n_B$ samples equally, so we recognize there are $N_T = B \cdot n_B$ samples as the total numbers of data.
Accordingly, we can establish the time index with such a unit sample, and the unit sample-based time system is natural in signal processing.  
However, the artificial intelligence framework such as the PyTorch works on a unit of mini-batch, not a unit sample. 
When we define the size of a mini-batch to be one, in other words, each mini-batch contains one sample, we can work the learning process to a unit sample. 
Even though such a process for a unit sample is possible, the process for a unit sample is not practically valuable for heavy computation time.

%=====================================================
\subsection{Transformation to Gaussian Wiener Process}
%=====================================================
%-----------------------------------------------------
\begin{algorithm}[tb]
    \caption{Virtual algorithm for transformation from the quantization error to a Gaussian random variable}
    \label{supp_alg-1}
\begin{algorithmic}[1]
    \State $Q_p \leftarrow Q_p(-1, \tau)$       \Comment {Set a Pure Quantization Parameter}
    \State Compute $h (\boldsymbol{X}_{\tau})$ at $\tau$   \Comment {Compute a Search Direction Vector}
    %\State $\bar{h}_{\tau} \leftarrow -\alpha \nabla f(\boldsymbol{X}_{\tau}) + r(\tau)$ \Comment {Compute the Directional vector with $r(\tau)$}
    \State $h_{\tau}^Q \leftarrow \frac{1}{Q_p}\lfloor Q_p \cdot (-\lambda h(\boldsymbol{X}_{\tau}) + 0.5 Q_p^{-1})\rfloor $ \Comment {Quantization of the Search Direction Vector}
    \State $\boldsymbol{\varepsilon}_{\tau} \leftarrow Q_p(h_{\tau}^Q - h(\boldsymbol{X}_{\tau}))$ \Comment {General Updating Rule for Learning}
    \State Compute $\boldsymbol{z}_{\tau} \leftarrow Q_p^{-1}(\boldsymbol{\varepsilon}_{\tau}, \tau) \boldsymbol{\varepsilon}_{\tau}$   \Comment {Compute Gaussian random variable}
\end{algorithmic}
\end{algorithm} 
%-----------------------------------------------------

Firstly, we derive an intuitive stochastic difference equation of the quantization learning based on the single-side time system. 
Since the single-sided time system is based on a unit mini-batch index, we should directly transform the quantization error into a standard Wiener process.
For the transformation, we exploit the quantization parameter with two parameters $Q_p(\varepsilon^q, t)$ as the transform function from the uniformly distributed random variable to a Gaussian random variable. 
The random number generation of a Gaussian random variable based on a uniformly distributed random variable is a standard and widely used method. 

Box-Muller algorithm (by \citet{Box_Muller_1958}), the Ziggurat algorithm (by \citet{Marsaglia_1963} and \citet{Marsaglia_2000}) and
the inverse transform sampling (by \citet{Thomas_2007}) are representative transforms.

Algorithm $\ref{supp_alg-1}$ describes the calculation of the quantization error $\boldsymbol{\epsilon_{\tau}^q}$, the factor for the quantization $\boldsymbol{\varepsilon}_{\tau}^q$, and the generation of a Gaussian random variable using the quantization parameter $Q_p(\varepsilon^q, t)$. 

If we establish the $\bar{\eta}(\boldsymbol{\varepsilon}_{\tau}^q) \triangleq \boldsymbol{z}_{\tau} \in \mathbb{R}^d$ to be a Gaussian random generator as the factorizing function of the quantization parameter, we can obtain the Gaussian random variable $\boldsymbol{z}_{\tau} \overset{i.i.d.}{\sim} \mathcal{N}(\boldsymbol{z}; 0, \boldsymbol{I}_d)$ as follows: 
\begin{equation}
    Q_p^{-1}(\boldsymbol{\varepsilon^q}, \tau)\boldsymbol{\varepsilon}_{\tau}^q 
    = b^{-\bar{p} (\tau)} \eta^{-1}(\boldsymbol{\varepsilon}_{\tau}^q)\cdot \boldsymbol{\varepsilon}_{\tau}^q 
    = \frac{\sqrt{\lambda} b^{-\bar{p} (\tau)}}{\| \boldsymbol{\varepsilon}_{\tau}^q \|^2} \bar{\eta}(\boldsymbol{\varepsilon}_{\tau}^q) \boldsymbol{\varepsilon}_{\tau}^q \cdot \boldsymbol{\varepsilon}_{\tau}^q
    = \sqrt{\lambda} b^{-\bar{p} (\tau)} \boldsymbol{z}_{\tau}, \; b^{-\bar{p} (\tau)} \in \mathbb{Q}
\label{ack-ch03-eq06}    
\end{equation}
, where $\eta^{-1}(\boldsymbol{\varepsilon}_{\tau}^q)$ is a mapping such that $\eta^{-1} : \mathbb{R}^d \mapsto \mathbb{R}^d$.

Additionally, if the input of $\eta^{-1}$ is $-1$, we define the quantization parameter as not a mapping but a scalar function such that $Q_p^{-1}(\tau) = \eta_0^{-1} b^{-\bar{p}(t)} \in \mathbb{Q}$.  

Using \eqref{ack-ch03-eq05}, we can rewrite \eqref{ch04-eq04} as QSGLD such that 
\begin{equation}
\boldsymbol{X}_{\tau + 1}^Q 
= \boldsymbol{X}_{\tau}^Q - \lambda \nabla_{\boldsymbol{x}} \tilde{f}_{\tau}(\boldsymbol{X}_{\tau}) +  \sqrt{\lambda} \cdot b^{-\bar{p}(\tau)} \boldsymbol{z}_{\tau}.
%\label{ch04-it-eq01}
\label{ack-ch03-eq07}
\end{equation}

The stochastic difference \eqref{ack-ch03-eq07} formulates the representative Euler-Maruyama approximation of the following SDE: 
\begin{equation}
d\boldsymbol{X}_{\tau}^Q = - \nabla_{\boldsymbol{x}} \tilde{f}_{\tau}(\boldsymbol{X}_{\tau}) d\tau + b^{-\bar{p}(\tau)} d\boldsymbol{B}_{\tau}
\label{ack-ch03-eq08}
\end{equation}
, where $\boldsymbol{B}_{\tau} \in \mathbb{R}^d$ is a standard Wiener process such that $\boldsymbol{B}_{\tau} \sim \mathcal{N}(\boldsymbol{z}; 0, \boldsymbol{I}_d)$.

Although we can define the transformation to the Gaussian Wiener process to obtain the SDE on the mini-batch indexed time system, the proposed algorithm does not show an advantage in comparison to the conventional SGLD. 
While we utilize the quantization error as a seed to generate the Gaussian random variable in the proposed algorithm, the conventional SGLD deploys the Gaussian random generator for the learning system. 
Hence, there is not any practical difference in the implementation of the algorithm, and the complexity of the implementation only increases.

%=====================================================
\subsection{Analysis based on Central Limit Theorem}
%=====================================================
The key idea is the simplest method to transform the uniformly distributed quantization error into a standard Wiener process.
For this purpose, we establish the sum of the learning equation within a unit epoch regarding the mini-batch-based time index based on the double-sided time index system as follows:
\begin{equation}
\begin{aligned}
\boldsymbol{X}_{t_e + B/B}^Q 
&= \boldsymbol{X}_{t_e + (B - 1)/B}^Q - \lambda \nabla_{\boldsymbol{x}} \tilde{f}_{B-1}(\boldsymbol{X}_{t_e+ (B - 1)/B}) + Q_p^{-1}(t_e) \boldsymbol{\varepsilon}_{t_e + (B - 1)/B}^q \\
\boldsymbol{X}_{t_e + (B - 1)/B}^Q 
&= \boldsymbol{X}_{t_e + (B - 2)/B}^Q - \lambda \nabla_{\boldsymbol{x}} \tilde{f}_{B-2}(\boldsymbol{X}_{t_e + (B - 2)/B}) + Q_p^{-1}(t_e) \boldsymbol{\varepsilon}_{t_e + (B - 2)/B}^q \\
&\cdots\\
\boldsymbol{X}_{t_e + 2/B}^Q 
&= \boldsymbol{X}_{t_e + 1/B}^Q - \lambda \nabla_{\boldsymbol{x}} \tilde{f}_{1}(\boldsymbol{X}_{t_e + 1/B}) + Q_p^{-1}(t_e) \boldsymbol{\varepsilon}_{t_e + 1/B}^q \\
\boldsymbol{X}_{t_e + 1/B}^Q 
&= \boldsymbol{X}_{t_e}^Q - \lambda \nabla_{\boldsymbol{x}} \tilde{f}_{0}(\boldsymbol{X}_{t_e}) + Q_p^{-1}(t_e) \boldsymbol{\varepsilon}_{t_e}^q.
\end{aligned}
\label{ack-ch03-eq09}
\end{equation}

Adding up each term in \eqref{ack-ch03-eq09} gives
\begin{equation}
\boldsymbol{X}_{t_e + 1}^Q 
= \boldsymbol{X}_{t_e}^Q - \lambda \sum_{\tau=0}^{B-1} \nabla_{\boldsymbol{x}} \tilde{f}_{\tau}(\boldsymbol{X}_{t_e+\tau/B}) + Q_p^{-1}(t_e)\sum_{\tau=0}^{B-1} \boldsymbol{\varepsilon}_{t_e+\tau/B}^q
%\label{ch04-it-eq02}
\label{ack-ch03-eq10}
\end{equation}
, where we set the initial time index to be $t_e = \tau$ and maintain the quantization parameter to be an equal value such that $Q_p^{-1}(t_e + \tau) = Q_p^{-1}(t_e), \forall \tau \in \mathbb{Z}[0, B)$. 

According to Assumption \ref{assum02} and Assumption \ref{assum03}, the quantization error $\{\boldsymbol{\epsilon}_{t_e+\tau/B}^q \}_{\tau=0}^{B-1} =  \{Q_p^{-1}(t_e) \boldsymbol{\varepsilon}_{t_e+\tau/B}^q \}_{\tau=0}^{B-1}$ are independent random variables of the equal uniform distributions $\mathcal{U}(\boldsymbol{\varepsilon}_{t_e+\tau/B}^q; 0, c_0 Q_p^{-2}(t_e) \, \boldsymbol{I}_{d})$. 
Therefore, we can apply the Lindeberg–Lévy central limit theorem(CLT) to \eqref{ack-ch03-eq10}, and we obtain a stochastic difference equation based on a standard Wiener process.

Let the average factor of quantization $S_B \in \mathbb{R}^d$ such that $S_B = 1/B \cdot \sum_{\tau=0}^{B-1} \boldsymbol{\varepsilon}_{t_e+\tau/B}^q = \lambda \cdot \sum_{\tau=0}^{B-1} \boldsymbol{\varepsilon}_{t_e+\tau/B}^q$.
Based on the expectation and the variance of $\boldsymbol{\varepsilon}_{t_e+\tau/B}^q$, we can obtain the expectation and variance straightforwardly as follows:
\begin{equation}
\begin{aligned}
&\mathbb{E} S_B = \frac{1}{B} \sum_{\tau=0}^{B-1} \mathbb{E} \boldsymbol{\varepsilon}_{t_e+\tau/B}^q = 0, \\
&cov \, S_B = \mathbb{E} S_B \otimes S_B = \frac{1}{B^2} \mathbb{E} \sum_{\tau=0}^{B-1} \left( \boldsymbol{\varepsilon}_{t_e+\tau/B}^q \otimes \sum_{\tau=0}^{B-1} \boldsymbol{\varepsilon}_{t_e+\tau/B}^q \right) = \frac{B \, c_0}{B^2} \cdot \boldsymbol{I}_d = \lambda \, c_0 \, \boldsymbol{I}_d
\end{aligned}
\end{equation}
Let the i-th component of $S_B$ denote $S_B^{(i)}$. 
According to the Lindeberg–Lévy CLT, we can obtain the random variable $S_B^{(i)}$ following a normal distribution such that
\begin{equation}
    \frac{1}{\sqrt{\lambda \, c_0 }} S_B^{(i)} = \frac{\sqrt{B}}{\sqrt{c_0}} S_B^{(i)} \sim \mathcal{N}(0, 1)
\end{equation}
Therefore, if we define the quantization parameter appropriately, we can establish a standard wiener process as the summation of the quantization error for a unit epoch. 

Since the i th component of $1/\sqrt{\lambda \, c_0} \, S_B^{(i)}$ follows a normal distribution, we establish the quantization parameter to extract a vector-valued Gaussian random variable from the summation of the quantization error as follows:
\begin{equation}
\begin{aligned}
&\frac{1}{\sqrt{\lambda \, c_0 }} S_B^{(i)} 
= \frac{1}{\sqrt{\lambda \, c_0 }\, B} \sum_{\tau=0}^{B-1} \boldsymbol{\varepsilon}_{t_e + \tau/B}^{(i)} 
= \sqrt{\frac{\lambda}{c_0}} \sum_{\tau=0}^{B-1} \boldsymbol{\varepsilon}_{t_e + \tau/B}^{(i)}
\triangleq \boldsymbol{z}_{t_e + \tau/B}^{(i)} \in \mathbb{R}, \quad \boldsymbol{z}_{t_e + \tau/B}^{(i)} \sim \mathcal{N}(0, 1) \\
&\implies \sum_{\tau=0}^{B-1} \boldsymbol{\varepsilon}_{t_e + \tau/B}^{(i)} 
= \sqrt{\frac{c_0}{\lambda}} \boldsymbol{z}_{t_e + \tau/B}^{(i)} \\
&\implies Q_p^{-1}(t_e) \sum_{\tau=0}^{B-1} \boldsymbol{\varepsilon}_{t_e + \tau/B}^{(i)} 
= Q_p^{-1}(t_e) \sqrt{\frac{c_0}{\lambda}} \boldsymbol{z}_{t_e + \tau/B}^{(i)} 
= \eta^{-1} b^{- \bar{p}(t_e)} \sqrt{\frac{c_0}{\lambda}} \boldsymbol{z}_{t_e + \tau/B}^{(i)}
\end{aligned}
\end{equation}
, where $\boldsymbol{z}^{(i)}$ denotes the i th component of $\boldsymbol{z} \in \mathbb{R}^d$, which denotes a vector-valued Gaussian random variable. 

Consequently, if we let $\eta \triangleq \frac{1}{\lambda}\sqrt{c_0/C_q}$, where $C_q \in \mathbb{R}^+$ is a positive constant value.
Applying the auxiliary parameter of the quantization parameter $\eta$ to \eqref{ack-ch03-eq10}, we can obtain the following stochastic difference equation: 
\begin{equation}
\begin{aligned}
\boldsymbol{X}_{t_e + 1}^Q 
&= \boldsymbol{X}_{t_e}^Q - \lambda \sum_{\tau=0}^{B-1} \nabla_{\boldsymbol{x}} \tilde{f}_{\tau}(\boldsymbol{X}_{t_e+\tau/B}) + b^{-\bar{p}(t_e)} \lambda \sqrt{\frac{C_q}{c_0}} \sum_{\tau=0}^{B-1} \boldsymbol{\varepsilon}_{t_e+\tau/B}^q \\
&= \boldsymbol{X}_{t_e}^Q - \lambda \sum_{\tau=0}^{B-1} \nabla_{\boldsymbol{x}} \tilde{f}_{\tau}(\boldsymbol{X}_{t_e+\tau/B}) + \sqrt{C_q}b^{-\bar{p}(t_e)} \sqrt{\lambda} \boldsymbol{z}_{t_e+\tau/B}
\end{aligned}
%\label{ch04-it-eq02}
\label{ack-ch03-eq11}
\end{equation}
, where $\{\boldsymbol{z}_{t_e + \tau/B} \}_{\tau=0}^{B-1}$ is a vector valued standard Wiener process such that  $\boldsymbol{z}_{\tau} \sim \mathcal{N}(0, \boldsymbol{I}_d), \; \forall \tau \in \mathbb{Z}^+$.

The proposed CLT-based QSGLD reveals the practical advantage that it does not require a random number generator.
The learning process itself generates a Gaussian random number based on the uniformly distributed quantization error.
From the practical perspective, at least after 50 units of mini-batch iterations, we can regard the proposed learning algorithm as operating in cooperation with a standard Wiener process. 

%%%%%%%%%%%%%%%%%%%%%%%%%%%%%%%%%%%%%%%%%%%%%%%%%%%%%%%%%%%%%%%%%%%%%%%
\section{Convergence Property of QSGLD}  % Theorem 3
%%%%%%%%%%%%%%%%%%%%%%%%%%%%%%%%%%%%%%%%%%%%%%%%%%%%%%%%%%%%%%%%%%%%%%%
We use the following lemma to prove the theorems.
%============================================================
\begin{lemma-aux}
\label{eq01:lemma}
%============================================================
For all $x \in \mathbf{R}$,
\begin{equation}
    ( 1 - x) \leq \exp (-x).
\label{eq02:lemma}
\end{equation}
\end{lemma-aux}
%============================================================
\begin{proof}
By the definition of the exponent, we write the exponential function as the following fundamental series : 
\begin{equation}
    \exp (-x) = \sum_{n=0}^{\infty} \frac{1}{n!} (-1)^n x^n  = \sum_{k=0}^{\infty} \left( \frac{1}{2k!} x^{2k} - \frac{1}{(2k+1)!} x^{2k+1}\right).
\end{equation}
Let $u_k$ as follows:
\begin{equation}
    u_k = \frac{1}{2k!} x^{2k} \left( 1 - \frac{1}{2k+1} x \right)
\end{equation}
then we can rewrite the series of exponents such that 
\begin{equation}
    \exp(-x) = u_0 + \sum_{k=1}^{\infty} u_k.
\end{equation}
For all $k > 0$, since each $u_k$ is positive, we have
\begin{equation}
    1 - x = u_0  \leq u_0 + \sum_{k=0}^{\infty} u_k.
\end{equation}
Alternatively, we can prove the lemma with differentiation.
Let $g(x) = (1 - x) - \exp(-x)$. Differentiating $g(x)$ to $x$, we get
\begin{equation}
    \frac{d g}{dx}(x) = -1 + exp(-x), \; \frac{d^2 g}{dx^2} = - \exp(-x)
\end{equation}
We note that $g(x)$ is a concave function from the fact that $\frac{d^2 g}{dx^2} < 0, \;  \forall x \in \mathbf{R}$. In addition, the maximum of $g(x)$ is zero at $x=0$ from which $\frac{d g}{dx}(x) = -1 + exp(-x) = 0$.
Therefore, $g(x) \leq 0$, so that it fulfills the Lemma. 
\end{proof}

%=====================================================
\subsection{Weak Convergence without Convex Assumption}
%=====================================================
We rewrite the QSGLD represented in \eqref{ack-ch03-eq11} with an increment of a vector-valued standard Wiener process $\Delta \boldsymbol{B}_{\tau} \in \mathbb{R}^d$ as follows:
\begin{equation}
\boldsymbol{X}_{t_e + 1}^Q 
= \boldsymbol{X}_{t_e}^Q - \lambda \sum_{\tau=0}^{B-1} \nabla_{\boldsymbol{x}} \tilde{f}_{\tau}(\boldsymbol{X}_{t_e+\tau/B}) + \sqrt{C_q}b^{-\bar{p}(t_e)} \sum_{\tau=0}^{B-1} \Delta \boldsymbol{B}_{t_e+\tau/B}
\label{ack-ch04-eq01}
\end{equation}
, where we substitute $\sqrt{\lambda} \boldsymbol{z}_{t_e+\tau/B}$ into $\sum_{\tau=0}^{B-1} \Delta \boldsymbol{B}_{t_e+\tau/B}$, since the variance of the increments is the time increment $\lambda$ such that $\mathbb{E} (\Delta B_{\tau})^2 = \lambda, \; \forall \Delta B_{\tau} \in \mathbb{R}$ whereas $\boldsymbol{z} \sim \mathcal{N}(0, 1)$.

In \eqref{ack-ch04-eq01}, we consider the learning rate $\lambda$ as an increment of time $t_e$ which is the numbers of a mini-batch.
Thus, if the learning rate $\lambda$ monotonically decreases to 0 such that $\lambda \downarrow 0$, we can obtain the following stochastic integration intuitively. 
\begin{equation}
\boldsymbol{X}_{t_e + 1}^Q 
= \boldsymbol{X}_{t_e}^Q - \int_{t_e}^{t_e + 1} \nabla_{\boldsymbol{x}} \tilde{f}_{s}(\boldsymbol{X}_{s}) ds + \sqrt{C_q}b^{-\bar{p}(t_e)} \int_{t_e}^{t_e + 1} d\boldsymbol{B}_{t_e+\tau/B}
\label{ack-ch04-eq02}
\end{equation}

Differentiate both terms with respect to $t = t_e + 1$, we get
\begin{equation}
d\boldsymbol{X}_{t}^Q 
= - \nabla_{\boldsymbol{x}} \tilde{f}_{t}(\boldsymbol{X}_{t}) dt + \sqrt{C_q}b^{-\bar{p}(t)}  d\boldsymbol{B}_{t}.
\label{ack-ch04-eq03}
\end{equation}
This intuitive deduction looks like a verification of the SDE approximation of the discrete stochastic difference equation from \eqref{ack-ch03-eq11} for QSGLD.

However, such a deduction does not provide any evidence of tightness regarding the state $\boldsymbol{X}_{t}^Q$, so we should provide more rigorous evidence of the SDE approximation.
For the rigorous proof of the SDE approximation, we investigated two approaches.
The one is weak convergence of a stochastic difference equation to corresponding SDE.
Unfortunately, the weak convergence criterion provided by \citet{Kushner-1974} requires the monotonically decreasing of the learning rate $\lambda \downarrow 0$ to time index increases, i.e., $t \uparrow \infty$
Some of the weak convergence criteria cannot be satisfied without such a monotonically decreasing learning rate, despite the limited drift and diffusion terms.

The other approach is a weak approximation of SDE. 
This approach is based on the numerical analysis of SDE approximation. 
The fundamental concept of the approach is that if the statistical quantities between a discrete stochastic difference equation and approximated continuous SDE are equivalent, we consider the approximation to be well-defined. 

In this paper, we provide the SDE approximation regarding the proposed quantization-based learning algorithm as following lemma. 

%====================================
%\begin{lemma} % theorem for weak convergence 
%\label{lemma_01}
\textbf{Lemma 3.2}
%====================================
The approximated Langvin SDE for QSGLD represented in \eqref{ch04-it-eq02} is as follows: 
\begin{equation}
d\boldsymbol{X}_t = - \nabla_{\boldsymbol{x}} f(\boldsymbol{X}_t) dt + \sqrt{C_q} \cdot \sigma(t) d\boldsymbol{B}_t, \quad \forall t > t_0 \in \displaystyle \R^+, \; \because \sigma(t) \triangleq b^{-\bar{p}(t)},
\label{sup_lemma01_eq01}
\end{equation}
The approximation $\eqref{sup_lemma01_eq01}$ satisfies the order-1 weak approximation described in Definition $\ref{def_04}$.
\begin{comment}
If the quantization parameter is given as $Q_p(t) = \frac{1}{\lambda}\, \sqrt{\frac{c_0}{C_q}} \, b^{\bar{p}(t)}$, the QSGLD represented as \eqref{ch04-it-eq02} weakly converges to Langevin SDE with respect to $\boldsymbol{X}_t \in \mathbb{R}^d$ as follows:
\begin{equation}
d\boldsymbol{X}_t = - \nabla_{\boldsymbol{x}} f(\boldsymbol{X}_t) dt + \sqrt{C_q} \cdot \sigma(t) d\boldsymbol{B}_t, \quad \forall t > t_0 \in \displaystyle \R^+, \; \because \sigma(t) \triangleq b^{-\bar{p}(t)},
%\label{lemma01_eq01}
\label{ack-ch04-eq01}
\end{equation}
, where $\boldsymbol{B}_t$ denotes a standard vector valued Wiener process, $C_q \in \displaystyle \R^+$ denotes a constant value, and $\lambda$ denotes the reciprocal of the size of mini-batch $B$, 
%\end{lemma}
\end{comment}
%------------------------------------
\begin{proof}
\textbf{Preparation}
Let the transition probability from $t=\tau$ to $t+1=\tau+\lambda$ such that $p(t, \boldsymbol{X}_t^Q, t+1, \boldsymbol{X}_{t+1}^Q)$. 
We define the following one-step changes of QSLD/QSLGD from \eqref{ch04-eq03} and \eqref{ch04-eq04} in the main manuscript such that
\begin{equation}
\Delta (\boldsymbol{X}_t) \triangleq \lambda \, h(\boldsymbol{X}_{\tau}^Q), \quad 
\tilde{\Delta} (\boldsymbol{X}_t) \triangleq \boldsymbol{X}_{t+1} - \boldsymbol{X}_{\tau}^Q, \quad \Delta, \tilde{\Delta} \in \mathbf{R}^d
\end{equation}
, where $\boldsymbol{X}_{t+1}^Q \sim p(t, \boldsymbol{X}_t^Q, t+1, \boldsymbol{X}_{t+1}^Q)$, and $\boldsymbol{X}_t^Q = \boldsymbol{X}_{\tau}^Q$.

By proving the following conditions presented in the theorems of SDE approximation to the SGD equation as proposed by \citet{Li-2019-JMLR} and \citet{Malladi_NEURIPS2022}, we validate the applicability of the SDE approximation to the proposed equation.

\textbf{Lipschitz continuity and continuous differentiability of the drift and diffusion functions}
Assumption \ref{assum01} in the manuscript provides that the drift $\nabla f(\boldsymbol{X}_{t})$ is Lipschitz continuity.
As the diffusion function in QSLGD is the quantization parameter, it is Lipschitz continuity by Definition $\ref{def_q02}$ in the manuscript. 

\textbf{Bounded moments condition} The first order bounded moments condition is given by:
\begin{equation}
\left\vert \mathbb{E}(\Delta_i(\boldsymbol{X}_{t}) - \tilde{\Delta}_i(\boldsymbol{X}_{t}) \right\vert
\leq K (\boldsymbol{X}_{t}) \lambda^2 
\label{lmpf_eq01}
\end{equation}

From the definition of $\Delta, \, \tilde{\Delta}$, for the initial time $t = \tau/N_B$, we get
\begin{equation}
\begin{aligned}
&\Delta(\boldsymbol{X}_{t}) - \tilde{\Delta}(\boldsymbol{X}_{t}) 
= \lambda h(\boldsymbol{X}_t) - \lambda h(\boldsymbol{X}_t) - Q_p^{-1}(t) \boldsymbol{\epsilon}_t^q  = - Q_p^{-1}(t) \boldsymbol{\epsilon}_t^q \\
&\Rightarrow \mathbb{E} Q_p^{-1}(t) \boldsymbol{\epsilon}_t^q = Q_p^{-1}(t) \mathbb{E} \boldsymbol{\epsilon}_t^q = 0.
\end{aligned}
\label{lmpf_eq02}
\end{equation}
The \eqref{lmpf_eq02} represents that the proposed algorithm fulfills the first-order bounded moment condition.

For the bound of the second moment, we calculate the moment as follows:
\begin{equation}
\begin{aligned}
\Delta_i(\boldsymbol{X}_{t}) \Delta_j(\boldsymbol{X}_{t}) 
&= \lambda^2 h_i (\boldsymbol{X}_{t}) h_j (\boldsymbol{X}_{t})  \\
\tilde{\Delta}_i(\boldsymbol{X}_{t}) \tilde{\Delta}_j(\boldsymbol{X}_{t}) 
&= (\lambda h_i (\boldsymbol{X}_{t}) + Q_p^{-1}(t) {\epsilon_i}_t^q) (\lambda h_j (\boldsymbol{X}_{t}) + Q_p^{-1}(t) {\epsilon_j}_t^q) \\
&= \lambda^2 h_i (\boldsymbol{X}_{t}) h_j (\boldsymbol{X}_{t}) + \lambda (h_j (\boldsymbol{X}_{t}) Q_p^{-1}(t) {\epsilon_j}_t^q \\
&+ h_i (\boldsymbol{X}_{t})  Q_p^{-1}(t) {\epsilon_i}_t^q + Q_p^{-2}(t) {\epsilon_i}_t^q {\epsilon_j}_t^q. 
\end{aligned}
\label{lmpf_eq03}
\end{equation}

The \eqref{lmpf_eq03} implies that 
\begin{equation}
\begin{aligned}
&\Delta_i(\boldsymbol{X}_{t}) \Delta_j(\boldsymbol{X}_{t}) - \tilde{\Delta}_i(\boldsymbol{X}_{t}) \tilde{\Delta}_j(\boldsymbol{X}_{t}) \\
&= -\lambda (h_j (\boldsymbol{X}_{t}) Q_p^{-1}(t) {\epsilon_j}_t^q + h_i (\boldsymbol{X}_{t})  Q_p^{-1}(t) {\epsilon_i}_t^q )- Q_p^{-2}(t) {\epsilon_i}_t^q {\epsilon_j}_t^q. \\
&\Rightarrow 
\mathbb{E}(\Delta_i(\boldsymbol{X}_{t}) \Delta_j(\boldsymbol{X}_{t}) - \tilde{\Delta}_i(\boldsymbol{X}_{t})  \tilde{\Delta}_j(\boldsymbol{X}_{t}) ) \\
&= \lambda (h_j (\boldsymbol{X}_{t}) Q_p^{-1}(t) \mathbb{E} {\epsilon_j}_t^q + h_i (\boldsymbol{X}_{t})  Q_p^{-1}(t) \mathbb{E}{\epsilon_i}_t^q ) 
- Q_p^{-2}(t) \mathbb{E} {\epsilon_i}_t^q {\epsilon_j}_t^q 
= -\frac{1}{12} Q_p^{-2}(t) \delta(i-j).
\end{aligned}
\label{lmpf_eq04}
\end{equation}
, where $i, j \in \mathbf{N}[1, d]$ denotes the index of the vector components.

Thus, we can choose $t > t_0$ such that 
\begin{equation}
\left\vert \Delta_i(\boldsymbol{X}_{t}) \Delta_j(\boldsymbol{X}_{t}) - \tilde{\Delta}_i(\boldsymbol{X}_{t}) \tilde{\Delta}_j(\boldsymbol{X}_{t}) \right\vert = \frac{1}{12} Q_p^{-2}(t) \delta(i-j) \leq K_1 (\boldsymbol{X}_{t}) \lambda^2
\label{lmpf_eq05}
\end{equation}
, where $K_1$ denotes a secondary differentiable function such that $K_1 : \mathbf{R}^d \rightarrow \mathbf{R}, \; K_1 \in C^2$.

Similarly, we calculate the third moment as follows:
\begin{equation}
\begin{aligned}
\Delta_i(\boldsymbol{X}_{t}) \Delta_j(\boldsymbol{X}_{t}) \Delta_k(\boldsymbol{X}_{t}) 
= \lambda^3  h_i (\boldsymbol{X}_{t}) h_j (\boldsymbol{X}_{t}) h_k (\boldsymbol{X}_{t})
\end{aligned}
\label{lmpf_eq06}
\end{equation}
, and 
\begin{equation}
\begin{aligned}
&\tilde{\Delta}_i(\boldsymbol{X}_{t}) \tilde{\Delta}_j(\boldsymbol{X}_{t}) \tilde{\Delta}_k(\boldsymbol{X}_{t}) 
= \lambda^3  h_i (\boldsymbol{X}_{t}) h_j (\boldsymbol{X}_{t}) h_k (\boldsymbol{X}_{t}) \\
&+ \lambda^2 Q_p^{-1}(t) \left( h_i (\boldsymbol{X}_{t}) h_j (\boldsymbol{X}_{t}) {\epsilon_k}_t^q 
+ h_j (\boldsymbol{X}_{t}) h_k (\boldsymbol{X}_{t}) {\epsilon_i}_t^q
+ h_k (\boldsymbol{X}_{t}) h_i (\boldsymbol{X}_{t}) {\epsilon_j}_t^q \right) \\
&+ \lambda Q_p^{-2}(t) \left( h_i (\boldsymbol{X}_{t}) {\epsilon_j}_t^q {\epsilon_k}_t^q  
+ h_j (\boldsymbol{X}_{t}) {\epsilon_k}_t^q {\epsilon_i}_t^q
+ h_k (\boldsymbol{X}_{t}) {\epsilon_i}_t^q {\epsilon_j}_t^q \right)
+ Q_p^{-3}(t) {\epsilon_i}_t^q {\epsilon_j}_t^q {\epsilon_k}_t^q.
\end{aligned}
\label{lmpf_eq07}
\end{equation}

Based on the white noise hypothesis(WNH) of the quantization error, we can determine the expectation value of \eqref{lmpf_eq07} such that
\begin{equation}
\begin{aligned}
&\mathbb{E}\tilde{\Delta}_i(\boldsymbol{X}_{t}) \tilde{\Delta}_j(\boldsymbol{X}_{t}) \tilde{\Delta}_k(\boldsymbol{X}_{t}) 
= \lambda^3  h_i (\boldsymbol{X}_{t}) h_j (\boldsymbol{X}_{t}) h_k (\boldsymbol{X}_{t}) \\
&+ \frac{1}{12}\lambda Q_p^{-2}(t) \left( h_i (\boldsymbol{X}_{t}) \delta(j-k)
+ h_j (\boldsymbol{X}_{t}) \delta(k-i) + h_k (\boldsymbol{X}_{t}) \delta(i-j) \right) \\
&= \lambda^3  h_i (\boldsymbol{X}_{t}) h_j (\boldsymbol{X}_{t}) h_k (\boldsymbol{X}_{t}) + \frac{1}{4} \lambda Q_p^{-2}(t) h_i (\boldsymbol{X}_{t}) \vert_{i=j=k}.
\end{aligned}
\label{lmpf_eq08}
\end{equation}
By \eqref{lmpf_eq08} and \eqref{lmpf_eq06}, we get 
\begin{equation}
\begin{aligned}
&\mathbb{E}\tilde{\Delta}_i(\boldsymbol{X}_{t}) \tilde{\Delta}_j(\boldsymbol{X}_{t}) \tilde{\Delta}_k(\boldsymbol{X}_{t}) 
- \mathbb{E} \Delta_i(\boldsymbol{X}_{t}) \Delta_j(\boldsymbol{X}_{t}) \Delta_k(\boldsymbol{X}_{t}) \\
&= \lambda^3  h_i (\boldsymbol{X}_{t}) h_j (\boldsymbol{X}_{t}) h_k (\boldsymbol{X}_{t}) + \frac{1}{4} \lambda Q_p^{-2}(t) h_i (\boldsymbol{X}_{t}) \vert_{i=j=k} 
- \lambda^3  h_i (\boldsymbol{X}_{t}) h_j (\boldsymbol{X}_{t}) h_k (\boldsymbol{X}_{t}) \\
&= \frac{1}{4} \lambda Q_p^{-2}(t) h_i (\boldsymbol{X}_{t}) \vert_{i=j=k} 
\end{aligned}
\label{lmpf_eq09}
\end{equation}

According to Assumption \ref{assum01}, there exists an optimal point $\boldsymbol{X}^* \in \mathcal{D} \subset B^o(\boldsymbol{X}^*, \rho)$ such that $\nabla f (\boldsymbol{X}^*) = 0$ allowing us to obtain 
\begin{equation}
\begin{aligned}
&\vert \mathbb{E}\tilde{\Delta}_i(\boldsymbol{X}_{t}) \tilde{\Delta}_j(\boldsymbol{X}_{t}) \tilde{\Delta}_k(\boldsymbol{X}_{t}) 
- \mathbb{E} \Delta_i(\boldsymbol{X}_{t}) \Delta_j(\boldsymbol{X}_{t}) \Delta_k(\boldsymbol{X}_{t}) \vert \\
&= \frac{1}{4} \lambda Q_p^{-2}(t) \vert h_i (\boldsymbol{X}_{t}) \vert_{i=j=k} 
= \frac{1}{4} \lambda Q_p^{-2}(t) \vert  h_i (\boldsymbol{X}_{t}) + \nabla f_i (\boldsymbol{X}^*) \vert_{i=j=k}. 
\end{aligned}
\label{lmpf_eq10}
\end{equation}

In QSLGD, since $h(\boldsymbol{X}_t)$ is equal to $-\nabla f(\boldsymbol{X}_t)$, we apply the $L_1$ Lipschitz condition to calculate the bound as follows:
\begin{equation}
\begin{aligned}
\vert \mathbb{E}\tilde{\Delta}_i(\boldsymbol{X}_{t}) \tilde{\Delta}_j(\boldsymbol{X}_{t}) \tilde{\Delta}_k(\boldsymbol{X}_{t}) 
- \mathbb{E} \Delta_i(\boldsymbol{X}_{t}) \Delta_j(\boldsymbol{X}_{t}) \Delta_k(\boldsymbol{X}_{t}) \vert 
&= \frac{1}{4} \lambda Q_p^{-2}(t) \vert  h_i (\boldsymbol{X}_{t}) + \nabla f_i (\boldsymbol{X}^*) \vert_{i=j=k} \\
&= \frac{1}{4} \lambda Q_p^{-2}(t) \vert  \nabla f_i (\boldsymbol{X}^*) - \nabla f_i (\boldsymbol{X}_{t}) \vert_{i=j=k} \\
&\leq \frac{1}{4} \lambda Q_p^{-2}(t) L_1 \vert  \boldsymbol{X}^* - \boldsymbol{X}_{t} \vert_{i=j=k} \\
&\leq \frac{1}{4} \lambda Q_p^{-2}(t) L_1 \rho \\
\end{aligned}
\label{lmpf_eq11}
\end{equation}
By selecting $t > t_0$ such that $Q_p(t)^{-2} < \lambda$, the bound condition of the third-order moment is satisfied, resulting in
\begin{equation}
\begin{aligned}
\vert \mathbb{E}\tilde{\Delta}_i(\boldsymbol{X}_{t}) \tilde{\Delta}_j(\boldsymbol{X}_{t}) \tilde{\Delta}_k(\boldsymbol{X}_{t}) 
- \mathbb{E} \Delta_i(\boldsymbol{X}_{t}) \Delta_j(\boldsymbol{X}_{t}) \Delta_k(\boldsymbol{X}_{t}) \vert 
\leq \frac{1}{4} \lambda Q_p^{-2}(t) L_1 \rho < \frac{1}{4} \lambda^2  L_1 = K_1(\boldsymbol{X}_t) \lambda^2
\end{aligned}
\label{lmpf_eq12}
\end{equation}

\textbf{Adaptation Lemma by \citet{Li-2019-JMLR} and \citet{Malladi_NEURIPS2022}}
Under the verification of the bounded momentum condition \eqref{lmpf_eq02}, \eqref{lmpf_eq04}, \eqref{lmpf_eq10}, we can establish the  following polynomial growth condition:
There exists a subset $P$ of the index set $\triangleq \{1, 2, \cdots, d\}$ such that the following holds. Below we use the notations $\| \boldsymbol{x}_P \| \triangleq \sqrt{\sum_{i \in P} x_i^2}$ and $\| \boldsymbol{x}_R \| \triangleq \sqrt{\sum_{i \notin P} x_i^2}$.
\begin{enumerate}
    \item There is a constant $C_1 > 0$, which is independent of $\lambda$, so that for all $k \leq N$
    \begin{equation}
    \begin{aligned}
        \| \mathbb{E} \Delta(\boldsymbol{X}_{\tau}) \|_P &\leq C_{1} \lambda (1 + \| \boldsymbol{X}_{\tau} \|_P )\\
        \| \mathbb{E} \Delta(\boldsymbol{X}_{\tau}) \|_R &\leq C_{1} \lambda (1 + \| \boldsymbol{X}_{\tau} \|_P^{w_1})(1 + \| \boldsymbol{X}_{\tau} \|_R ) 
    \end{aligned}
    \end{equation}    
    \item For all $m \geq 1$, there are constant $C_{2m}, w_{2m} > 0$ (independent of $\lambda$) such that for all $k < N$
    \begin{equation}
    \begin{aligned}
        \| \mathbb{E} \Delta(\boldsymbol{X}_{\tau}) \|_P^{2m} &\leq C_{1} \lambda (1 + \| \boldsymbol{X}_{\tau} \|_P^{2m} )\\
        \| \mathbb{E} \Delta(\boldsymbol{X}_{\tau}) \|_R &\leq C_{1} \lambda (1 + \| \boldsymbol{X}_{\tau} \|_P^{w_{2m}})(1 + \| \boldsymbol{X}_{\tau} \|_R^{2m} ) 
    \end{aligned}
    \end{equation}    
\end{enumerate}
The analysis shows that the function $K_1$ is differentiable to the first order, allowing for the bound on the third moment. Consequently, a function $g$ twice differentiable ($g \in C^2$) is enough to achieve a first-order weak approximation for QSLGD.

\textbf{Adaptation Theorem 3 in by \citet{Li-2019-JMLR} \citet{Malladi_NEURIPS2022}}
Since QSLGD satisfies the Lipschitz continuity, the bounded moment conditions, and the adaptation lemma, QSLGD fulfills a first-order weak approximation defined as Definition \ref{def_04} in the manuscript, by the theorem provided by Theorem 3 in by \citet{Li_NEURIPS2022}, as follows: 
For each function $g \in C^2$, there exists a constant $C > 0$ (independent of $\lambda$) such that
\begin{equation}
    \max_{0 \leq k \leq N} \vert \mathbb{E} g(\boldsymbol{X}_{\tau}^Q) - \mathbb{E} g(\boldsymbol{X}_t) \vert \leq C \lambda 
\end{equation}

\end{proof}
%------------------------------------
%====================================
%\begin{theorem} % theorem for weak convergence 
%\label{theorem_02}
\textbf{Theorem 3.3}
%====================================
Consider the transition probability, denoted as $p(t, \boldsymbol{X}_t, t+\bar{\tau}, \boldsymbol{x}^*)$, from an arbitrary state $\boldsymbol{X}_{t} \in \mathbb{R}^d$ to the optimal point $\boldsymbol{x}^* \in \mathbb{R}^d, \; \boldsymbol{X}_t \neq \boldsymbol{x}^*$ after a time interval $\bar{\tau} \in \displaystyle \R^+$, for all $t > t_0$. If the quantization parameter is bounded as follows:
\begin{equation}
\sup_{t \geq 0} Q_p(t) = \frac{1}{C} \cdot{\log (t + 2)}, \quad C \in \displaystyle \R^{++},
\end{equation}
for $\boldsymbol{X}_t \neq \boldsymbol{\bar{X}}_t$, QSGLD represented as \eqref{sup_lemma01_eq01} converges with distribution in the sense of Cauchy convergence such that
\begin{equation}
\overline{\lim_{\bar{\tau} \rightarrow \infty}} \sup_{\boldsymbol{X}_t, \bar{\boldsymbol{X}}_{t} \in \displaystyle \R^n} \| p(t, \boldsymbol{\bar{X}}_t, t + \bar{\tau},  \boldsymbol{x}^*) - p(t, \boldsymbol{X}_t, t + \bar{\tau},  \boldsymbol{x}^*) \| \leq \tilde{C} \cdot \exp \left(-\sum_{\bar{\tau}=0}^{\infty} \delta_{t+\bar{\tau}} \right)
\end{equation}
, where $\delta_{t}$ denotes the infimum of the transition probability from time $t$ to $t+1$ given by $\delta_t = \inf_{x, y \in \displaystyle \R^d} p(t, x, t+1, y)$, satisfying $\sum_{\bar{\tau}=0}^{\infty} \delta_{t+\bar{\tau}} = \infty$, and $\tilde{C}$ denotes a positive value. 
%\end{theorem}
%------------------------------------
\begin{proof}
%------------------------------------
We depend on the lemmas in works of \citet{Geman-1986} to prove the theorem.
Herein, we prove the following convergence of the transition probability:
\begin{equation}
\overline{\lim_{\bar{\tau} \rightarrow \infty}} \sup_{\boldsymbol{X}_t, \bar{\boldsymbol{X}}_{t} \in \displaystyle \R^n} \| p(t, \boldsymbol{\bar{X}}_t, t + \bar{\tau},  \boldsymbol{x}^*) - p(t, \boldsymbol{X}_t, t + \bar{\tau},  \boldsymbol{x}^*) \| = 0
\label{lmpf2.7_eq01}    
\end{equation}
, where $t$ and $\tau$ denote the current time index and the process time index, respectively. $\boldsymbol{x}^* \in \mathbb{R}^d$ denotes an global optimum for the objective function $f(\boldsymbol{X}_t)$ such that $f(\boldsymbol{x}^*) < f(\boldsymbol{X}_t), \; \forall t \geq 0$.

Let the infimum of the transition probability from $t$ to $t+1$ such that
\begin{equation}
\delta_t = \inf_{\boldsymbol{x}, \boldsymbol{y} \in \mathbf{R}^d} p(t, \boldsymbol{x}, t+1, \boldsymbol{y}) 
\label{lmpf2.7_eq02}    
\end{equation}

According to the lemma in \citet{Geman-1986}, we can evaluate the upper bound of \eqref{lmpf2.7_eq01} as follows:
For state vectors $\boldsymbol{v}, \boldsymbol{w}, \boldsymbol{z}, \boldsymbol{f} \in \mathbf{R}^d$ and time indexes $s, t \in \mathbf{R}^+$,
\begin{equation}
\begin{aligned}
&\overline{\lim_{t \rightarrow \infty}} \sup_{\boldsymbol{v}, \boldsymbol{w} \in \mathbb{R}^d} \vert p(s,\boldsymbol{v}, t, \boldsymbol{f}) - p(s, \boldsymbol{w}, t, \boldsymbol{f}) \vert \\
&= \overline{\lim_{t \rightarrow \infty}} \sup_{\boldsymbol{v}, \boldsymbol{w} \in \mathbb{R}^d}
\left\vert \int p(s,\boldsymbol{v}, s+1, \boldsymbol{z})p(s+1,\boldsymbol{z}, t, \boldsymbol{f}) dz
- \int p(s, \boldsymbol{w}, s+1, \boldsymbol{z}) p(s+1, \boldsymbol{z}, t, \boldsymbol{f}) dz \right\vert \\
&= \overline{\lim_{t \rightarrow \infty}} \sup_{\boldsymbol{v}, \boldsymbol{w} \in \mathbb{R}^d}
\left\vert \int p(s,\boldsymbol{v}, s+1, \boldsymbol{z})  p(s+1,\boldsymbol{z}, t, \boldsymbol{f}) dz 
- \int p(s, \boldsymbol{w}, s+1, \boldsymbol{z}) p(s+1, \boldsymbol{z}, t, \boldsymbol{f}) dz \right. \\
& \left. - (\delta_s - \delta_s) p(s+1,\boldsymbol{z}; t, \boldsymbol{f}) \right\vert \\
&= \overline{\lim_{t \rightarrow \infty}} \sup_{\boldsymbol{v}, \boldsymbol{w} \in \mathbb{R}^d} 
\left\vert \int (p(s,\boldsymbol{v}; s+1, \boldsymbol{z}) - \delta_s) p(s+1,\boldsymbol{z}; t, \boldsymbol{f}) dz - \int (p(s, \boldsymbol{w}; s+1, \boldsymbol{z}) - \delta_s) p(s+1, \boldsymbol{z}; t, \boldsymbol{f}) dz \right\vert  \\
&\leq \overline{\lim_{t \rightarrow \infty}} \sup_{\boldsymbol{v}, \boldsymbol{w} \in \mathbb{R}^d}
\left\vert \int (p(s,\boldsymbol{v}; s+1, \boldsymbol{z}) - \delta_s) \sup_{\boldsymbol{z} \in \mathbb{R}^d} p(s+1,\boldsymbol{z}; t, \boldsymbol{f}) dz 
- \int (p(s, \boldsymbol{w};, s+1, \boldsymbol{z}) - \delta_s) \inf_{\boldsymbol{z} \in \mathbb{R}^d} p(s+1, \boldsymbol{z}; t, \boldsymbol{f}) dz \right\vert \\
&= \overline{\lim_{t \rightarrow \infty}} \sup_{\boldsymbol{v}, \boldsymbol{w} \in \mathbb{R}^d} 
\left\vert \sup_{\boldsymbol{z} \in \mathbb{R}^d} p(s+1,\boldsymbol{z}; t, \boldsymbol{f}) \int (p(s,\boldsymbol{v}; s+1, \boldsymbol{z}) - \delta_s) dz
- \inf_{\boldsymbol{z} \in \mathbb{R}^d} p(s+1, \boldsymbol{z}; t, \boldsymbol{f}) \int ( p(s, \boldsymbol{w}; s+1, \boldsymbol{z}) - \delta_s) dz \right\vert \\
&\leq \overline{\lim_{t \rightarrow \infty}} \sup_{\boldsymbol{v}, \boldsymbol{w} \in \mathbb{R}^d} 
\left\vert ( 1 - \delta_s) \sup_{\boldsymbol{z} \in \mathbb{R}^d} p(s+1,\boldsymbol{z}; t, \boldsymbol{f}) - (1 - \delta_s) \inf_{\boldsymbol{z} \in \mathbb{R}^d} p(s+1,\boldsymbol{z}; t, \boldsymbol{f}) \right\vert \\
&= \overline{\lim_{t \rightarrow \infty}} 
\sup_{\boldsymbol{v}, \boldsymbol{w} \in \mathbb{R}^d} ( 1 - \delta_s) \left\vert \sup_{\boldsymbol{z} \in \mathbb{R}^d} p(s+1,\boldsymbol{z}; t, \boldsymbol{f}) - \inf_{\boldsymbol{z} \in \mathbb{R}^d} p(s+1,\boldsymbol{z}; t, \boldsymbol{f}) \right\vert \\
&\cdots \\
&\leq \overline{\lim_{t \rightarrow \infty}}
\left(\prod_{k=0}^{(t-s)-1} (1 - \delta_{s+k}) \right) \cdot
\sup_{\boldsymbol{v}, \boldsymbol{w} \in \mathbb{R}^d} \left\vert p(s+(t-s),\boldsymbol{v}; t, \boldsymbol{f}) - p(s+(t-s), \boldsymbol{w}; t, \boldsymbol{f}) \right\vert \\
&\leq \overline{\lim_{t \rightarrow \infty}} \left(\prod_{k=0}^{(t-s)-1} (1 - \delta_{s+k}) \right)
= \prod_{k=0}^{\infty}(1 - \delta_{s+k}).
\end{aligned}
\end{equation}

Thus, we obtain
\begin{equation}
\overline{\lim_{\tau \rightarrow \infty}} \sup_{\boldsymbol{X}_t, \bar{\boldsymbol{X}}_{t} \in \displaystyle \R^n} \| p(t, \boldsymbol{\bar{X}}_t, t + \bar{\tau},  \boldsymbol{x}^*) - p(t, \boldsymbol{X}_t, t + \bar{\tau},  \boldsymbol{x}^*) \| \leq \prod_{k=0}^\infty (1 - \delta_{t+k}).
\label{lmpf2.7_eq03}    
\end{equation}

From the exponential approximation \eqref{eq01:lemma} in \textbf{Lemma:Auxiliary},  we rewrite \eqref{lmpf2.7_eq03} as follows:
\begin{equation}
\overline{\lim_{\tau \rightarrow \infty}} \sup_{\boldsymbol{X}_t, \bar{\boldsymbol{X}}_{t} \in \displaystyle \R^n} \| p(t, \boldsymbol{\bar{X}}_t, t + \bar{\tau},  \boldsymbol{x}^*) - p(t, \boldsymbol{X}_t, t + \bar{\tau},  \boldsymbol{x}^*) \| \leq \exp(-\sum_{k=0}^{\infty} \delta_{t+k}) ).
\label{lmpf2.7_eq04}    
\end{equation}
To verify the existence of an upper bound of the right-hand side in \eqref{lmpf2.7_eq04}, we rephrase the approximation SDE governing the dynamics of the proposed algorithm, as stated in Lemma \ref{lemma_01}:
\begin{equation}
d\boldsymbol{X}_s = - \nabla f(\boldsymbol{X}_{s}) ds + \sigma(s) \sqrt{C_q} d\boldsymbol{B}_s, \quad s \in \mathbf{R}(t, t+1).
\label{lmpf2.7_eq05}    
\end{equation}
%, where $\sigma(s) \triangleq 1/\sqrt{C_q} \, Q_p^{-1}(s)$.

Let $P_x$ be the probability measures on $\mathcal{F}$ induced by \eqref {lmpf2.7_eq05} and the probability distribution $Q_x$ given by the following equation:
\begin{equation}
d\bar{\boldsymbol{X}}_{s} = \sigma(s) \sqrt{C_q} d\boldsymbol{B}_{s}, \quad s \in \mathbf{R}(t, t+1). 
\label{lmpf2.7_eq06}    
\end{equation}
According to the Girsanov theorem (\citet{Bernt_2003, Klebaner_2011}), we obtain
%-----------------------------------
% (76) : {lmpf2.7_eq07}    
%-----------------------------------
\begin{equation}
\frac{dP_{\boldsymbol{X}}}{dQ_{\bar{\boldsymbol{X}}}}
= \exp \left\{-\int_t^{t+1} \frac{C_q^{-1}}{\sigma^2 (s)} \nabla f(\boldsymbol{X}_{s}) d\bar{\boldsymbol{X}}_{s} -\frac{1}{2} \int_t^{t+1} \frac{C_q^{-1}}{\sigma^2 (s)} \|  \nabla f(\boldsymbol{X}_{s}) \|^2 ds \right\}.
\label{lmpf2.7_eq07}    
\end{equation}

To compute the upper bound of \eqref{lmpf2.7_eq07}, we will check the upper bound of $\| \nabla f \|$.  
Considering Assumption $\ref{assum02}$, the gradient of $f(\boldsymbol{X}_t) \in C^{2}$ fulfills the Lipschitz continuous condition as well. 
Thereby, there exists a positive value $L_1 \in \mathbf{R}^+$  such that 
\begin{equation}
\| \nabla f(\boldsymbol{X}_{s}) - \nabla f(\boldsymbol{x}^*) \| \leq L_1 \| \boldsymbol{X}_{s} - \boldsymbol{x}^* \|, \quad \forall s > 0.
\label{lmpf2.7_eq09}    
\end{equation}
Successively, since $\nabla f(\boldsymbol{x}^*) = 0$, the Lipschitz condition forms simply as follows : 
\begin{equation}
\| \nabla f(\boldsymbol{X}_t) \| \leq L_1 \rho = C_0
\label{lmpf2.7_eq10}    
\end{equation}
, where $\rho = \| \boldsymbol{X}_t - \boldsymbol{x}^* \|$ from Assumption \ref{assum01}.

Additionally, holding the $L_1$ Lipschitz continuity and the assumption of which $f \in C^2$, it implies that there exists a positive value $L_2 
 \in \mathbf{R}^+$ such that 
\begin{equation}
\| \boldsymbol{H}_{\boldsymbol{x}}(f)(\boldsymbol{x}) \|_{\mathbb{R}^{d \times d}} < L_2 \text{ and }  | \Delta_{\boldsymbol{x}}^2 f (\boldsymbol{x})| < L_2 d, \quad  \forall \boldsymbol{x} \in B^o(\boldsymbol{x}, \rho)
\label{lmpf2.7_eq10-01}    
\end{equation}
, where $\boldsymbol{H}_{\boldsymbol{x}}(f) \in \mathbb{R}^{d \times d}$ denotes the Hessian of $f : \mathbb{R}^d \mapsto \mathbb{R}$, and $\Delta_{\boldsymbol{x}}^2 f \in \mathbb{R}$ denotes the Laplacian of $f$.

To evaluate the first term of the right-hand side in \eqref{lmpf2.7_eq07}, we derive the stochastic differential of $f(\boldsymbol{X}_s)$ with respect to $d\boldsymbol{X}_t$ described in \eqref{sup_lemma01_eq01} as follows:
\begin{equation}
\begin{aligned}
&df(\boldsymbol{X}_s) =  \nabla f(\boldsymbol{X}_s) \cdot d \bar{\boldsymbol{X}}_s + \frac{1}{2} C_q \sigma^2(s) \Delta f(\boldsymbol{X}_s) ds \\
&\implies 
\frac{C_q^{-1}}{\sigma^{2}(s)}  \nabla f(\boldsymbol{X}_s) \cdot d \bar{\boldsymbol{X}}_s 
= \frac{C_q^{-1}}{\sigma^{2}(s)} df(\boldsymbol{X}_s) - \frac{1}{2} \Delta f(\boldsymbol{X}_s) ds \\
&\implies 
\int_{t}^{t+1} \frac{C_q^{-1}}{\sigma^{2}(s)}  \nabla f(\boldsymbol{X}_s) \cdot d \bar{\boldsymbol{X}}_s 
= \int_{t}^{t+1} \frac{C_q^{-1}}{\sigma^{2}(s)} df(\boldsymbol{X}_s) - \frac{1}{2} \int_{t}^{t+1} \Delta f(\boldsymbol{X}_s) ds.
\end{aligned}
\label{lmpf2.7_eq10-02}    
\end{equation}

To get a feasible result for the stochastic integration, We integrate the first term of the right-hand side in \eqref{lmpf2.7_eq10-02} partially such that
\begin{equation}
\int_{t}^{t+1} \frac{C_q^{-1}}{\sigma^{2}(s)} df(\boldsymbol{X}_s)
= C_q^{-1} \left( \frac{f(\boldsymbol{X}_s)}{\sigma^2(s)} \Bigg\vert_{t}^{t+1} - \int_t^{t+1}  f(\boldsymbol{X}_s) d \left(\frac{1}{\sigma^{2}(s)} \right) \right).
\label{lmpf2.7_eq10-03}    
\end{equation}

Since \eqref{lmpf2.7_eq10-03} relies on the random variable $\boldsymbol{X}_t$, we should the upper bound of \eqref{lmpf2.7_eq10-03} for the positive result such that $\int_{t}^{t+1} \frac{C_q^{-1}}{\sigma^{2}(s)} df(\boldsymbol{X}_s) \geq 0$, and the negative result $\int_{t}^{t+1} \frac{C_q^{-1}}{\sigma^{2}(s)} df(\boldsymbol{X}_s) < 0$.
For $\int_{t}^{t+1} \frac{C_q^{-1}}{\sigma^{2}(s)} df(\boldsymbol{X}_s) \geq 0$, we can obtain 
\begin{equation}
\begin{aligned}
\Bigg\vert \int_{t}^{t+1} \frac{C_q^{-1}}{\sigma^{2}(s)} df(\boldsymbol{X}_s) \Bigg\vert
&\leq C_q^{-1} \Bigg\vert \frac{f(\boldsymbol{X}_{t+1})}{\sigma^2(t+1)} - \frac{f(\boldsymbol{X}_{t})}{\sigma^2(t)} - \int_t^{t+1} f(\boldsymbol{X}_s) d \left(\frac{1}{\sigma^{2}(s)} \right) \Bigg \vert \\ 
&\leq C_q^{-1} \Bigg\vert \frac{\sup_{x \in \mathbf{R}^d} f}{\sigma^2(t+1)} - \frac{\inf_{x \in \mathbf{R}^d} f}{\sigma^2(t)} - \int_t^{t+1} (\inf_{x \in \mathbf{R}^d} f) d \left(\frac{1}{\sigma^{2}(s)} \right) \Bigg \vert \\
&= C_q^{-1} \Bigg\vert \frac{\sup_{x \in \mathbf{R}^d} f}{\sigma^2(t+1)} - \inf_{x \in \mathbf{R}^d} f \left(\frac{1}{\sigma^2(t)} + \int_t^{t+1} d \left(\frac{1}{\sigma^{2}(s)} \right) \right) \Bigg \vert \\
&= C_q^{-1} \Bigg\vert \frac{\sup_{x \in \mathbf{R}^d} f}{\sigma^2(t+1)} - \inf_{x \in \mathbf{R}^d} f \left(\frac{1}{\sigma^2(t)} + \frac{1}{\sigma^{2}(t+1)} - \frac{1}{\sigma^2(t)} \right) \Bigg \vert \\
&= \frac{C_q^{-1}}{\sigma^2(t+1)} \Bigg\vert \sup_{x \in \mathbf{R}^d} f - \inf_{x \in \mathbf{R}^d} f  \Bigg \vert
\leq \frac{C_q^{-1}L_0 \rho }{\sigma^2(t+1)}. 
\end{aligned}
\label{lmpf2.7_eq10-04}    
\end{equation}

For $\int_{t}^{t+1} \frac{C_q^{-1}}{\sigma^{2}(s)} df(\boldsymbol{X}_s) < 0$, in the same manner, we get
\begin{equation}
\begin{aligned}
\Bigg\vert \int_{t}^{t+1} \frac{C_q^{-1}}{\sigma^{2}(s)} df(\boldsymbol{X}_s) \Bigg\vert
&\leq C_q^{-1} \Bigg\vert \frac{f(\boldsymbol{X}_{t+1})}{\sigma^2(t+1)} - \frac{f(\boldsymbol{X}_{t})}{\sigma^2(t)} - \int_t^{t+1} f(\boldsymbol{X}_s) d \left(\frac{1}{\sigma^{2}(s)} \right) \Bigg \vert \\ 
&\leq C_q^{-1} \Bigg\vert \frac{\inf_{x \in \mathbf{R}^d} f}{\sigma^2(t+1)} - \frac{\sup_{x \in \mathbf{R}^d} f}{\sigma^2(t)} - \int_t^{t+1} (\sup_{x \in \mathbf{R}^d} f) d \left(\frac{1}{\sigma^{2}(s)} \right) \Bigg \vert \\
&= C_q^{-1} \Bigg\vert \frac{\inf_{x \in \mathbf{R}^d} f}{\sigma^2(t+1)} - \sup_{x \in \mathbf{R}^d} f \left(\frac{1}{\sigma^2(t)} + \int_t^{t+1} d \left(\frac{1}{\sigma^{2}(s)} \right) \right) \Bigg \vert \\
&= C_q^{-1} \Bigg\vert \frac{\inf_{x \in \mathbf{R}^d} f}{\sigma^2(t+1)} - \sup_{x \in \mathbf{R}^d} f \left(\frac{1}{\sigma^2(t)} + \frac{1}{\sigma^{2}(t+1)} - \frac{1}{\sigma^2(t)} \right) \Bigg \vert \\
&= \frac{C_q^{-1}}{\sigma^2(t+1)} \Bigg\vert \inf_{x \in \mathbf{R}^d} f - \sup_{x \in \mathbf{R}^d} f  \Bigg \vert
\leq \frac{C_q^{-1}L_0 \rho }{\sigma^2(t+1)}. 
\end{aligned}
\label{lmpf2.7_eq10-05}    
\end{equation}

Hence, we get the upper bound of the first term in \eqref{lmpf2.7_eq10-02} as follows:
\begin{equation}
\Bigg\vert \int_{t}^{t+1} \frac{C_q^{-1}}{\sigma^{2}(s)} df(\boldsymbol{X}_s) \Bigg\vert \leq \frac{C_q^{-1}L_0 \rho }{\sigma^2(t+1)}    
\label{lmpf2.7_eq10-06}    
\end{equation}

The second term in \eqref{lmpf2.7_eq10-02} is the integration to deterministic time index $s$, so we can obtain the upper-bound conveniently holding \eqref{lmpf2.7_eq10-01} such that
\begin{equation}
\Bigg \vert \frac{1}{2} \int_{t}^{t+1} \Delta_{\boldsymbol{x}} f(\boldsymbol{X}_s) ds \Bigg \vert 
\leq \frac{1}{2} \sup_{\forall x \in \mathcal{D}} | \Delta_{\boldsymbol{x}} f(x) | = \frac{1}{2} L_2 d.
\label{lmpf2.7_eq10-07}    
\end{equation}

The result of \eqref{lmpf2.7_eq10-06} and \eqref{lmpf2.7_eq10-07} implies the first term of the right-hand side in \eqref{lmpf2.7_eq07} as follows:
\begin{equation}
\begin{aligned}
\Bigg \vert \int_{t}^{t+1} \frac{C_q^{-1}}{\sigma^{2}(s)} \nabla f(\boldsymbol{X}_s) \cdot d \bar{\boldsymbol{X}}_s \Bigg \vert
&\leq \Bigg\vert \int_{t}^{t+1} \frac{C_q^{-1}}{\sigma^{2}(s)} df(\boldsymbol{X}_s) \Bigg\vert + \Bigg \vert \frac{1}{2} \int_{t}^{t+1} \Delta f(\boldsymbol{X}_s) ds \Bigg \vert  \\
&\leq \frac{C_q^{-1}L_0 \rho }{\sigma^2(t+1)} + \frac{1}{2} L_2 d
< \frac{C_q^{-1}L_0 \rho + 0.5 L_2 d \cdot \sigma^2(0) }{\sigma^2(t+1)}. 
\end{aligned}
\label{lmpf2.7_eq10-08}    
\end{equation}
Since $\sigma(t)$ is a monotonic decreasing function, there exists a positive value $\bar{s} > 0$ such that $\sigma(t) \leq \bar{s}^{-1} \sigma(t+1)$. It implies that
\begin{equation}
\Bigg \vert -\int_{t}^{t+1} \frac{C_q^{-1}}{\sigma^{2}(s)} \nabla f(\boldsymbol{X}_s) \cdot d \bar{\boldsymbol{X}}_s \Bigg \vert 
\leq \frac{C_q^{-1}L_0 \rho + 0.5 L_D d \sigma^2(0)}{\bar{s}} \frac{1}{\sigma^2(t)}
= \frac{C_1}{\sigma^2(t)}
\label{lmpf2.7_eq10-09}
\end{equation}
 where $C_1$ denotes a positive value such that $C_1 > \frac{C_q^{-1}L_0 \rho + 0.5 L_D d \sigma^2(0)}{\bar{s}}$.

Furthermore, We can straightforwardly obtain the upper bound of the second term in the right-hand side of \eqref{lmpf2.7_eq07} as follows:
\begin{equation}
\begin{aligned}
\frac{1}{2} \Bigg\vert \int_t^{t+1} \frac{C_q^{-1}}{\sigma^2 (s) } \|  \nabla f(\boldsymbol{X}_{s}) \|^2 ds \Bigg\vert
&\leq \frac{1}{2} \frac{ C_q^{-1} }{\sigma^2 (t+1)} \sup \|  \nabla_x f(\boldsymbol{X}_{s}) \|^2 \int_t^{t+1} ds \\
&\leq \frac{1}{2 \sigma^2 (t+1)} C_q^{-1} \cdot C_0^2 
\leq \frac{C_2}{\sigma^2 (t)},   \quad \because C_2 > \frac{C_0^2}{2C_q \bar{s}}. 
\end{aligned}
\label{lmpf2.7_eq16}
\end{equation}

Since $\sigma(s) \triangleq b^{-\bar{p}(t)}$ is monotone decreasing function, the supremum of  $\sigma(s)$ is $\sigma(0) $ for all $s \in \mathbf{R}[0, \infty)$, i.e. $\sup_{s \in \mathbf{R}[0, \infty]}  \sigma(s) = \sigma (0) \triangleq \sigma$.
With the supremum of each term in \eqref{lmpf2.7_eq07}, we can obtain the lower bound of the Radon-Nykodym derivative \eqref{lmpf2.7_eq07}  such that
\begin{equation}
\frac{dP_{\boldsymbol{X}}}{dQ_{\bar{\boldsymbol{X}}}} \geq \exp \left( - \frac{C_1 + C_2}{\sigma^2 (t)}\right) \geq \exp \left(- \frac{C_3}{\sigma^2 (t)}  \right), \quad \because C_3 > C_2 + C_1.
\label{lmpf2.7_eq17}    
\end{equation}

Accordingly, for any $\varepsilon > 0$  and $\boldsymbol{X}_t, \; \boldsymbol{x}^* \in \mathbf{R}^d$,  the infimum of $P_x (|X_{t+1} - \boldsymbol{x}^*| < \varepsilon) $ is 
\begin{equation}
P_{\boldsymbol{X}} (\left\|\boldsymbol{X}_{t} - \boldsymbol{x}^*\right\| < \varepsilon) 
\geq \exp\left(- \frac{C_3}{\sigma^2(t)} \right) Q_{\bar{\boldsymbol{X}}} (\left\|\boldsymbol{X}_{t} - \boldsymbol{x}^*\right\| < \varepsilon).
\label{lmpf2.7_eq18}    
\end{equation}
As $Q_w$ is a normal distribution based on \eqref{lmpf2.7_eq06},  we have
\begin{equation}
\begin{aligned}
P_{\boldsymbol{X}} (|\boldsymbol{X}_{t+1} - \boldsymbol{x}^*| < \varepsilon) 
&\geq \exp\left(- \frac{C_3}{\sigma^2 (t)} \right) \int_{\left\| \boldsymbol{X} - \boldsymbol{x}^* \right\| < \varepsilon} \frac{1}{\sigma(t) \sqrt{2 \pi \int_{t}^{t+1}C_q d\tau}} \exp \left( -\frac{ (\boldsymbol{X} - \boldsymbol{x}^*)^2}{2 \int_{t}^{t+1} C_q \sigma^2(\tau) d\tau}  \right) dx \\
&\geq \exp\left(- \frac{C_3}{\sigma^2 (t)} \right) \int_{\| \boldsymbol{X} - \boldsymbol{x}^* \| < \varepsilon} \frac{1}{\sigma(t) \sqrt{2 \pi C_q \int_{t}^{t+1} d\tau}} \exp \left( -\frac{ (\sqrt{\rho} + \varepsilon)^2}{2  \sigma^2(0) C_q \int_{t}^{t+1} d\tau}  \right) dx  \\
&\geq \exp\left(- \frac{C_3}{\sigma^2 (t)} \right) \frac{1}{\sigma(0) \sqrt{2 \pi C_q}} \exp \left( -\frac{ (\sqrt{\rho} + \varepsilon)^2}{2 \sigma(0) C_q } \right) \int_{\| \boldsymbol{X} - \boldsymbol{x}^* \| < \varepsilon} dx \\
&= \exp\left(- \frac{C_3}{\sigma^2 (t)} \right) \frac{1}{\sqrt{2 \pi C_q}} \exp \left( -\frac{ (\sqrt{\rho} + \varepsilon)^2}{2 C_q } \right) \frac{2 \pi^{d/2}\varepsilon^{d}}{\Gamma(d/2+1)} \because \sigma(0) = 1 \\
&\geq \exp\left(- \frac{C_3}{\sigma^2 (t)} \right) \frac{1}{\sqrt{2 \pi C_q}} \left(1 + \frac{ (\sqrt{\rho} + \varepsilon)^2}{2 C_q } \right) \frac{2 \pi^{d/2}\varepsilon^{d}}{\Gamma(d/2+1)} \\
&\geq \exp\left(- \frac{C_3}{\sigma^2 (t)} \right) \frac{1}{\sqrt{2 \pi C_q}} \left(\frac{2  C_q+ (\sqrt{\rho} + \varepsilon)^2}{  C_q } \right) \Bigg\vert_{\rho=0, \varepsilon=0} \cdot \frac{\pi^{d/2}\varepsilon^{d}}{\Gamma(d/2+1)} \\
&\geq \exp\left(- \frac{C_3}{\sigma^2 (t)} \right) \cdot C_4 \cdot \varepsilon, \quad \because C_4 = \frac{\sqrt{2}}{\sqrt{\pi C_q}} \cdot \frac{\pi^{d/2}\varepsilon^{d-1}}{\Gamma(d/2+1)}. 
\end{aligned}
\label{lmpf2.7_eq19}    
\end{equation}

Finally, we obtain the lower bound of the transition probability  such that 
\begin{equation}
\label{lmpf2.7_eq20}
\begin{aligned}
\delta_t 
&= \inf_{\boldsymbol{x}, \boldsymbol{y} \in \mathbf{R}^d} p(t, \boldsymbol{x}, t+1, \boldsymbol{y}) \bigg\vert_{\boldsymbol{x} = \boldsymbol{X}_t,\; y=\boldsymbol{x}^*} \nonumber\\
&= \inf_{\boldsymbol{x}, \boldsymbol{y} \in \mathbf{R}^d} \lim_{\varepsilon \rightarrow 0} \frac{1}{\varepsilon} P_{\boldsymbol{X}} (\left\|\boldsymbol{X}_{t+1} - \boldsymbol{x}^*\right| < \varepsilon) \\
&\geq \inf_{\boldsymbol{x}, \boldsymbol{y} \in \mathbf{R}^d} \lim_{\varepsilon \rightarrow 0} \frac{1}{\varepsilon} \cdot C_4 \cdot \exp\left(- \frac{C_3}{\sigma^2 (t)} \right) \cdot  \varepsilon \\
&\geq \exp \left(-\frac{C_5}{\sigma^2(t)}  \right), \quad \because C_5 > C_3 + \cdot | \ln C_4 |
\end{aligned}    
\end{equation}
The above inequality implies that if there exists a monotone decreasing function such that $\sigma^2(s) \geq \frac{C_5}{ \log (t+2)}$, it satisfies that the convergence condition given by \eqref{lmpf2.7_eq04} such that
\begin{equation}
\sum_{k=0}^{\infty} \delta_{t+k} 
\geq \sum_{k=0}^{\infty} \exp \left( -\frac{C_5}{C_5} \log(t+2+k) \right) 
= \sum_{k=0}^{\infty} \frac{1}{t+2+k} 
= \infty, \quad \forall k \geq 0.    
\label{lmpf2.7_eq21}
\end{equation}
Substitute \eqref{lmpf2.7_eq21} into \eqref{lmpf2.7_eq04}, we obtain
\begin{equation}
    \overline{\lim_{\tau \rightarrow \infty}} \sup_{\boldsymbol{X}_t, x_{t+\tau} \in \mathbf{R}^d} \| p(t, \boldsymbol{X}_t, t + \tau,  \boldsymbol{x}^*) - p(t, \boldsymbol{X}_t, t + \tau,  \boldsymbol{x}^*) \| \leq \exp(-\sum_{k=0}^{\infty} \delta_{t+k}) ) = 0.
\end{equation}
 
%------------------------------------
\end{proof}
%------------------------------------

%=====================================================
\subsection{Local Convergence under Convex Assumption}
%=====================================================
%====================================
%\begin{assumption}  %theorem for local convergence 
%\label{assum_04}
\textbf{Assumption 4.}
%====================================
The Hessian of the objective function $\boldsymbol{H}(f): \mathbb{R}^d \mapsto \mathbb{R}^d$ around the optimal point is non-singular and positive definite,  
%\end{assumption}

%====================================
%\begin{theorem} % theorem for local convergence 
%\label{theorem_03}
\textbf{Theorem 3.4}
%====================================
The expectation value of the objective function derived by the proposed QSGLD converges to a locally optimal point asymptotically under Assumption $\ref{assum_04}$. 
%\end{theorem}
%------------------------------------
\begin{proof}   %for Theorem 3.4
%------------------------------------
Given the learning equation derived by QSGLD as \eqref{ack-ch03-eq01}, we can calculate the one-step difference of the objective function as follows:
\begin{equation}
\begin{aligned}
f(\boldsymbol{X}_{\tau+1}^Q) - f(\boldsymbol{X}_{\tau}^Q) 
&= \langle \nabla_{\boldsymbol{x}} f(\boldsymbol{X}_{\tau}^Q) , -\lambda \nabla_{\boldsymbol{x}} f(\boldsymbol{X}_{\tau}^Q) + Q_p^{-1}(\tau) \boldsymbol{\varepsilon}_{\tau}^q \rangle \\
&+ \lambda^2 \int_0^1 (1 - s) \langle \nabla_{\boldsymbol{x}} f(\boldsymbol{X}_{\tau}^Q), \boldsymbol{H}_{\boldsymbol{x}}(f)(\boldsymbol{X}_{\tau}^Q + s(\boldsymbol{X}_{\tau + 1}^Q - \boldsymbol{X}_{\tau}^Q)) \nabla_{\boldsymbol{x}} f(\boldsymbol{X}_{\tau}^Q) \rangle ds 
\end{aligned}
\label{thpf-eq01}
\end{equation}
, where $\boldsymbol{H}_{\boldsymbol{x}}(f)(\cdot) : \mathbf{R}^d \rightarrow \mathbf{R}^{d \times d}$ denotes Hessian of the objective function $f$. 

Assumption \ref{assum_04} and Definition \ref{def_st01} indicates that there exists the eigenvalue $M_{\max} \in \mathbf{R}^++$ of the Hessian such that $(\boldsymbol{H}_{\boldsymbol{x}}(f))_{\mathbf{R}^{d \times d}} \leq M_{\max}$. It implies 
\begin{equation}
\begin{aligned}
f(\boldsymbol{X}_{\tau+1}^Q) - f(\boldsymbol{X}_{\tau}^Q) 
&\leq \langle \nabla_{\boldsymbol{x}} f(\boldsymbol{X}_{\tau}^Q) , -\lambda \nabla_{\boldsymbol{x}} f(\boldsymbol{X}_{\tau}^Q) \rangle 
+ Q_p^{-1}(\tau) \langle \nabla_{\boldsymbol{x}} f(\boldsymbol{X}_{\tau}^Q), \boldsymbol{\varepsilon}_{\tau}^q \rangle 
+ \frac{1}{2} \lambda^2 M_{\max} \| \nabla_{\boldsymbol{x}} f(\boldsymbol{X}_{\tau}^Q) \|^2 \\
&= -\lambda \| \nabla_{\boldsymbol{x}} f(\boldsymbol{X}_{\tau}^Q) \|^2 
+ \frac{1}{2} \lambda^2 M_{\max} \| \nabla_{\boldsymbol{x}} f(\boldsymbol{X}_{\tau}^Q) \|^2
+ Q_p^{-1}(\tau) \langle \nabla_{\boldsymbol{x}} f(\boldsymbol{X}_{\tau}^Q), \boldsymbol{\varepsilon}_{\tau}^q \rangle \\
&= \lambda \| \nabla_{\boldsymbol{x}} f(\boldsymbol{X}_{\tau}^Q) \|^2 \left( \frac{1}{2} \lambda M_{\max} - 1 \right)
+ Q_p^{-1}(\tau) \langle \nabla_{\boldsymbol{x}} f(\boldsymbol{X}_{\tau}^Q), \boldsymbol{\varepsilon}_{\tau}^q \rangle.
\end{aligned}
\label{thpf-eq02}
\end{equation}
\begin{comment}
The $L_1$-Lipschitz condition and the finite domain assumption in Assumption \ref{assum01} implies that
\begin{equation}
\begin{aligned}
Q_p^{-1}(\tau) \langle \nabla_{\boldsymbol{x}} f(\boldsymbol{X}_{\tau}^Q), \boldsymbol{\varepsilon}_{\tau}^q \rangle
&\leq Q_p^{-1}(\tau) \| \nabla_{\boldsymbol{x}} f(\boldsymbol{X}_{\tau}^Q) \| \| \boldsymbol{\varepsilon}_{\tau}^q \| \\
&= q Q_p^{-1}(\tau) \| \nabla_{\boldsymbol{x}} f(\boldsymbol{X}_{\tau}^Q) - \nabla_{\boldsymbol{x}} f(\boldsymbol{X}^*) \| \| \boldsymbol{\varepsilon}_{\tau}^q \| \\
&\leq \frac{1}{2} Q_p^{-1}(\tau) L_1 \| \boldsymbol{X}_{\tau}^Q - \boldsymbol{X}^* \|, \quad \because \| \boldsymbol{\varepsilon}_{\tau}^q \| \leq \frac{1}{2} \\
&\leq \frac{1}{2} Q_p^{-1}(\tau) L_1 \rho
\end{aligned}
\label{thpf-eq03}
\end{equation}

Substituting $\eqref{thpf-eq01}$ into $\eqref{thpf-eq02}$, we can obtain the upper-bound of $f(\boldsymbol{X}_{\tau+1}^Q) - f(\boldsymbol{X}_{\tau}^Q)$ such that
\begin{equation}
\begin{aligned}
f(\boldsymbol{X}_{\tau+1}^Q) - f(\boldsymbol{X}_{\tau}^Q) 
\leq \lambda \| \nabla_{\boldsymbol{x}} f(\boldsymbol{X}_{\tau}^Q) \|^2 \left( \frac{1}{2} \lambda M_{\max} - 1 \right)
+ \frac{1}{2} Q_p^{-1}(\tau) L_1 \rho
\end{aligned}
\label{thpf-eq04}
\end{equation}
\end{comment}
Furthermore, assuming the existence of the minimum eigenvalue $m_{\min} \in \mathbf{R}$ of the Hessian matrix $H$ such that $(H)_{\mathbf{R}^{d \times d}} \geq m_{\min}$, we can compute the difference between the optimal point $\boldsymbol{x}^* \in \mathbb{R}^d$ and $\boldsymbol{X}_{\tau}$ as follows:
\begin{equation}
\begin{aligned}
f(\boldsymbol{x}^*) - f(\boldsymbol{X}_{\tau}^Q) 
&\geq -\lambda \| \nabla_{\boldsymbol{x}} f(\boldsymbol{X}_{\tau}^Q) \|^2 + \frac{1}{2} \lambda^2 m_{\min} \| \nabla_{\boldsymbol{x}} f(\boldsymbol{X}_{\tau}^Q) \|^2 \\
&=\frac{1}{2} m_{\min} \| \nabla_{\boldsymbol{x}} f(\boldsymbol{X}_{\tau}^Q) \|^2 \left(\lambda^2 - \frac{2}{m_{\min}} \lambda \right) \\
&=\frac{1}{2} m_{\min} \| \nabla_{\boldsymbol{x}} f(\boldsymbol{X}_{\tau}^Q \|^2) \left( \left(\lambda - \frac{1}{m_{\min}} \right)^2 -\frac{1}{m_{\min}^2}  \right) 
\geq -\frac{1}{2m_{min}} \| \nabla_{\boldsymbol{x}} f(\boldsymbol{X}_{\tau}) \|^2.
\end{aligned}
\label{thpf-eq05}
\end{equation}
For convenience, we abbreviate $m_{\min}$ as $m$ and $M_{\max}$ as $M$.
\eqref{thpf-eq05} implies that 
\begin{equation}
\| \nabla_{\boldsymbol{x}} f(\boldsymbol{X}_{\tau}) \|^2 \geq 2m \left( f(\boldsymbol{x}^*) - f(\boldsymbol{X}_{\tau}) \right).
\label{thpf-eq06}
\end{equation}

Let us assume that the learning rate is sufficiently small, such that $\lambda < \min \{1, \frac{2}{M} \}$.
By substituting the inequality \eqref{thpf-eq06} into \eqref{thpf-eq05}, we can derive the following equation:
\begin{equation}
\begin{aligned}
&f(\boldsymbol{X}_{\tau+1}^Q) - f(\boldsymbol{x}^*) + f(\boldsymbol{x}^*) - f(\boldsymbol{X}_{\tau}^Q) \\
&\leq \frac{1}{2}M \cdot 2m \left(f(\boldsymbol{X}_{\tau}^Q) - f(\boldsymbol{x}^*) \right) 
\left( \lambda^2 - \frac{2}{M} \lambda \right) + Q_p^{-1}(\tau) \langle \nabla_{\boldsymbol{x}} f(\boldsymbol{X}_{\tau}^Q), \boldsymbol{\varepsilon}_{\tau}^q \rangle \\
&\Rightarrow  f(\boldsymbol{X}_{\tau+1}^Q) - f(\boldsymbol{x}^*)\\
&\leq \left( 1 + M \cdot m \left( \lambda^2 - \frac{2}{M} \lambda \right) \right)\left( f(\boldsymbol{X}_{\tau}^Q) - f(\boldsymbol{x}^*) \right) + Q_p^{-1}(\tau) \langle \nabla_{\boldsymbol{x}} f(\boldsymbol{X}_{\tau}^Q), \boldsymbol{\varepsilon}_{\tau}^q \rangle.
\end{aligned}
\label{thpf-eq07}
\end{equation}

Applying the expectation of the quantization to both terms, we obtain 
\begin{equation}
\begin{aligned}
&\mathbb{E}_{\boldsymbol{\varepsilon_{\tau}}^q} f(\boldsymbol{X}_{\tau+1}^Q) - \mathbb{E}_{\boldsymbol{\varepsilon_{\tau}}^q} f(\boldsymbol{x}^*) \\
&\leq \left( 1 + M \cdot m \left( \lambda^2 - \frac{2}{M} \lambda \right) \right)\left( \mathbb{E}_{\boldsymbol{\varepsilon_{\tau}}^q} f(\boldsymbol{X}_{\tau}^Q) 
- \mathbb{E}_{\boldsymbol{\varepsilon_{\tau}}^q} f(\boldsymbol{x}^*) \right) \\
&+ Q_p^{-1}(\tau) \langle \nabla_{\boldsymbol{x}} f(\boldsymbol{X}_{\tau}^Q), \mathbb{E}_{\boldsymbol{\varepsilon_{\tau}}^q} \boldsymbol{\varepsilon}_{\tau}^q \rangle \\
&=  \left( 1 + M \cdot m \left( \lambda^2 - \frac{2}{M} \lambda \right) \right)\left( \mathbb{E}_{\boldsymbol{\varepsilon_{\tau}}^q} f(\boldsymbol{X}_{\tau}^Q) 
- f(\boldsymbol{x}^*) \right)
\end{aligned}
\label{thpf-eq07-01}
\end{equation}
, where $\mathbb{E}_{\boldsymbol{\varepsilon_{\tau}}^q} f(\boldsymbol{x}^*) = f(\boldsymbol{x}^*)$.

To assess the convergence, we extend the inequality \eqref{thpf-eq07-01} to $t+k$ for $k > 0$.
\begin{equation}
\begin{aligned}
\mathbb{E}_{\boldsymbol{\varepsilon_{\tau}}^q} f(\boldsymbol{X}_{\tau+k}^Q) - f(\boldsymbol{x}^*) 
\leq  \prod_{j=0}^{k-1}\left( 1 + M \cdot m \left( \lambda^2 - \frac{2}{M} \lambda \right) \right)\left( \mathbb{E}_{\boldsymbol{\varepsilon_{\tau}}^q} f(\boldsymbol{X}_{\tau}^Q) 
- f(\boldsymbol{x}^*) \right).
\end{aligned}
\label{thpf-eq08}
\end{equation}

The exponential lemma \eqref{eq02:lemma} to the \eqref{thpf-eq08} yields 
\begin{equation}
\begin{aligned}
&\mathbb{E}_{\boldsymbol{\varepsilon_{\tau}}^q} f(\boldsymbol{X}_{\tau+k}^Q) - f(\boldsymbol{x}^*) \\
&\leq  \exp \left( M \cdot m \cdot \lambda \cdot \sum_{j=0}^{k-1}\left( \lambda - \frac{2}{M} \right) \right)\left( \mathbb{E}_{\boldsymbol{\varepsilon_{\tau}}^q} f(\boldsymbol{X}_{\tau}^Q) 
- f(\boldsymbol{x}^*) \right).
\end{aligned}
\label{thpf-eq09}
\end{equation}
The assumption on $\lambda$ such that $\lambda < \min\{1, \frac{2}{M}\}$ implies the existence of a negative value, denoted as $\bar{h}^{-}(k)$, which depends on $k$ and is defined as
\begin{equation}
\bar{h}^{-}(k) \triangleq \frac{1}{\lambda} \sum_{j=0}^{k-1}\left( \lambda - \frac{2}{M} \right) = \sum_{j=0}^{k-1}\left( 1 - \frac{2}{\lambda M} \right) = -c_1 k, \quad \because c_1 = \left| 1 - \frac{2}{\lambda M} \right| > 0.   
\end{equation}

Thus, we obtain 
\begin{equation}
\begin{aligned}
\mathbb{E}_{\boldsymbol{\varepsilon_{\tau}}^q} f(\boldsymbol{X}_{\tau+k}^Q) - f(\boldsymbol{x}^*) 
&\leq  \exp \left( M m \lambda^2 \bar{h}^{-}(k) \right)\left( \mathbb{E}_{\boldsymbol{\varepsilon_{\tau}}^q} f(\boldsymbol{X}_{\tau}^Q) 
- f(\boldsymbol{x}^*) \right) \\
&\leq  \exp \left( - C_0 \cdot k \right)\left( \mathbb{E}_{\boldsymbol{\varepsilon_{\tau}}^q} f(\boldsymbol{X}_{\tau}^Q) 
- f(\boldsymbol{x}^*) \right)
\end{aligned}
\label{thpf-eq10}
\end{equation}
, where $C_0$ denotes $M \cdot m \cdot \lambda^2 \cdot c_1$.

The Lipschitz continuity assumption implies that 
\begin{equation}
\begin{aligned}
\mathbb{E}_{\boldsymbol{\varepsilon_{\tau}}^q}  f(\boldsymbol{X}_{\tau}^Q) 
- f(\boldsymbol{x}^*) 
&\leq \sup_{\boldsymbol{X}_{\tau}^Q \in B^o(\boldsymbol{x}^*, \rho)} 
\| f(\boldsymbol{X}_{\tau}^Q) - f(\boldsymbol{x}^*) \| 
\leq L_0 \| \boldsymbol{X}_{\tau}^Q - \boldsymbol{x}^* \| \\
&\leq L_0 \rho
\end{aligned}
\end{equation}

Consequently, applying an absolute value to both terms, we get
\begin{equation}
\begin{aligned}
\mathbb{E}_{\boldsymbol{\varepsilon_{\tau}}^q} f(\boldsymbol{X}_{\tau+k}^Q) - f(\boldsymbol{x}^*) 
&\leq  \exp \left( - C_0 \cdot k \right) L_0 \rho \\
&=     \exp \left( - C_0 \cdot k + \ln L_0\right) \rho
\end{aligned}
\label{thpf-eq11}
\end{equation}
The result of \eqref{thpf-eq11} describes that for all $\rho > 0$, we can find an appropriate positive value $\delta > 0$, which implies $\delta = \exp(-c_0 k + a) \rho$ so that $\mathbb{E}_{\boldsymbol{\varepsilon_{\tau}}^q} f(\boldsymbol{X}_{\tau+k}^Q) \rightarrow f(\boldsymbol{x}^*) $.
Therefore, we can pick a $k > k_0 = \left\lceil \frac{1}{C_0} \ln L_0 \right\rceil$ satisfying the following proposition of convergence:
\begin{equation}
\forall \varepsilon > 0, \; \exists \rho>0 \text{ such that } \| \boldsymbol{X}_{\tau}^Q - \boldsymbol{x}^* \| < \rho 
\implies \vert \mathbb{E}_{\boldsymbol{\varepsilon_{\tau}}^q} f(\boldsymbol{X}_{\tau+k}^Q) - f(\boldsymbol{x}^*) \vert < \varepsilon(\rho). 
\end{equation}

%------------------------------------    
\end{proof}
%------------------------------------

%%%%%%%%%%%%%%%%%%%%%%%%%%%%%%%%%%%%%%%%%%%%%%%%%%%%%%%%%%%%%%%%%%%%%%%
\section{Detailed information of Experimental Results}
%%%%%%%%%%%%%%%%%%%%%%%%%%%%%%%%%%%%%%%%%%%%%%%%%%%%%%%%%%%%%%%%%%%%%%%
%----------------------------------------------------- 
% Figures - 1 Fashion MNIST
%----------------------------------------------------- 
\begin{figure}[t]
\centering
{
    \begin{subfigure}[b]{0.45\textwidth}
    \includegraphics[width=\textwidth]{FashionMNIST-05-24-01.png}
    \caption{FashionMNIST via CNN}
    \label{fig-fm-01}
    \end{subfigure}
    \hfill
    \begin{subfigure}[b]{0.45\textwidth}
    \includegraphics[width=\textwidth]{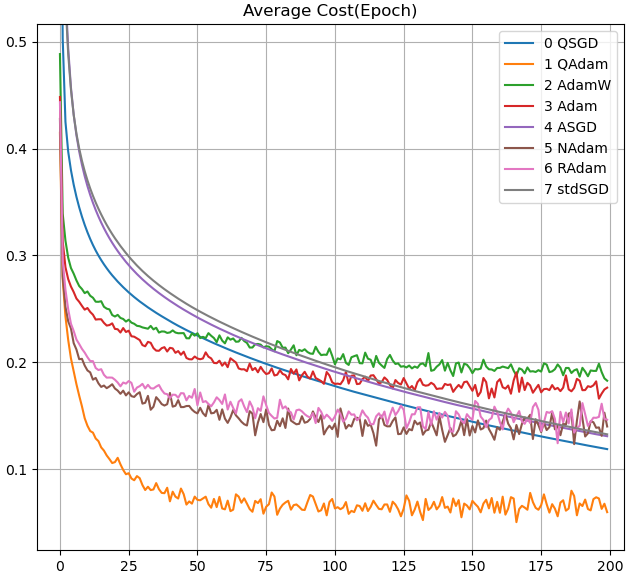}
    \caption{Enlarged plot}
    \label{fig-fm-02}
    \end{subfigure}
    \hfill
}    
\caption{The error trends of test algorithms to the dataset and neural models: (a) Training error trends of the CNN model on FashionMNIST dataset.(b) Enlarged Training error trends}
\label{fig-fm}
\end{figure}
%----------------------------------------------------- 

\begin{table}[]
\caption{Experimental Environment}
\centering
{
\begin{tabular}{l|lll}
\hline
PC Name & OS                & GPU                       & CPU           \\ \hline
PC-1    & Linux Ubuntu 22.0 & NVIDIA GeForce GTX 1080Ti & Intel i9 7900 \\
PC-2    & Windows 11        & NVIDIA GeForce RTX 3050   & Intel i9 7900 \\
PC-3    & Windows 11        & NVIDIA GeForce GTXTi      & Intel i7 6700 \\ \hline
\end{tabular}
}
\label{table-01}
\end{table}

%=====================================================
%\subsection{Figures and Graph for the simulation result}
%=====================================================
%----------------------------------------------------- 
% Figures - 2 CIFAR-10
%-----------------------------------------------------
\begin{figure}[t]
\centering
{
    \begin{subfigure}[b]{0.45\textwidth}
    \includegraphics[width=\textwidth]{CIFAR-10-05-24-01.png}
    \caption{CIFAR10 via ResNet50}
    \label{fig-cf-01}
    \end{subfigure}
    \hfill
    \begin{subfigure}[b]{0.45\textwidth}
    \includegraphics[width=\textwidth]{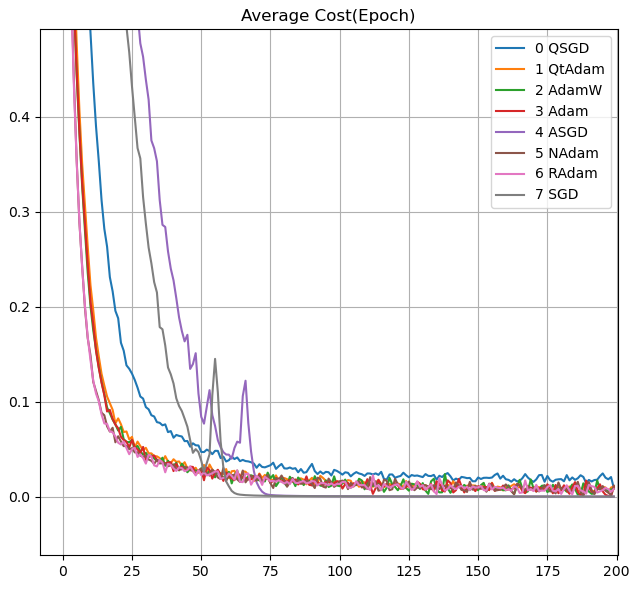}
    \caption{Enlarged plot}
    \label{fig-cf-02}
    \end{subfigure}
    \hfill
}    
\caption{The error trends of test algorithms to the dataset and neural models: (a) Training error trends of ResNet-50 on CIFAR-10 dataset.(b) Enlarged Training error trends}
\label{fig-cf10}
\end{figure}

%----------------------------------------------------- 
%  Table-2 전체 결과
%----------------------------------------------------- 
\begin{table}[]
\caption{Comparison of test performance among optimizers with a fixed learning rate 0.01. Evaluation is based on the Top-1 accuracy of the training and testing data.}
\centering
{
\scriptsize
\begin{tabular}{l|ccc|cccccc|}
\hline
Data Set   & \multicolumn{3}{c|}{FashionMNIST}      & \multicolumn{3}{c|}{CIFAR10}                             & \multicolumn{3}{c|}{CIFAR100}       \\ \hline
Model      & \multicolumn{3}{c|}{CNN with 3-Layer Blocks} & \multicolumn{6}{c|}{ResNet-50 (56 Layer Blocks)}                                                                 \\ \hline
Algorithms & Training  & Testing  & Training Error  & Training & Testing & \multicolumn{1}{c|}{Training Error} & Training & Testing & Training Error \\ \hline
QSGD       & 97.10     & 91.59    & 0.085426        & 99.90    & 73.80   & \multicolumn{1}{c|}{0.009253}       & 99.04    & 37.77   & 0.030104       \\
QADAM      & 98.43     & 89.29    & 0.059952        & 99.99    & 85.09   & \multicolumn{1}{c|}{0.011456}       & 98.62    & 49.60   & 0.037855       \\
SGD        & 95.59     & 91.47    & 0.132747        & 99.99    & 63.31   & \multicolumn{1}{c|}{0.001042}       & 98.24    & 25.90   & 0.005478       \\
ASGD       & 95.60     & 91.42    & 0.130992        & 99.99    & 63.46   & \multicolumn{1}{c|}{0.001166}       & 98.36    & 26.43   & 0.004981       \\
ADAM       & 92.45     & 87.12    & 0.176379        & 99.75    & 82.08   & \multicolumn{1}{c|}{0.012421}       & 98.85    & 46.32   & 0.038741       \\
ADAMW      & 91.72     & 86.81    & 0.182867        & 99.57    & 82.20   & \multicolumn{1}{c|}{0.012551}       & 98.86    & 47.01   & 0.038002       \\
NADAM      & 96.25     & 87.55    & 0.140066        & 99.56    & 82.46   & \multicolumn{1}{c|}{0.014377}       & 98.62    & 48.56   & 0.037409       \\
RADAM      & 95.03     & 87.75    & 0.146404        & 99.65    & 82.26   & \multicolumn{1}{c|}{0.010526}       & 98.17    & 48.61   & 0.044193       \\ \hline
\end{tabular}
}
\label {table-02}
\end{table}

We conducted the experiments using a Python program based on the PyTorch framework version 1.13.1.
For the experiments, we utilized three computers, and the detailed specifications of each computer are provided in Table \ref{table-01}.
The Python version used was 3.10.0, and the Anaconda version was 23.10.
We conducted the experiments for the FashionMNIST dataset using a vanilla CNN with three-layer blocks. For the CIFAR-10 and CIFAR-100 datasets, we used the ResNet-50 model with 56 layer blocks.
The representative experimental results are presented in Table \ref{table-02}.
A fixed learning rate of 0.01 is utilized in all experiments, and 200 epochs are conducted for all datasets.
The batch sizes for FashionMNIST, CIFAR-10, and CIFAR-100 are 100 samples, 128 samples, and 100 samples, respectively.
%=====================================================
\subsection{Learning Equations used in the Experiment}
%=====================================================
%----------------------------------------------------- 
% Figures - 3 CIFAR-100
%-----------------------------------------------------
\begin{figure}[t]
\centering
{
    \begin{subfigure}[b]{0.45\textwidth}
    \includegraphics[width=\textwidth]{CIFAR-100-05-24-01.png}
    \caption{CIFAR100 via ResNet50}
    \label{fig-ack-01-002}
    \end{subfigure}
    \hfill
    \begin{subfigure}[b]{0.45\textwidth}
    \includegraphics[width=\textwidth]{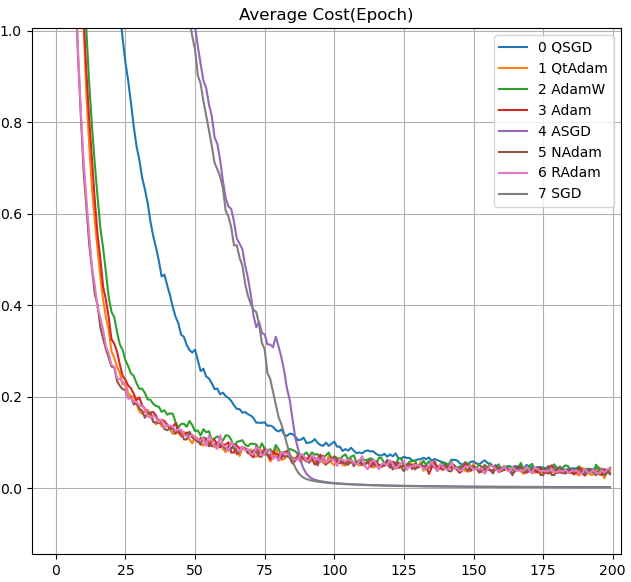}
    \caption{Enlarged plot}
    \label{fig-ack-02-002}
    \end{subfigure}
    \hfill
}    
\caption{The error trends of test algorithms to the dataset and neural models: (a) Training error trends of ResNet-50 on CIFAR-100 dataset.(b) Enlarged Training error trends}
\label{fig-cf100}
\end{figure}

%----------------------------------------------------- 
\textbf{QSGLD and QSLD-ADAM}

The fundamental learning equation of the proposed algorithm is as follows:
\begin{equation}
\boldsymbol{X}_{\tau + 1}^Q = \boldsymbol{X}_{\tau}^Q + Q_p^{-1}(\tau) \left[ Q_p(\tau) \cdot \lambda h(\boldsymbol{X}_{\tau}^Q)\right]^Q.
\label{ack-ex01}
\end{equation}
, where $h(\boldsymbol{X}_{\tau}) = -\nabla_{\boldsymbol{x}} f((\boldsymbol{X}_{\tau})$ for QSGLD, and  $h(\boldsymbol{X}_{\tau}) = -\frac{\hat{\boldsymbol{m}}_{\tau}}{\sqrt{\hat{\boldsymbol{v}}_{\tau}} + \epsilon}$ for Adam-based QSLD, respectively.

The quantization parameter $Q_p$ is defined as follows:
\begin{equation}
Q_p = \eta \cdot b^{\bar{p}(t_e)}.
\label{ack-ex02}
\end{equation}
where $t_e$ represents the time index of each epoch, given by $t_e = \frac{\tau}{B}$ for the number of mini-batches per epoch $B$, and $\bar{p}(t_e)$ denotes the power function, derived as follows:"
\begin{equation}
\begin{aligned}
Q_p = 
\eta \cdot b^{\bar{p}(t_e)} \vert_{t_e = \tau/B} &\leq \sqrt{\frac{1}{C} \log(\tau + 2)} \\
b^{2\bar{p}(t_e)} \vert_{t_e = \tau/B} & \leq \frac{1}{\eta^2 \, C} \log(\tau + 2) \\
\bar{p}(t_e) \vert_{t_e = \tau/B} &\leq \frac{1}{2} \log_b \left( \frac{1}{\eta^2 \, C} \log(\tau + 2) \right).
\end{aligned}
\label{ack-ex03}
\end{equation}

For convenience, we define the constant $C$ in \eqref{ack-ex03} as the reciprocal value of $\eta^2$.
Additionally, Considering the upper bound of the quantization parameter, which needs to be a rational number, we apply the floor function to the upper bound of the quantization parameter.
Therefore, the power function for the quantization parameter is as follows:
\begin{equation}
\bar{p}(t_e) \vert_{t_e = \tau/B} \triangleq \lfloor 0.5 \cdot \log_b \log(\tau + 2) \rfloor.
\label{ack-ex04}
\end{equation}

Finally, we introduce an enforcement function to prevent early paralysis, defined as follows:
\begin{equation}
r(\tau, \boldsymbol{X}_\tau) = \lambda \cdot \left(\frac{\exp(-\varkappa(\tau - \tau_0))}{1 + \exp(-\varkappa(\tau - \tau_0))} \cdot \frac{h(\boldsymbol{X}_{\tau})}{\| h(\boldsymbol{X}_{\tau}) \|} \right),\quad \tau_0 \in \mathbf{Z}^d
\label{ack-ex05}
\end{equation}
, where $\tau_0$ is a parameter, measured in mini-batches, that determines the interval for applying the enforcement function during the learning process. The parameter $\varkappa$ represents the shape of the enforcement function, with larger values causing a rapid decrease toward zero. 

The following is the summary of all the equations for the proposed algorithm:
\begin{equation}
\begin{aligned}    
t_e &= \tau/B \\
\bar{p}(t_e) &= \lfloor 0.5 \cdot \log_b \log(\tau + 2) \rfloor \\
Q_p &= \eta \cdot b^{\bar{p}(t_e)} \\
r(\tau) &= \lambda \cdot \left(\frac{\exp(-\varkappa(\tau - \tau_0))}{1 + \exp(-\varkappa(\tau - \tau_0))} \cdot \frac{h(\boldsymbol{X}_{\tau}^Q)}{\| h(\boldsymbol{X}_{\tau}^Q) \|} \right) \\
\boldsymbol{X}_{\tau + 1}^Q &= \boldsymbol{X}_{\tau}^Q + Q_p^{-1}(\tau) \left[ Q_p(\tau) \cdot \left(\lambda h(\boldsymbol{X}_{\tau}^Q) + r(\tau, \boldsymbol{X}_\tau^Q) \right) \right]^Q.
\end{aligned}
\end{equation}

We recommend the hyper-parameters represented as follows:
\begin{equation}
\eta^2 \in 2^{19} \approx 0.5 \times 10^6, \; C = 1/\eta^2, \; b=2 \; \kappa=2.0 \text{ or } 4.0, \; t_0 = 5\text{ ~ }20 \text{\% of all epochs}.
\end{equation}

In the following section, we present an empirical analysis of the impact of changing hyperparameters.

\textbf{SGD}
We set the SGD for the experiments using standard gradient descent form, as follows:

\begin{equation}
\boldsymbol{X}_{\tau + 1} = \boldsymbol{X}_{\tau} - \lambda \nabla_{\boldsymbol{x}} f(\boldsymbol{X}_{\tau}).
\end{equation}

\textbf{ASGD} (Average SGD by \citet{Shamir_ICML_2013}) optimizer updates the parameters using the following equation:

\begin{equation}
\boldsymbol{X}_{\tau + 1} = \frac{1}{t}\sum_{i=0}^{t-1} \nabla_{\boldsymbol{x}} f(\boldsymbol{X}_{\tau - i} ),
\end{equation}
where $\boldsymbol{X}_{\tau}$ represents the updated parameter at time step $\tau$.

\textbf{ADAM} (Adaptive Moment Estimation by \citet{Kingma_2015}) optimizer updates the parameters using the following equations:

\begin{equation}
\begin{aligned}
\boldsymbol{m}_{\tau} &= \beta_1 \cdot \boldsymbol{m}_{\tau-1} + (1 - \beta_1) \cdot \boldsymbol{g}_{\tau}, \\
\boldsymbol{v}_{\tau} &= \beta_2 \cdot \boldsymbol{v}_{\tau-1} + (1 - \beta_2) \cdot \boldsymbol{g}_{\tau}^2, \\
\hat{\boldsymbol{m}}_{\tau} &= \frac{\boldsymbol{m}_{\tau}}{1 - \beta_1^{\tau}}, \\
\hat{\boldsymbol{v}}_{\tau} &= \frac{\boldsymbol{v}_{\tau}}{1 - \beta_2^{\tau}}, \\
\boldsymbol{X}_{\tau} &= \boldsymbol{X}_{{\tau}-1} - \frac{\eta}{\sqrt{\hat{\boldsymbol{v}}_{\tau}} + \epsilon} \cdot \boldsymbol{\hat{m}}_{\tau},
\end{aligned}
\end{equation}

where $\boldsymbol{m}_{\tau} \in \mathbf{R}^d$ and $\boldsymbol{v}_{\tau} \in \mathbf{R}^d$ are the first and second moment estimates respectively, $\boldsymbol{g}_{\tau} \in \mathbf{R}^d$ is the gradient at time step $t$ i.e. $-\nabla_{\boldsymbol{x}} f(\boldsymbol{X}_t)$, $\beta_1$ and $\beta_2$ are the decay rates for the moments, $\hat{\boldsymbol{m}}_{\tau} \in \mathbf{R}^d$ and $\hat{\boldsymbol{v}}_{\tau} \in \mathbf{R}^d$ are the bias-corrected moment estimates, $\boldsymbol{X}_{\tau}$ represents the updated parameter at time step $t$, $\eta$ is the learning rate, and $\epsilon$ is a small constant to avoid division by zero.
We set the hyperparameters for ADAM such that
\begin{equation}
\beta_1 = 0.9, \; \beta_2 = 0.999, \; \epsilon = 10^{-8}. 
\end{equation}
We utilize the ADAMW optimizer implemented in PyTorch.

\textbf{NADAM} (Nestrov momentum incooperated ADAM by \citet{dozat_ICLR_WH_2016})
\begin{equation}
\begin{aligned}
\mu_\tau &= \beta_1 \left(1 - \frac{1}{2} 0.96^{\tau \psi})\right) \\
\mu_{\tau+1} &= \beta_1 \left(1 - \frac{1}{2} 0.96^{(\tau + 1)\psi}\right) \\
\boldsymbol{m}_{\tau} &= \beta_1 \cdot \boldsymbol{m}_{\tau-1} + (1 - \beta_1) \cdot \boldsymbol{g}_{\tau}, \\
\boldsymbol{v}_{\tau} &= \beta_2 \cdot \boldsymbol{v}_{\tau-1} + (1 - \beta_2) \cdot \boldsymbol{g}_{\tau}^2, \\
\hat{\boldsymbol{m}}_{\tau} &= \frac{\mu_{\tau + 1}}{1 - \prod_{i=1}^{t+1}} \boldsymbol{m}_{\tau} + \frac{1 - \mu_{\tau}}{1 - \prod_{i=1}^{t}} \boldsymbol{g}_{\tau}, \\
\hat{\boldsymbol{v}}_{\tau} &= \frac{\boldsymbol{v}_{\tau}}{1 - \beta_2^{\tau}}, \\
\boldsymbol{X}_{\tau} &= \boldsymbol{X}_{{\tau}-1} - \frac{\eta}{\sqrt{\hat{\boldsymbol{v}}_{\tau}} + \epsilon} \cdot \boldsymbol{\hat{m}}_{\tau},
\end{aligned}
\end{equation}

\textbf{RADAM} (Rectified Adam by \citet{Liyuan_ICLR_2020}) optimizer updates the parameters using the following equations:

\begin{equation}
\begin{aligned}
\boldsymbol{m}_{\tau} &= \beta_1 \cdot \boldsymbol{m}_{\tau-1} + (1 - \beta_1) \cdot \boldsymbol{g}_{\tau}, \\
\boldsymbol{v}_{\tau} &= \beta_2 \cdot \boldsymbol{v}_{\tau-1} + (1 - \beta_2) \cdot \boldsymbol{g}_{\tau}^2, \\
\hat{\boldsymbol{m}}_{\tau} &= \frac{\boldsymbol{m}_{\tau}}{1 - \beta_1^{\tau}}, \\
\hat{\boldsymbol{v}}_{\tau} &= \frac{\boldsymbol{v}_{\tau}}{1 - \beta_2^{\tau}}, \\
\rho_{\tau} &= \rho_{\infty} - \frac{2 t \beta_2^t}{1 - \beta_2^t} \\
\boldsymbol{r}_{\tau} &= \sqrt{\frac{(\rho_{\tau} - 4)(\rho_{\tau}-2)\rho_{\infty}}{(\rho_{\infty} - 4)(\rho_{\infty}-2)\rho_{\tau}}}, \\
\boldsymbol{X}_{\tau} &= \boldsymbol{X}_{\tau-1} - \eta \, \boldsymbol{m}_{\tau} \cdot
\begin{cases}
\boldsymbol{r}_{\tau} \frac{\sqrt{1 - \beta_2^{\tau}}}{\sqrt{\boldsymbol{v}_{\tau}} + \epsilon} & \rho_t > 5 \\
1 & \text{else}
\end{cases}
\end{aligned}
\end{equation}
, where $\boldsymbol{r}_{\tau} \in \mathbf{R}^d$ is the "leaky" update term, and $\rho_{\tau}, \rho_{\infty}$ is an additional hyperparameter introduced in RADAM. $\rho_{\infty}$ is initialized with $\rho_{\infty} = \frac{2}{1 - \beta_2} - 1$.
We utilize the RADAM optimizer implemented in PyTorch.

\textbf{ADAMW} optimizer updates the parameters using the following equations:

\begin{equation}
\begin{aligned}
\boldsymbol{m}_{\tau} &= \beta_1 \cdot \boldsymbol{m}_{\tau-1} + (1 - \beta_1) \cdot \boldsymbol{g}_{\tau}, \\
\boldsymbol{v}_{\tau} &= \beta_2 \cdot \boldsymbol{v}_{\tau-1} + (1 - \beta_2) \cdot \boldsymbol{g}_{\tau}^2, \\
\hat{\boldsymbol{m}}_{\tau} &= \frac{\boldsymbol{m}_{\tau}}{1 - \beta_1^{\tau}}, \\
\hat{\boldsymbol{v}}_{\tau} &= \frac{\boldsymbol{v}_{\tau}}{1 - \beta_2^{\tau}}, \\
\boldsymbol{X}_{\tau} &= \boldsymbol{X}_{\tau-1} - \frac{\eta}{\sqrt{\hat{v}_{\tau}} + \epsilon} \cdot (\boldsymbol{\hat{m}}_{\tau} + \lambda \boldsymbol{X}_{\tau-1}),
\end{aligned}
\end{equation}
where $\lambda$ is a weight decay coefficient or regularization term added in ADAMW.
\citet{loshchilov_ICLR_2019} provided the algorithm.
We utilize the ADAMW optimizer implemented in PyTorch.

%=====================================================
\subsection{Experimental Results According to Datasets}
%=====================================================
%----------------------------------------------------- 
% Figures - 3 CIFAR-100
%----------------------------------------------------- 
\begin{figure}[t]
\centering
{
    \begin{subfigure}[b]{0.45\textwidth}
    \includegraphics[width=\textwidth]{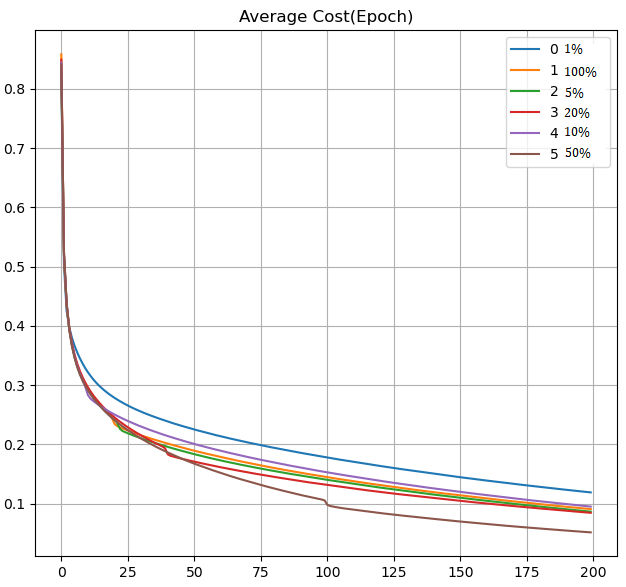}
    \caption{Application Period of the Enforcement}
    \label{fig-ack-04-001}
    \end{subfigure}
    \hfill
    \begin{subfigure}[b]{0.45\textwidth}
    \includegraphics[width=\textwidth]{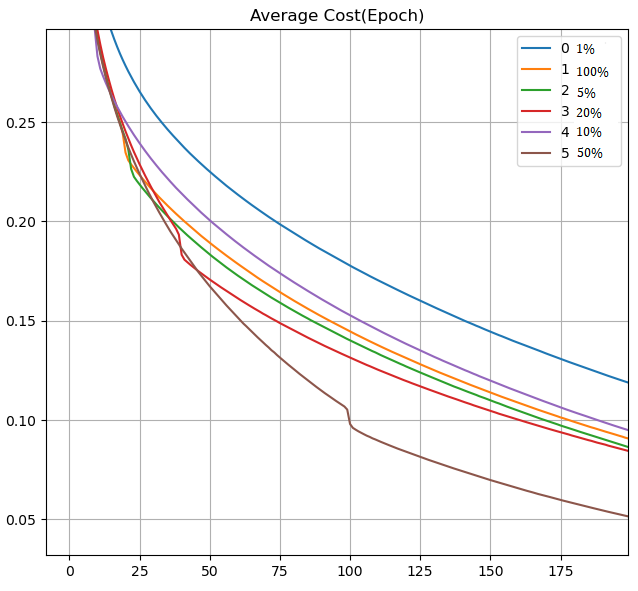}
    \caption{Enlarged plot}
    \label{fig-ack-04-002}
    \end{subfigure}
    \hfill
}    
\caption{The error trends of performance Variation of Adam-Based QSLD based on the application period of the enforcement function: (a) Training error trends of Vanilla CNN on FashionMNIST dataset.(b) Enlarged Training error trends}
\label{fig-ack-03}
\end{figure}

%----------------------------------------------------- 

\begin{table}[]
\caption{Performance Variation of Adam-Based QSLD Based on the Application Period of the Enforcement Function }
\centering
{
\scriptsize
\begin{tabular}{l|ccc|ccc|ccc}
\hline
Period Ratio (\%) & \multicolumn{3}{l|}{FashionMNIST} & \multicolumn{3}{l|}{CIFAR-10} & \multicolumn{3}{l}{CIFAR-100} \\ \cline{2-10} 
                  & Training   & Testing  & Error     & Training & Testing & Error    & Training & Testing & Error    \\ \hline
0.5               & 96.05      & 91.04    & 0.92853   & 99.74    & 83.61   & 0.007327 & 98.66    & 44.51   & 0.036558 \\
1.0               & 96.11      & 91.59    & 0.118914  & 99.84    & 83.29   & 0.006766 & 98.62    & 44.38   & 0.032726 \\
5.0               & 96.75      & 91.38    & 0.094954  & 99.87    & 83.39   & 0.006549 & 98.61    & 49.60   & 0.037855 \\
10.0              & 96.95      & 91.59    & 0.090756  & 99.84    & 84.18   & 0.0      & 98.84    & 44.59   & 0.035554 \\
20.0              & 97.05      & 90.84    & 0.084489  & 100.0    & 85.08   & 0.0      & 99.13    & 46.78   & 0.035354 \\
50.0              & 97.82      & 90.64    & 0.051491  & 99.91    & 84.04   & 0.005456 & 98.94    & 48.49   & 0.029179 \\
100.0             & 99.46      & 90.01    & 0.104510  & 99.87    & 83.39   & 0.010629 & 98.81    & 46.76   & 0.044370 \\ \hline
\end{tabular}
}
\label {table-03}
\end{table}

\textbf{FashionMNIST}
The FashionMNIST dataset is a drop-in replacement for the original MNIST dataset, which consists of handwritten digits. 
The FashionMNIST dataset contains 60,000 grayscale images of 10 different fashion categories, each with a 28x28 pixel representation.
The ten fashion categories in FashionMNIST include T-shirts/tops, trousers, pullovers, dresses, coats, sandals, shirts, sneakers, bags, and ankle boots. Each image in the dataset is associated with a corresponding label indicating the category of the depicted fashion item.

Simple vanilla multilayer networks, equipped with well-tuned optimizers and moderately wide hidden layers, can achieve high accuracy in classifying each category of the MNIST dataset, resulting in minimal accuracy errors. This poses challenges in meaningfully evaluating the performance of different optimizers.

However, while the classification test scores for FashionMNIST are higher than CIFAR-10 and CIFAR-100, standard SGD performs superior to the ADAM optimizer family in evaluation tests on the FashionMNIST dataset. This result suggests that the objective function of FashionMNIST exhibits a more convex property around the optimal point, and previous research (\citet{xie_ICLR_2021}) has revealed that the ADAM optimizer may struggle to select flat minima.

\textbf{Experiments to FashionMNIST}
The experiments on the FashionMNIST dataset yielded interesting results.
Firstly, similar to the MNIST dataset, the experimental results showed that the training classification performance exceeded 90\% for all models.
However, there was a noticeable performance difference between the SGD and ADAM optimizers. The SGD optimizer exhibited better classification performance compared to the ADAM optimizer.
As shown in Figure \ref{fig-fm}, the error trend indicated that ADAM converged faster initially, but as the number of epochs increased, SGD gradually reduced the error more effectively than the ADAM optimizer.
In general, once the number of epochs exceeded 400, the SGD optimizer consistently achieved a training classification accuracy of 100\% regardless of the learning rate. However, the test classification accuracy reached a limit of around 91.25\%.

The proposed QSGLD and ADAM-based QSLD perform better than the conventional SGD and ADAM optimizer, but they do not exhibit significant improvements on the FashionMNIST dataset.
QSGLD achieves a slight improvement of approximately 0.2\% in classification accuracy for both training and testing, while Adam-Based QSLD achieves an improvement of around 2\%.
In Figure \ref{fig-fm-01}, it can be observed that QSGLD converges faster than SGD due to the similar learning rate, although the convergence speed is similar.
On the other hand, Adam-based QSLD exhibits a convergence trend similar to the Adam optimizer. 
Nevertheless, Adam-based QSLD shows lower error trends than conventional ADAM optimizers.

%----------------------------------------------------- 
% Figures - 5 CIFAR-10 vs ResNet20 /CIFAR-100 via ResNet 100
%----------------------------------------------------- 
\begin{figure}[t]
\centering
{
    \begin{subfigure}[b]{0.45\textwidth}
    \includegraphics[width=\textwidth]{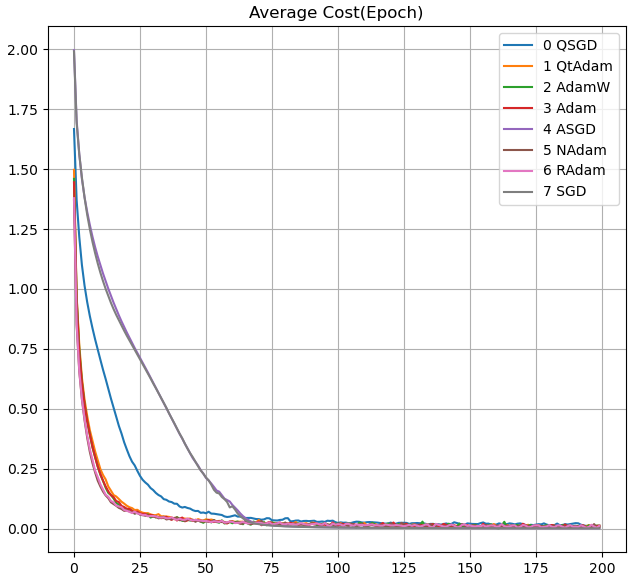}
    \caption{Test for CIFAR-10 on ResNet-20}
    \label{fig-ack-04-003}
    \end{subfigure}
    \hfill
    \begin{subfigure}[b]{0.45\textwidth}
    \includegraphics[width=\textwidth]{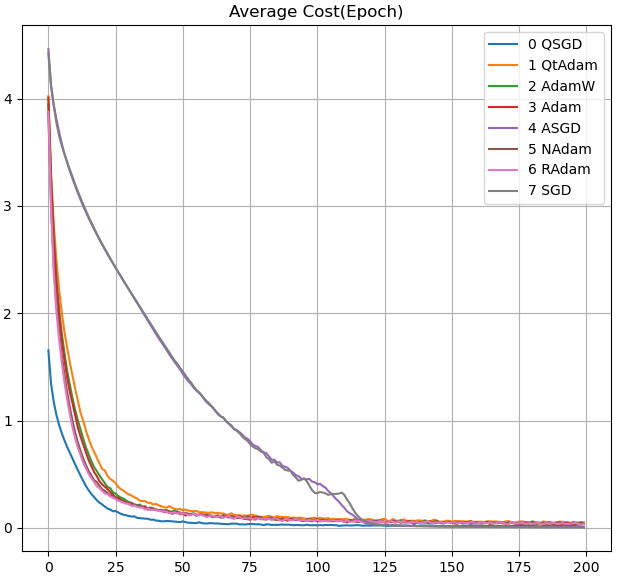}
    \caption{Test for CIFAR-100 on ResNet-32}
    \label{fig-ack-04-004}
    \end{subfigure}
    \hfill
}    
\caption{Performance Comparison of the Proposed Algorithm on Networks with Different Depths: (a) When testing on the CIFAR-10 dataset, the algorithm was evaluated on ResNet-20.(b) When testing on the CIFAR-100 dataset, the algorithm was evaluated on ResNet-30.}
\label{fig-ack-05}
\end{figure}

\textbf{CIFAR10 and CIFAR100}
We don't provide any specific explanation for the CIFAR-10 and CIFAR-100 datasets.
For the CIFAR datasets, separate experiments were conducted based on the depth of ResNet. 
This was done to verify that the proposed algorithm works effectively regardless of the depth of the neural network model.
The experimental results demonstrated the superiority of the proposed algorithm for all depths of ResNet. 
As shown in Table \ref{table-02}, when classifying the CIFAR-10 dataset using ResNet-50, QSGD outperformed SGD by 8\% in terms of test accuracy.

As depicted in Figure \ref{fig-cf10}, QSGLD exhibited significant improvements in both convergence speed and error reduction compared to the conventional SGD methods. On the other hand, Adam-Based QSLD showed a performance advantage of approximately 1.5\% for test accuracy.

Similar trends were observed for the CIFAR-100 dataset. QSGD demonstrated a performance advantage of around 11\% over conventional SGD methods for test accuracy. In contrast, Adam-based QSLD showed an improvement of approximately 1.0\% compared to conventional Adam-based optimizers.

The experiments on the CIFAR datasets revealed interesting characteristics of QSGLD and Adam-based QSLD.

%=====================================================
\subsection{Experimental Results of Changing of Hyper-Parameters}
%=====================================================
%=====================================================
\textbf{Period Parameter for Enforcement Function }
%=====================================================
We conduct experiments to analyze the effect of the period parameter $\tau_0$ for the enforcement function.
Regarding the quantization parameter, there were variations depending on the algorithm, but generally, the size of the search vector was around $10^{-6}$, so the eta value was set to half of it, which is 524288.
Firstly, the optimal application period of the enforcement function varied depending on the dataset. 
For FashionMNIST and CIFAR-10, the best performance improvement was observed when the enforcement function was applied for about 10-20\% of the epochs. 
Applying it for longer or shorter periods did not result in significant performance improvement.
On the other hand, for CIFAR-100, the best performance improvement was achieved when the enforcement function was applied throughout all the epochs.
Since CIFAR-100 has a higher classification complexity than FashionMNIST or CIFAR-10, it is considered optimal to integrate the enforcement function with a measure that can assess problem complexity, such as the Fisher Information Matrix, rather than using it as a function of time.
Regarding another hyperparameter, $\kappa$, no significant performance variations were observed with changes in its value.

\begin{comment}
%%%%%%%%%%%%%%%%%%%%%%%%%%%%%%%%%%%%%%%%%%%%%%%%%%%%%%%%%%%%%%%%%%%%%%%
\section{Acknowledgement}
%%%%%%%%%%%%%%%%%%%%%%%%%%%%%%%%%%%%%%%%%%%%%%%%%%%%%%%%%%%%%%%%%%%%%%%
This work was supported by Institute for Information and Communications Technology Promotion(IITP) grant funded by the Korean government(MSIP) (2021-0- 00766, Development of Integrated Development Framework that supports Automatic Neural Network Generation and Deployment optimized for Runtime Environment)
\end{comment}

\begin{comment}
%%%%%%%%%%%%%%%%%%%%%%%%%%%%%%%%%%%%%%%%%%%%%%%%%%%%%%%%%%%%%%%%%%%%%%%
\section{Conclusion}
%%%%%%%%%%%%%%%%%%%%%%%%%%%%%%%%%%%%%%%%%%%%%%%%%%%%%%%%%%%%%%%%%%%%%%%
In the supplementary material, we provide detailed proofs of the theorems and additional propositions presented in the submitted manuscript. Additionally, we present comprehensive information on the experiments conducted to validate the proposed quantization algorithm. 

Additionally, we intended to upload the test code to platforms like GitHub and provide the link in the Supplementary Material. However, due to a recent policy change on Github, anonymous uploads are no longer possible. Therefore, we had no choice but to include the code in the Supplementary Material.

\end{comment}

%%%%%%%%%%%%%%%%%%%%%%%%%%%%%%%%%%%%%%%%%%%%%%%%%%%%%%%%%%%%
\newpage
%\section*{References}
%\bibliography{supplementary-4810}
%%%%%%%%%%%%%%%%%%%%%%%%%%%%%%%%%%%%%%%%%%%%%%%%%%%%%%%%%%%%

%% file: main.bbl
\begin{thebibliography}{61}
\providecommand{\natexlab}[1]{#1}
\providecommand{\url}[1]{\texttt{#1}}
\expandafter\ifx\csname urlstyle\endcsname\relax
  \providecommand{\doi}[1]{doi: #1}\else
  \providecommand{\doi}{doi: \begingroup \urlstyle{rm}\Url}\fi

\bibitem[Altschuler and Talwar(2022)]{Altschuler_NEURIPS2022}
Jason Altschuler and Kunal Talwar.
\newblock Privacy of noisy stochastic gradient descent: More iterations without more privacy loss.
\newblock In S.~Koyejo, S.~Mohamed, A.~Agarwal, D.~Belgrave, K.~Cho, and A.~Oh, editors, \emph{Advances in Neural Information Processing Systems}, volume~35, pages 3788--3800. Curran Associates, Inc., 2022.
\newblock URL \url{https://proceedings.neurips.cc/paper_files/paper/2022/file/18561617ca0b4ffa293166b3186e04b0-Paper-Conference.pdf}.

\bibitem[Benedetto et~al.(2004)Benedetto, Yilmaz, and Powell]{Benedetto_2004}
J.J. Benedetto, O.~Yilmaz, and A.M. Powell.
\newblock Sigma-delta quantization and finite frames.
\newblock In \emph{2004 IEEE International Conference on Acoustics, Speech, and Signal Processing}, volume~3, pages iii--937, 2004.
\newblock \doi{10.1109/ICASSP.2004.1326700}.

\bibitem[Box and Muller(1958)]{Box_Muller_1958}
G.~E.~P. Box and Mervin~E. Muller.
\newblock {A Note on the Generation of Random Normal Deviates}.
\newblock \emph{The Annals of Mathematical Statistics}, 29\penalty0 (2):\penalty0 610 -- 611, 1958.
\newblock \doi{10.1214/aoms/1177706645}.
\newblock URL \url{https://doi.org/10.1214/aoms/1177706645}.

\bibitem[Brosse et~al.(2018)Brosse, Durmus, and Moulines]{Brosse_NEURIPS_2018}
Nicolas Brosse, Alain Durmus, and Eric Moulines.
\newblock The promises and pitfalls of stochastic gradient langevin dynamics.
\newblock In S.~Bengio, H.~Wallach, H.~Larochelle, K.~Grauman, N.~Cesa-Bianchi, and R.~Garnett, editors, \emph{Advances in Neural Information Processing Systems}, volume~31. Curran Associates, Inc., 2018.
\newblock URL \url{https://proceedings.neurips.cc/paper_files/paper/2018/file/335cd1b90bfa4ee70b39d08a4ae0cf2d-Paper.pdf}.

\bibitem[Chaudhari and Soatto(2018)]{Pratik_2018}
Pratik Chaudhari and Stefano Soatto.
\newblock Stochastic gradient descent performs variational inference, converges to limit cycles for deep networks.
\newblock In \emph{2018 Information Theory and Applications Workshop, {ITA} 2018, SanDiego, CA, USA, February 11-16, 2018}, pages 1--10. {IEEE}, 2018.
\newblock \doi{10.1109/ITA.2018.8503224}.
\newblock URL \url{https://doi.org/10.1109/ITA.2018.8503224}.

\bibitem[Cheng et~al.(2020{\natexlab{a}})Cheng, Yin, Bartlett, and Jordan]{Cheng_2020}
Xiang Cheng, Dong Yin, Peter Bartlett, and Michael Jordan.
\newblock Stochastic gradient and langevin processes.
\newblock In \emph{Proceedings of the 37th International Conference on Machine Learning}, ICML'20. JMLR.org, 2020{\natexlab{a}}.

\bibitem[Cheng et~al.(2020{\natexlab{b}})Cheng, Yin, Bartlett, and Jordan]{Cheng_2020_ICML}
Xiang Cheng, Dong Yin, Peter Bartlett, and Michael Jordan.
\newblock Stochastic gradient and {L}angevin processes.
\newblock In Hal~Daumé III and Aarti Singh, editors, \emph{Proceedings of the 37th International Conference on Machine Learning}, volume 119 of \emph{Proceedings of Machine Learning Research}, pages 1810--1819. PMLR, 13--18 Jul 2020{\natexlab{b}}.
\newblock URL \url{https://proceedings.mlr.press/v119/cheng20e.html}.

\bibitem[Dalalyan and Karagulyan(2019)]{Dalalyan_2019}
Arnak~S. Dalalyan and Avetik Karagulyan.
\newblock User-friendly guarantees for the langevin monte carlo with inaccurate gradient.
\newblock \emph{Stochastic Processes and their Applications}, 129\penalty0 (12):\penalty0 5278--5311, 2019.
\newblock ISSN 0304-4149.
\newblock \doi{https://doi.org/10.1016/j.spa.2019.02.016}.
\newblock URL \url{https://www.sciencedirect.com/science/article/pii/S0304414918304824}.

\bibitem[De~Sa et~al.(2015)De~Sa, Zhang, Olukotun, R\'{e}, and R\'{e}]{Christopher_2015}
Christopher~M De~Sa, Ce~Zhang, Kunle Olukotun, Christopher R\'{e}, and Christopher R\'{e}.
\newblock Taming the wild: A unified analysis of hogwild-style algorithms.
\newblock In C.~Cortes, N.~D. Lawrence, D.~D. Lee, M.~Sugiyama, and R.~Garnett, editors, \emph{Advances in Neural Information Processing Systems 28}, pages 2674--2682. NIPS, 2015.

\bibitem[Dozat(2016)]{dozat_ICLR_WH_2016}
Timothy Dozat.
\newblock Incorporating {Nesterov Momentum into Adam}.
\newblock In \emph{Proceedings of the 4th International Conference on Learning Representations: Workshop Track}, pages 1--4, 2016.

\bibitem[Fonseca and Saporito(2022)]{Fonseca_NEURIPS2022}
Yuri Fonseca and Yuri Saporito.
\newblock Statistical learning and inverse problems: A stochastic gradient approach.
\newblock In S.~Koyejo, S.~Mohamed, A.~Agarwal, D.~Belgrave, K.~Cho, and A.~Oh, editors, \emph{Advances in Neural Information Processing Systems}, volume~35, pages 9591--9602. Curran Associates, Inc., 2022.
\newblock URL \url{https://proceedings.neurips.cc/paper_files/paper/2022/file/3e8b1835833ef809059efa74b9df6805-Paper-Conference.pdf}.

\bibitem[Geman and Hwang(1986)]{Geman-1986}
Stuart Geman and Chii-Ruey Hwang.
\newblock Diffusions for global optimization.
\newblock \emph{SIAM Journal on Control and Optimization}, 24\penalty0 (5):\penalty0 1031--1043, 1986.

\bibitem[Goyal et~al.(2018)Goyal, Dollár, Girshick, Noordhuis, Wesolowski, Kyrola, Tulloch, Jia, and He]{goyal_2018}
Priya Goyal, Piotr Dollár, Ross Girshick, Pieter Noordhuis, Lukasz Wesolowski, Aapo Kyrola, Andrew Tulloch, Yangqing Jia, and Kaiming He.
\newblock Accurate, large minibatch sgd: Training imagenet in 1 hour, 2018.
\newblock URL \url{https://arxiv.org/abs/1706.02677}.

\bibitem[Granziol et~al.(2022)Granziol, Zohren, and Roberts]{Granziol_2022_JMLR}
Diego Granziol, Stefan Zohren, and Stephen Roberts.
\newblock Learning rates as a function of batch size: A random matrix theory approach to neural network training.
\newblock \emph{Journal of Machine Learning Research}, 23\penalty0 (173):\penalty0 1--65, 2022.
\newblock URL \url{http://jmlr.org/papers/v23/20-1258.html}.

\bibitem[Gray and Neuhoff(2006)]{Gray:2006}
Robert.~M. Gray and David~L. Neuhoff.
\newblock Quantization.
\newblock \emph{IEEE Transactions on Information Theory}, 44\penalty0 (6):\penalty0 2325--2383, 2006.

\bibitem[Han et~al.(2015)Han, Pool, Tran, and Dally]{Song_2015}
Song Han, Jeff Pool, John Tran, and William Dally.
\newblock Learning both weights and connections for efficient neural network.
\newblock In C.~Cortes, N.~D. Lawrence, D.~D. Lee, M.~Sugiyama, and R.~Garnett, editors, \emph{Advances in Neural Information Processing Systems 28}, pages 1135--1143. Curran Associates, Inc., 2015.

\bibitem[Hoffer et~al.(2017)Hoffer, Hubara, and Soudry]{Hoffer_2017}
Elad Hoffer, Itay Hubara, and Daniel Soudry.
\newblock Train longer, generalize better: Closing the generalization gap in large batch training of neural networks.
\newblock In \emph{Proceedings of the 31st International Conference on Neural Information Processing Systems}, NIPS'17, page 1729–1739, Red Hook, NY, USA, 2017. Curran Associates Inc.
\newblock ISBN 9781510860964.

\bibitem[Jim{\'{e}}nez et~al.(2007)Jim{\'{e}}nez, Wang, and Wang]{Jimnez_2007}
David Jim{\'{e}}nez, Long Wang, and Yang Wang.
\newblock White noise hypothesis for uniform quantization errors.
\newblock \emph{{SIAM} J. Math. Analysis}, 38\penalty0 (6):\penalty0 2042--2056, 2007.

\bibitem[Jung et~al.(2019)Jung, Son, Lee, Son, Han, Kwak, Hwang, and Choi]{Sangil_2019}
Sangil Jung, Changyong Son, Seohyung Lee, JinWoo Son, Jae{-}Joon Han, Youngjun Kwak, Sung~Ju Hwang, and Changkyu Choi.
\newblock Learning to quantize deep networks by optimizing quantization intervals with task loss.
\newblock In \emph{{IEEE} Conference on Computer Vision and Pattern Recognition, {CVPR} 2019, Long Beach, CA, USA, June 16-20, 2019}, pages 4350--4359, 2019.

\bibitem[Kalil~Lauand and Meyn(2022)]{Kalil_NEURIPS2022}
Caio Kalil~Lauand and Sean Meyn.
\newblock Approaching quartic convergence rates for quasi-stochastic approximation with application to gradient-free optimization.
\newblock In S.~Koyejo, S.~Mohamed, A.~Agarwal, D.~Belgrave, K.~Cho, and A.~Oh, editors, \emph{Advances in Neural Information Processing Systems}, volume~35, pages 15743--15756. Curran Associates, Inc., 2022.
\newblock URL \url{https://proceedings.neurips.cc/paper_files/paper/2022/file/6530274c68e81047e1f4a2ceb0b8c0ef-Paper-Conference.pdf}.

\bibitem[Kidambi et~al.(2018)Kidambi, Netrapalli, Jain, and Kakade]{Kidambi_2018_ITA}
Rahul Kidambi, Praneeth Netrapalli, Prateek Jain, and Sham~M. Kakade.
\newblock On the insufficiency of existing momentum schemes for stochastic optimization.
\newblock In \emph{ITA}, pages 1--9. IEEE, 2018.
\newblock ISBN 978-1-7281-0124-8.
\newblock URL \url{http://dblp.uni-trier.de/db/conf/ita/ita2018.html#KidambiN0K18}.

\bibitem[Kingma and Ba(2015)]{Kingma_2015}
Diederik~P. Kingma and Jimmy Ba.
\newblock Adam: {A} method for stochastic optimization.
\newblock In Yoshua Bengio and Yann LeCun, editors, \emph{3rd International Conference on Learning Representations, {ICLR} 2015, San Diego, CA, USA, May 7-9, 2015, Conference Track Proceedings}, 2015.
\newblock URL \url{http://arxiv.org/abs/1412.6980}.

\bibitem[Klebaner(2012)]{Klebaner_2011}
Fima~C. Klebaner.
\newblock \emph{Introduction To Stochastic Calculus With Applications}.
\newblock Imperial College Press, 3rd edition, 2012.
\newblock ISBN 1848168322.

\bibitem[Krizhevsky(2014)]{krizhevsky}
Alex Krizhevsky.
\newblock One weird trick for parallelizing convolutional neural networks, 2014.
\newblock URL \url{https://arxiv.org/abs/1404.5997}.

\bibitem[Kushner(1974)]{Kushner-1974}
Harold~J. Kushner.
\newblock {On the Weak Convergence of Interpolated Markov Chains to a Diffusion}.
\newblock \emph{The Annals of Probability}, 2\penalty0 (1):\penalty0 40 -- 50, 1974.
\newblock \doi{10.1214/aop/1176996750}.
\newblock URL \url{https://doi.org/10.1214/aop/1176996750}.

\bibitem[Li et~al.(2019)Li, Tai, and E]{Li-2019-JMLR}
Qianxiao Li, Cheng Tai, and Weinan E.
\newblock Stochastic modified equations and dynamics of stochastic gradient algorithms i: Mathematical foundations.
\newblock \emph{Journal of Machine Learning Research}, 20\penalty0 (40):\penalty0 1--47, 2019.
\newblock URL \url{http://jmlr.org/papers/v20/17-526.html}.

\bibitem[Li and Li(2019)]{Xiaoyun_2019}
Xiaoyun Li and Ping Li.
\newblock Generalization error analysis of quantized compressive learning.
\newblock In H.~Wallach, H.~Larochelle, A.~Beygelzimer, F.~d\textquotesingle Alch\'{e}-Buc, E.~Fox, and R.~Garnett, editors, \emph{Advances in Neural Information Processing Systems 32}, pages 15124--15134. NIPS, 2019.

\bibitem[Li et~al.(2021)Li, Malladi, and Arora]{Li_NEURIPS2021}
Zhiyuan Li, Sadhika Malladi, and Sanjeev Arora.
\newblock On the validity of modeling sgd with stochastic differential equations (sdes).
\newblock In M.~Ranzato, A.~Beygelzimer, Y.~Dauphin, P.S. Liang, and J.~Wortman Vaughan, editors, \emph{Advances in Neural Information Processing Systems}, volume~34, pages 12712--12725. Curran Associates, Inc., 2021.
\newblock URL \url{https://proceedings.neurips.cc/paper_files/paper/2021/file/69f62956429865909921fa916d61c1f8-Paper.pdf}.

\bibitem[Li et~al.(2022)Li, Wang, and Yu]{Li_NEURIPS2022}
Zhiyuan Li, Tianhao Wang, and Dingli Yu.
\newblock Fast mixing of stochastic gradient descent with normalization and weight decay.
\newblock In S.~Koyejo, S.~Mohamed, A.~Agarwal, D.~Belgrave, K.~Cho, and A.~Oh, editors, \emph{Advances in Neural Information Processing Systems}, volume~35, pages 9233--9248. Curran Associates, Inc., 2022.
\newblock URL \url{https://proceedings.neurips.cc/paper_files/paper/2022/file/3c215225324f9988858602dc92219615-Paper-Conference.pdf}.

\bibitem[Lin et~al.(2020)Lin, Stich, Patel, and Jaggi]{Tao_ICLR_2020}
Tao Lin, Sebastian~U. Stich, Kumar~Kshitij Patel, and Martin Jaggi.
\newblock Don't use large mini-batches, use local sgd.
\newblock In \emph{International Conference on Learning Representations}, 2020.
\newblock URL \url{https://openreview.net/forum?id=B1eyO1BFPr}.

\bibitem[Liu and Belkin(2020)]{Liu_ICLR_2020}
Chaoyue Liu and Mikhail Belkin.
\newblock Accelerating sgd with momentum for over-parameterized learning.
\newblock In \emph{International Conference on Learning Representations}, 2020.
\newblock URL \url{https://openreview.net/forum?id=r1gixp4FPH}.

\bibitem[Liu et~al.(2020)Liu, Jiang, He, Chen, Liu, Gao, and Han]{Liyuan_ICLR_2020}
Liyuan Liu, Haoming Jiang, Pengcheng He, Weizhu Chen, Xiaodong Liu, Jianfeng Gao, and Jiawei Han.
\newblock On the variance of the adaptive learning rate and beyond.
\newblock In \emph{International Conference on Learning Representations}, 2020.
\newblock URL \url{https://openreview.net/forum?id=rkgz2aEKDr}.

\bibitem[Loshchilov and Hutter(2019)]{loshchilov_ICLR_2019}
Ilya Loshchilov and Frank Hutter.
\newblock Decoupled weight decay regularization.
\newblock In \emph{International Conference on Learning Representations}, 2019.
\newblock URL \url{https://openreview.net/forum?id=Bkg6RiCqY7}.

\bibitem[Ma et~al.(2018)Ma, Bassily, and Belkin]{Ma_pmlr-2018}
Siyuan Ma, Raef Bassily, and Mikhail Belkin.
\newblock The power of interpolation: Understanding the effectiveness of {SGD} in modern over-parametrized learning.
\newblock In Jennifer Dy and Andreas Krause, editors, \emph{Proceedings of the 35th International Conference on Machine Learning}, volume~80 of \emph{Proceedings of Machine Learning Research}, pages 3325--3334. PMLR, 10--15 Jul 2018.
\newblock URL \url{https://proceedings.mlr.press/v80/ma18a.html}.

\bibitem[Malladi et~al.(2022)Malladi, Lyu, Panigrahi, and Arora]{Malladi_NEURIPS2022}
Sadhika Malladi, Kaifeng Lyu, Abhishek Panigrahi, and Sanjeev Arora.
\newblock On the sdes and scaling rules for adaptive gradient algorithms.
\newblock In S.~Koyejo, S.~Mohamed, A.~Agarwal, D.~Belgrave, K.~Cho, and A.~Oh, editors, \emph{Advances in Neural Information Processing Systems}, volume~35, pages 7697--7711. Curran Associates, Inc., 2022.
\newblock URL \url{https://proceedings.neurips.cc/paper_files/paper/2022/file/32ac710102f0620d0f28d5d05a44fe08-Paper-Conference.pdf}.

\bibitem[Mandt et~al.(2017)Mandt, Hoffman, and Blei]{Stephan_JMLR}
Stephan Mandt, Matthew~D. Hoffman, and David~M. Blei.
\newblock Stochastic gradient descent as approximate bayesian inference.
\newblock \emph{Journal of Machine Learning Research}, 18\penalty0 (134):\penalty0 1--35, 2017.
\newblock URL \url{http://jmlr.org/papers/v18/17-214.html}.

\bibitem[Marco and Neuhoff(2005)]{Marco_2005}
D.~Marco and D.L. Neuhoff.
\newblock The validity of the additive noise model for uniform scalar quantizers.
\newblock \emph{IEEE Transactions on Information Theory}, 51\penalty0 (5):\penalty0 1739--1755, 2005.
\newblock \doi{10.1109/TIT.2005.846397}.

\bibitem[Marsaglia(1963)]{Marsaglia_1963}
G.~Marsaglia.
\newblock Generating a variable from the tail of the normal distribution.
\newblock In \emph{Mathematical Note No. 322}, pages 1--3. Boeing Scientific Research Labs., 1963.

\bibitem[Marsaglia and Tsang(2000)]{Marsaglia_2000}
George Marsaglia and Wai~Wan Tsang.
\newblock The ziggurat method for generating random variables.
\newblock \emph{Journal of Statistical Software}, 5\penalty0 (8):\penalty0 1–7, 2000.
\newblock \doi{10.18637/jss.v005.i08}.
\newblock URL \url{https://www.jstatsoft.org/index.php/jss/article/view/v005i08}.

\bibitem[Masiha et~al.(2022)Masiha, Salehkaleybar, He, Kiyavash, and Thiran]{Masiha_NEURIPS2022}
Saeed Masiha, Saber Salehkaleybar, Niao He, Negar Kiyavash, and Patrick Thiran.
\newblock Stochastic second-order methods improve best-known sample complexity of sgd for gradient-dominated functions.
\newblock In S.~Koyejo, S.~Mohamed, A.~Agarwal, D.~Belgrave, K.~Cho, and A.~Oh, editors, \emph{Advances in Neural Information Processing Systems}, volume~35, pages 10862--10875. Curran Associates, Inc., 2022.
\newblock URL \url{https://proceedings.neurips.cc/paper_files/paper/2022/file/46323351ebc2afa42b30a6122815cb95-Paper-Conference.pdf}.

\bibitem[Mou et~al.(2018)Mou, Wang, Zhai, and Zheng]{Wenlong_ICML_2018}
Wenlong Mou, Liwei Wang, Xiyu Zhai, and Kai Zheng.
\newblock Generalization bounds of sgld for non-convex learning: Two theoretical viewpoints.
\newblock In Sébastien Bubeck, Vianney Perchet, and Philippe Rigollet, editors, \emph{Proceedings of the 31st Conference On Learning Theory}, volume~75 of \emph{Proceedings of Machine Learning Research}, pages 605--638. PMLR, 06--09 Jul 2018.
\newblock URL \url{https://proceedings.mlr.press/v75/mou18a.html}.

\bibitem[Nguyen et~al.(2019)Nguyen, Simsekli, Gurbuzbalaban, and RICHARD]{Nguyen_NEURIPS2019}
Thanh~Huy Nguyen, Umut Simsekli, Mert Gurbuzbalaban, and Ga\"{e}l RICHARD.
\newblock First exit time analysis of stochastic gradient descent under heavy-tailed gradient noise.
\newblock In H.~Wallach, H.~Larochelle, A.~Beygelzimer, F.~d\textquotesingle Alch\'{e}-Buc, E.~Fox, and R.~Garnett, editors, \emph{Advances in Neural Information Processing Systems}, volume~32. Curran Associates, Inc., 2019.
\newblock URL \url{https://proceedings.neurips.cc/paper_files/paper/2019/file/a97da629b098b75c294dffdc3e463904-Paper.pdf}.

\bibitem[Raginsky et~al.(2017)Raginsky, Rakhlin, and Telgarsky]{Raginsky_2017}
Maxim Raginsky, Alexander Rakhlin, and Matus Telgarsky.
\newblock Non-convex learning via stochastic gradient langevin dynamics: a nonasymptotic analysis.
\newblock In Satyen Kale and Ohad Shamir, editors, \emph{Proceedings of the 2017 Conference on Learning Theory}, volume~65 of \emph{Proceedings of Machine Learning Research}, pages 1674--1703. PMLR, 07--10 Jul 2017.
\newblock URL \url{https://proceedings.mlr.press/v65/raginsky17a.html}.

\bibitem[Seide et~al.(2014)Seide, Fu, Droppo, Li, and Yu]{Seide_2014}
Frank Seide, Hao Fu, Jasha Droppo, Gang Li, and Dong Yu.
\newblock 1-bit stochastic gradient descent and application to data-parallel distributed training of speech dnns.
\newblock In \emph{Interspeech 2014}, September 2014.

\bibitem[Shallue et~al.(2018)Shallue, Lee, Antognini, Sohl-dickstein, Frostig, and Dahl]{Shallue_2018}
Chris Shallue, Jaehoon Lee, Joseph Antognini, Jascha Sohl-dickstein, Roy Frostig, and George Dahl.
\newblock Measuring the effects of data parallelism on neural network training.
\newblock \emph{Journal of Machine Learning Research (JMLR)}, 2018.
\newblock URL \url{https://arxiv.org/pdf/1811.03600.pdf}.

\bibitem[Shamir and Zhang(2013)]{Shamir_ICML_2013}
Ohad Shamir and Tong Zhang.
\newblock Stochastic gradient descent for non-smooth optimization: Convergence results and optimal averaging schemes.
\newblock In Sanjoy Dasgupta and David McAllester, editors, \emph{Proceedings of the 30th International Conference on Machine Learning}, volume~28 of \emph{Proceedings of Machine Learning Research}, pages 71--79, Atlanta, Georgia, USA, 17--19 Jun 2013. PMLR.
\newblock URL \url{https://proceedings.mlr.press/v28/shamir13.html}.

\bibitem[Simsekli et~al.(2019)Simsekli, Sagun, and Gurbuzbalaban]{Simsekli_pmlr_2019}
Umut Simsekli, Levent Sagun, and Mert Gurbuzbalaban.
\newblock A tail-index analysis of stochastic gradient noise in deep neural networks.
\newblock In Kamalika Chaudhuri and Ruslan Salakhutdinov, editors, \emph{Proceedings of the 36th International Conference on Machine Learning}, volume~97 of \emph{Proceedings of Machine Learning Research}, pages 5827--5837. PMLR, 09--15 Jun 2019.
\newblock URL \url{https://proceedings.mlr.press/v97/simsekli19a.html}.

\bibitem[Smith et~al.(2019)Smith, Elsen, and De]{Smith_ICML_2019}
Samuel~L. Smith, Erich Elsen, and Soham De.
\newblock Momentum enables large batch training.
\newblock In \emph{ICML workshop on Theoretical Physics in Deep Learning}, 2019.
\newblock URL \url{https://proceedings.neurips.cc/paper_files/paper/2019/file/e0eacd983971634327ae1819ea8b6214-Paper.pdf}.

\bibitem[Smith et~al.(2020)Smith, Elsen, and De]{Smith_ICML_2020}
Samuel~L. Smith, Erich Elsen, and Soham De.
\newblock On the generalization benefit of noise in stochastic gradient descent.
\newblock In \emph{Proceedings of the 37th International Conference on Machine Learning}, ICML'20. JMLR.org, 2020.

\bibitem[Thomas et~al.(2007)Thomas, Luk, Leong, and Villasenor]{Thomas_2007}
David~B. Thomas, Wayne Luk, Philip~H.W. Leong, and John~D. Villasenor.
\newblock Gaussian random number generators.
\newblock \emph{ACM Comput. Surv.}, 39\penalty0 (4):\penalty0 11–es, nov 2007.
\newblock ISSN 0360-0300.
\newblock \doi{10.1145/1287620.1287622}.
\newblock URL \url{https://doi.org/10.1145/1287620.1287622}.

\bibitem[Wang et~al.(2021)Wang, Huang, Gao, and Calmon]{Wang_NEURIPS_2021}
Hao Wang, Yizhe Huang, Rui Gao, and Flavio Calmon.
\newblock Analyzing the generalization capability of sgld using properties of gaussian channels.
\newblock In M.~Ranzato, A.~Beygelzimer, Y.~Dauphin, P.S. Liang, and J.~Wortman Vaughan, editors, \emph{Advances in Neural Information Processing Systems}, volume~34, pages 24222--24234. Curran Associates, Inc., 2021.
\newblock URL \url{https://proceedings.neurips.cc/paper_files/paper/2021/file/cb77649f5d53798edfa0ff40dae46322-Paper.pdf}.

\bibitem[Welling and Teh(2011)]{Welling_ICML_2011}
Max Welling and Yee~Whye Teh.
\newblock Bayesian learning via stochastic gradient langevin dynamics.
\newblock In \emph{Proceedings of the 28th International Conference on International Conference on Machine Learning}, ICML'11, page 681–688, Madison, WI, USA, 2011. Omnipress.
\newblock ISBN 9781450306195.

\bibitem[Wen et~al.(2016)Wen, Wu, Wang, Chen, and Li]{Wen_2016}
Wei Wen, Chunpeng Wu, Yandan Wang, Yiran Chen, and Hai Li.
\newblock Learning structured sparsity in deep neural networks.
\newblock In D.~D. Lee, M.~Sugiyama, U.~V. Luxburg, I.~Guyon, and R.~Garnett, editors, \emph{Advances in Neural Information Processing Systems 29}, pages 2074--2082. Curran Associates, Inc., 2016.

\bibitem[Wu et~al.(2020)Wu, Hu, Xiong, Huan, Braverman, and Zhu]{Wu_ICML_2020}
Jingfeng Wu, Wenqing Hu, Haoyi Xiong, Jun Huan, Vladimir Braverman, and Zhanxing Zhu.
\newblock On the noisy gradient descent that generalizes as sgd.
\newblock In \emph{Proceedings of the 37th International Conference on Machine Learning}, ICML'20. JMLR.org, 2020.

\bibitem[Xiao et~al.(2017)Xiao, Rasul, and Vollgraf]{xiao2017fashionmnist}
Han Xiao, Kashif Rasul, and Roland Vollgraf.
\newblock Fashion-mnist: a novel image dataset for benchmarking machine learning algorithms, 2017.

\bibitem[Xie et~al.(2021)Xie, Sato, and Sugiyama]{xie_ICLR_2021}
Zeke Xie, Issei Sato, and Masashi Sugiyama.
\newblock A diffusion theory for deep learning dynamics: Stochastic gradient descent exponentially favors flat minima.
\newblock In \emph{International Conference on Learning Representations}, 2021.
\newblock URL \url{https://openreview.net/forum?id=wXgk_iCiYGo}.

\bibitem[Xu et~al.(2018)Xu, Chen, Zou, and Gu]{Xu_NEURIPS_2018}
Pan Xu, Jinghui Chen, Difan Zou, and Quanquan Gu.
\newblock Global convergence of langevin dynamics based algorithms for nonconvex optimization.
\newblock In S.~Bengio, H.~Wallach, H.~Larochelle, K.~Grauman, N.~Cesa-Bianchi, and R.~Garnett, editors, \emph{Advances in Neural Information Processing Systems}, volume~31. Curran Associates, Inc., 2018.
\newblock URL \url{https://proceedings.neurips.cc/paper_files/paper/2018/file/9c19a2aa1d84e04b0bd4bc888792bd1e-Paper.pdf}.

\bibitem[Zamir and Feder(1996)]{Zamir_1996}
R.~Zamir and M.~Feder.
\newblock On lattice quantization noise.
\newblock \emph{IEEE Transactions on Information Theory}, 42\penalty0 (4):\penalty0 1152--1159, 1996.
\newblock \doi{10.1109/18.508838}.

\bibitem[Zhang et~al.(2019)Zhang, Li, Nado, Martens, Sachdeva, Dahl, Shallue, and Grosse]{Zhang_NEURIPS2019}
Guodong Zhang, Lala Li, Zachary Nado, James Martens, Sushant Sachdeva, George Dahl, Chris Shallue, and Roger~B Grosse.
\newblock Which algorithmic choices matter at which batch sizes? insights from a noisy quadratic model.
\newblock In H.~Wallach, H.~Larochelle, A.~Beygelzimer, F.~d\textquotesingle Alch\'{e}-Buc, E.~Fox, and R.~Garnett, editors, \emph{Advances in Neural Information Processing Systems}, volume~32. Curran Associates, Inc., 2019.
\newblock URL \url{https://proceedings.neurips.cc/paper_files/paper/2019/file/e0eacd983971634327ae1819ea8b6214-Paper.pdf}.

\bibitem[Zhang et~al.(2022)Zhang, Wilson, and Sa]{Zhang_ICML_2022}
Ruqi Zhang, Andrew~Gordon Wilson, and Christopher~De Sa.
\newblock Low-precision stochastic gradient langevin dynamics.
\newblock In \emph{International Conference on Machine Learning, {ICML} 2022, 17-23 July 2022, Baltimore, Maryland, {USA}}, volume 162 of \emph{Proceedings of Machine Learning Research}, pages 26624--26644. {PMLR}, 2022.
\newblock URL \url{https://proceedings.mlr.press/v162/zhang22ag.html}.

\bibitem[Øksendal(2003)]{Bernt_2003}
Bernt Øksendal.
\newblock \emph{Stochastic Differential Equations:An Introduction with Applications}.
\newblock Springer-Verlag, 6th edition, 2003.
\newblock ISBN 978-3-540-04758-2.

\end{thebibliography}
